\documentclass[epsfig,a4paper,12pt,titlepage]{book}
\usepackage{epsfig}
\usepackage{fancyhdr}
\usepackage{plain}
\usepackage{algorithm}  
\usepackage{algorithmic}

\usepackage{booktabs}

\usepackage{graphicx}
\usepackage{import}
\usepackage{xcolor}
\usepackage{amsmath}
\usepackage{amssymb}

\usepackage{times}
\usepackage{soul}
\usepackage{url}
\usepackage[hidelinks]{hyperref}
\usepackage[utf8]{inputenc}
\usepackage[small]{caption}
\usepackage{amsthm}
\usepackage[figuresright]{rotating}
\usepackage{graphicx} 
\usepackage{array}  

\usepackage{textcomp}
\usepackage{multirow}

\makeatletter
\renewcommand\part{%
  \if@openright
    \cleardoublepage
  \else
    \clearpage
  \fi
  \thispagestyle{empty}%
  \if@twocolumn
    \onecolumn
    \@tempswatrue
  \else
    \@tempswafalse
  \fi
  \null\vfil
  \secdef\@part\@spart}
\makeatother

\newcommand{\clearemptydoublepage}{\newpage{\pagestyle{empty}\cleardoublepage}}

\makeindex
\begin{document}
\pagestyle{plain}

\newpage
\clearemptydoublepage
\thispagestyle{empty}
\begin{center}

\begin{figure}[h!]
    \centering
    \includegraphics[width=0.5\linewidth]{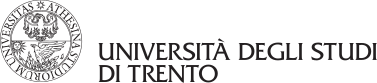}
\end{figure}

\hrulefill

DEPARTMENT OF INFORMATION ENGINEERING AND COMPUTER SCIENCE\\
\textbf{ICT International Doctoral School}\\

\vspace{1 cm} 
\Huge\textsc{Knowledge Graph Extension by Entity Type Recognition} \Large\textsc{}\\

\vspace{0.3 cm}

\begin{center}
\begin{tabular}{l}
\huge{Daqian Shi}\\
\end{tabular}
\vspace{0.3 cm}
\\
Ph.D. Degree Thesis
\vspace{0.3 cm}
\\
Academic Year 2019–2023

\end{center}
\vspace{1 cm} 
\begin{flushleft}
\begin{tabular}{ll}
\multicolumn{2}{l}{\large Advisor}\\
 & \large Prof. Fausto Giunchiglia\\
 & \large Universit\`a degli Studi di Trento\\
\end{tabular}
\end{flushleft}


\vspace{3 cm}

\hrulefill

\normalsize
January $2024$
\end{center}

\newpage
\thispagestyle{empty}
\large

{\bf \Huge Acknowledgements}

\vspace{1cm}

First of all, I wish to extend my appreciation and thanks to my advisor, Prof. Fausto Giunchiglia. Thanks for providing me with such an exceptional opportunity to study and work closely with him. I learned a lot from him, including the methodology for doing research, writing papers, and constructing social networks, which inspired me a lot at the beginning of being a young researcher. Meanwhile, he encouraged me and trusted me a lot during the past four years, I could not finish my Ph.D. study without his help and encouragement. I would say working with him really broadened my eyes and constructed the foundation of my future research career. Also thanks to \emph{InteropEHRate} project and the EU Horizon 2020 program that underpinned our research. Additionally, my deepest thanks to all the other professors and my colleagues for their invaluable help and support. I am immensely thankful for the opportunity to be involved in the LiveSchema project, which allows me to connect with passionate individuals from diverse backgrounds. Last but certainly not least, please allow me to convey my sincerest gratitude and love to my wife Xiaolei Diao, my mother Guangxu He, and my father Rongfu Shi for their companionship throughout my doctoral journey.

\vspace{2cm}

\hfill Daqian Shi 

\hfill January 2024 

\newpage
\thispagestyle{empty}
\large

{\bf \Huge Abstract}

\vspace{1cm}

Knowledge graphs have emerged as a sophisticated advancement and refinement of semantic networks, and their deployment is one of the critical methodologies in contemporary artificial intelligence. The construction of knowledge graphs is a multifaceted process involving various techniques, where researchers aim to extract the knowledge from existing resources for the construction since building from scratch entails significant labor and time costs.  However, due to the pervasive issue of heterogeneity, the description diversity across different knowledge graphs can lead to mismatches between concepts, thereby impacting the efficacy of knowledge extraction. This Ph.D. study focuses on automatic knowledge graph extension, i.e., properly extending the reference knowledge graph by extracting and integrating concepts from one or more candidate knowledge graphs. We propose a novel knowledge graph extension framework based on entity type recognition. The framework aims to achieve high-quality knowledge extraction by aligning the schemas and entities across different knowledge graphs, thereby enhancing the performance of the extension. This paper elucidates three major contributions: (i) we propose an entity type recognition method exploiting machine learning and property-based similarities to enhance knowledge extraction; (ii) we introduce a set of assessment metrics to validate the quality of the extended knowledge graphs; (iii) we develop a platform for knowledge graph acquisition, management, and extension to benefit knowledge engineers practically. Our evaluation comprehensively demonstrated the feasibility and effectiveness of the proposed extension framework and its functionalities through quantitative experiments and case studies.

\vspace{1cm}

\clearemptydoublepage

\pagenumbering{roman}
\tableofcontents
\clearemptydoublepage
\listoftables
\clearemptydoublepage
\listoffigures
\clearemptydoublepage

\pagenumbering{arabic}

\chapter{Introduction}

\section{The Context}
\label{sec:context}

Semantic networks, serving as a depiction and representation of human knowledge, graphically illustrate concepts and their interconnections through the nodes and edges, thereby offering an intuitive method for understanding and processing complex information. They are instrumental in establishing a universal medium for data exchange, enabling the sharing and processing of data by both automated systems and individuals \cite{hogan2021knowledge}. Semantic networks play an essential role in information systems that rely on structured knowledge. Researchers have tried to support intelligent systems by formalizing knowledge in the classic artificial intelligence research since the 1980s \cite{russel2003artificial}, enhancing not only the quality and precision of information retrieval but also significantly impacting the field of artificial intelligence.

The notion of knowledge graphs emerges as a sophisticated advancement and refinement of semantic networks. While both aim to articulate knowledge through network structures, knowledge graphs surpass traditional semantic networks in both scale and complexity \cite{shalaby2016entity}. Knowledge graphs yield a more enriched and comprehensive depiction of knowledge, aggregating extensive data from diverse sources. The applications of knowledge graphs in computer science are immensely diverse. They not only facilitate machine comprehension and natural language processing (NLP) but also occupy a central role in the evolution of knowledge representation and the design of reasoning algorithms. Recently, with the emergence of linked open data \cite{bizer2011linked} sources like DBpedia \cite{lehmann2015dbpedia} and Linked Open Vocabularies (LOV) \cite{vandenbussche2017linked}, and by Google’s announcement of the Google knowledge graph, knowledge graph developing becomes a prominent research topic in the knowledge representation community. 

The construction of knowledge graphs is a multifaceted process that involves various methodologies, according to different scenarios and applications. Typically, constructing a knowledge graph from scratch necessitates extensive manual efforts, where researchers introduced crowd-sourcing to gather information from a large number of individuals to achieve comprehensive data \cite{bordes2014constructing}. This approach is beneficial for covering a wide range of topics and gathering real-time data, but it needs to ensure the accuracy and reliability of the provided information. For instance, the construction and maintenance of Wikidata \cite{vrandevcic2014wikidata} and Freebase \cite{bollacker2008freebase} heavily rely on community contributions. Thus, aiming to avoid collecting data from scratch, automatic information extraction methods are introduced during knowledge graph construction. NLP techniques are applied, e.g., named entity recognition (NER) and relation extraction (RE), to identify and categorize entities and their interrelations within unstructured data \cite{mondal2021end}, where NELL \cite{carlson2010coupled} and PROSPERA \cite{nakashole2011scalable} being notable examples constructed by NLP-based methods. Meanwhile, researchers explore knowledge extraction from structured data sources, such as databases or spreadsheets \cite{lehmann2015dbpedia}. For example, the construction of DBpedia exploits knowledge extracted from the large-scale, semi-structured knowledge base of Wikipedia. These information extraction approaches are particularly useful for collecting knowledge from vast amounts of data, although they can be challenging due to the complexity and ambiguity of language presentation.

With the increasing requirements of contemporary artificial intelligence systems for knowledge graphs, the efficient development of new knowledge graphs through the reuse of existing knowledge is emerging as a compelling research area. The principal idea entails the development of new knowledge graphs by repurposing existing knowledge graphs and resources, rather than developing from scratch \cite{pinto1999some}. This process includes merging information from diverse data sources to forge a more comprehensive knowledge graph, which involves the integration of high-quality knowledge graphs from disparate domains or the combining of structured data with text data, e.g., research topics such as knowledge graph completion and ontology matching. A significant challenge in this endeavor is to reconcile inconsistencies and redundancies to preserve the integrity of the knowledge graph \cite{choi2023knowledge}.  Our study concentrates on one of the most challenging tasks in the knowledge reuse field, namely knowledge graph extension. The objective is to automatically augment the target knowledge graph by assimilating several resource knowledge graphs.

\section{The Problem}
In the context of knowledge reuse, the investigation into automatic knowledge graph extension constitutes a strategic response to the requirements of large-scale, dynamic, and complex knowledge management, which is driven by several practical reasons listed below. To begin with, given the vast array of existing knowledge graphs from various domains, coupled with the annual addition of thousands of knowledge graphs, there arises a demand for scalability \cite{chen2019identifying}. This necessitates the capability to rapidly process and integrate vast data volumes, thus markedly accelerating the extension of real-time data and enhancing the scale and depth of knowledge. Automatic extension significantly boosts efficiency, reducing the repetitive manual costs required by data scientists and domain experts. Additionally, it furnishes consistent algorithmic approaches to mitigate human biases in standardizing diverse data types and synchronizing different data sources. Manual extension of knowledge graphs, particularly for large-scale ones like DBpedia and YAGO, which encompass millions of entities, is labor-intensive and time-consuming \cite{gracia2007solving}. Therefore, there is a critical need for high-quality methodologies for automatic knowledge graph extension.

One of the major challenges for automatic extension lies in semantic heterogeneity. Semantic heterogeneity is inevitable due to the diversity in data formats, taxonomies, and sources, which is a manifestation of the broader diversity of the world and its descriptions \cite{lonsdale2010reusing}. Such diversity results in a high degree of variation in the structural granularity and terminology between different knowledge graphs \cite{hogan2021knowledge}. Specifically, two distinct diversities of description are found: linguistic diversity refers to the many-to-many alignments between words and their intended meanings, and knowledge diversity implies many-to-many alignments between entity types and the properties used to describe them \cite{giunchiglia2020entity}. The diversity of description poses a challenge in precisely locating target knowledge through general indexing methods from the vast data distribution. Meanwhile, the diversity of data sources brings difficulty to the uniformity of automatic knowledge management and processing. Thus, the process of knowledge graph extension should be organized as a cohesive framework consisting of scientifically planned stages, which should also encompass the capability of proper assessment of extension outcomes \cite{fumagalli2021ranking}.  

The main problems that we identify in this thesis are listed below:
\begin{itemize}
\item A standardized framework is requisite to properly model the automatic knowledge graph extension task. 
\item Precise aligning of knowledge graph concepts is essential to overcome the description diversity caused by semantic heterogeneity.
\item Appropriate assessment methods are needed to validate the effectiveness of the extended knowledge graph.
\item A unified platform is needed to customize knowledge graph extension services with consistent processing. 
\end{itemize}

\section{The Solution}
Modeling the task of knowledge graph extension as a framework encompassing multiple concrete stages is fundamental for facilitating automatic processing. In addressing the given user requirements, our emphasis lies on extending both the schema-level and instance-level knowledge graph. This task's modeling includes four stages: data preparation, entity type recognition, knowledge graph extension, and performance assessment. Data collection entails accumulating, organizing, and standardizing knowledge graphs and user requirements via a knowledge management platform. Entity type recognition is tasked with precisely aligning knowledge graph concepts at both schema and instance levels, respectively. The knowledge graph extension phase leverages the results of this recognition, utilizing a specially designed extension algorithm to integrate necessary knowledge into the reference knowledge graph. Subsequently, the extended knowledge graph is analyzed and assessed using proposed measurements, with those knowledge graphs that satisfy the established criteria being made available to knowledge engineers.

The description diversity problem arises whenever there is a need to exploit knowledge graphs from heterogeneous resources. To tackle this issue, we introduce a machine learning-based entity type recognition method, aimed at enhancing the aligning of various concepts within knowledge graphs by exploring a range of machine learning models. In addition, we propose a set of property-based metrics designed to measure the similarity between different entities and entity types. These metrics serve as an improvement to traditional string-based similarity measures, thereby further enhancing the performance of knowledge extraction by the proposed entity type recognition method.

To conduct a systematic evaluation of the extended knowledge graph, we propose an assessment measure anchored in categorization purpose, which we structured as a set of quantifiable metrics. These metrics were employed with classical ranking metrics at the assessment stage. Concurrently, we created an online knowledge management platform, namely LiveSchema, which principally aims to integrate the functionalities of the aforementioned framework into a cohesive suite of features to facilitate the automatic extension of knowledge graphs. Furthermore, LiveSchemea is developed with an extensive array of functionalities, encompassing aspects of knowledge acquisition, management, and visualization.

Corresponding to the problems, we conclude the main contributions of this thesis as follows:
\begin{itemize}
\item We introduce a framework for organizing functionalities into four stages, to achieve automatic knowledge graph extension. 
\item We propose an entity type recognition method exploiting machine learning models and property-based similarity metrics to enhance the performance of knowledge extraction.
\item We propose an assessment method by using a set of ranking metrics to validate the quality of the extended knowledge graph.
\item We develop the online platform LiveSchema to integrate the functionalities for knowledge graph acquisition, management, and extension services to benefit knowledge engineers. 
\end{itemize}

\section{Structure of the Thesis}
In Chapter 2, the state of the art on four related research topics are introduced. Firstly, We introduce the evolution of the methods for schema matching, including the lexical-based, contextual-based, and machine learning-based methods. Concurrently, we demonstrate the entity typing task and its corresponding methods, which are about recognizing the entity type of entities from diverse resources. Then, we talk about entity mapping methods which refer to the alignment of entities. Finally, we discuss knowledge graph completion, which is a broad research topic including several sub-topics, where knowledge graph extension is one of them. Additionally, we also demonstrate two sub-topics, namely knowledge graph refinement and self-completion. 

Chapter 3 introduces the modeling of the automatic knowledge graph extension task. We first introduce the terminology that appears in this thesis to improve the readability. We also define the extension task and the exploited data in symbolic language. Then, we detail the knowledge graph extension framework and its involved functionalities.

In Chapter 4, we primarily introduce the preparation for achieving the entity type recognition task. Firstly, we analyze the existing knowledge graphs and discuss the intuition of exploiting properties for measuring the similarity between entity types. We also introduced a knowledge formalization method based on formal concept analysis lattices. Then, we propose the property-based similarity metrics and theoretically compare them with existing similarity metrics. At last, we validate the proposed metrics by qualitative experiments. This chapter is based on the following publication:
\begin{itemize}
\item \textbf{Daqian Shi}, and Fausto Giunchiglia, 2023. \textit{Recognizing Entity Types via Properties [C].} In Proceeding of the 13th international conference on formal ontology in information systems.
\end{itemize}

Chapter 5 demonstrates the machine learning-based entity type recognition method. Firstly, the overall method and its components are introduced, including knowledge pre-processing, the calculation of property-based similarity metrics, and machine learning-based recognizers. We identify the specific cases of schema-level and instance-level recognition, and also the alignment of properties from different resources. A very detailed discussion of learning algorithm selection is presented, where we include classic machine learning models and also novel neural networks we investigated in various studies. Then, we focus on introducing the model training, giving the training details and parameter settings. Finally, we involve a set of experiments to evaluate our entity type recognition method and the proposed property-based similarity metrics, both quantitatively and qualitatively. Chapter 5 is based on the following publications:
\begin{itemize}
\item Fausto Giunchiglia, and \textbf{Daqian Shi}, 2021. \textit{Property-based Entity Type Graph Matching [C].} The 16th International Workshop on Ontology Matching co-located with the 20th ISWC. 

\item Lida Shi, Fausto Giunchiglia, Rui Song, \textbf{Daqian Shi} and Tongtong Liu, Xiaolei Diao, and Xu Hao, 2022. \textit{A Simple Contrastive Learning Framework for Interactive Argument Pair Identification via Argument-Context Extraction [C].} In Proceedings of the 2022 Conference on Empirical Methods in Natural Language Processing.

\item Jinge Wu, \textbf{Daqian Shi}, Abul Hasan, Honghan Wu, 2023. \textit{KnowLab at RadSum23: comparing pre-trained language models in radiology report summarization [C].} The 22nd Workshop on Biomedical Natural Language Processing and BioNLP Shared Tasks.


\item Jian Li, Ziyao Meng, \textbf{Daqian Shi}, Rui Song, Xiaolei Diao,  Jingwen Wang, and Hao Xu, 2023. \textit{FCC: Feature Clusters Compression for Long-Tailed Visual Recognition [C].} In Proceedings of the IEEE/CVF Conference on Computer Vision and Pattern Recognition.
\end{itemize}

Chapter 6 mainly demonstrates a set of assessment metrics for evaluating the quality of the extended knowledge graphs. Firstly, we introduce several existing metrics that are exploited in knowledge graph assessment. Then, we propose our metrics, namely Focus, whose primary idea is to measure the categorization purpose of the given entity types by their associated properties. The experimental results show the validation of Focus metrics. This chapter is based on the publication:
\begin{itemize}
\item Mattia Fumagalli, \textbf{Daqian Shi}, and Fausto Giunchiglia. \textit{Ranking Schemas by Focus: A Cognitively-Inspired Approach [C].} In Graph-Based Representation and Reasoning: 26th International Conference on Conceptual Structures, ICCS 2021.
\end{itemize}

In Chapter 7, we develop an online knowledge management platform, namely Liveschema, aiming to integrate all introduced functionalities into a cohesive system. We demonstrate the details of the data layer and service layer of the platform, providing a visual representation of Liveschema's development. Crucially, we introduce a knowledge graph extension algorithm, leveraging the functionalities offered by Liveschema. At last, we present several case studies to show the effectiveness of the extended knowledge graph. The content of this chapter is grounded in research from the following publications:
\begin{itemize}
\item \textbf{Daqian Shi}, Xiaoyue Li, and Fausto Giunchiglia, 2023. \textit{KAE: A Property-based Method for Knowledge Graph Alignment and Extension [J].} Journal of Web Semantics. (Under Reviewing)

\item Mattia Fumagalli, Marco Boffo, \textbf{Daqian Shi}, Mayukh Bagchi,  and Fausto Giunchiglia, 2023. \textit{Liveschema: A gateway towards learning on knowledge graph schemas [C].} In Proceeding of the 13th international conference on formal ontology in information systems.

\item \textbf{Daqian Shi}, Ting Wang, Hao Xing, and Hao Xu, 2020. \textit{A learning path recommendation model based on a multidimensional knowledge graph framework for e-learning [J].} Knowledge-Based Systems, 195, p.105618.

\item Ting Wang, \textbf{Daqian Shi}, Zhaodan Wang, Shuai Xu, and Hao Xu, 2020. \textit{MRP2Rec: exploring multiple-step relation path semantics for knowledge graph-based recommendations [J].} IEEE Access, 8, pp.134817-134825. 

\item Yang Chi, Fausto Giunchiglia, \textbf{Daqian Shi}, Xiaolei Diao, Chuntao Li, Hao Xu, 2022. \textit{ZiNet: Linking Chinese Characters Spanning Three Thousand Years [C].} Findings of the Association for Computational Linguistics: ACL 2022.
\end{itemize}

Finally, Chapter 8 presents the conclusions and future work respectively.

\chapter{State of the Art}
In this study, our objective is to address the challenge presented in the introduction chapter, specifically, identifying and assimilating pertinent knowledge from diverse data sources to augment the existing knowledge graph. This chapter delves into an exhaustive analysis of state-of-the-art methodologies that contribute to the knowledge graph extension. Our exploration encompasses studies related to both schema-level and entity-level knowledge graphs. Initially, we introduce schema aligning techniques, which are instrumental in aligning entity types and properties within knowledge graph schemas. Furthermore, we exhibit a series of entity typing and entity aligning approaches, aimed to correlate identical entities across disparate knowledge resources. Subsequently, we investigate various completion methods based on their design purpose, including knowledge graph refinement, knowledge graph completion, and knowledge graph extension.


\section{Schema Alignment}
The schema alignment task aims to determine the correspondences between concepts in knowledge graph schemas. This section investigates schema aligning methods based on different techniques, including lexical-based, contextual-based, and machine learning-based methods. 

\subsection{Lexical-based Methods}
In the early stages of the research, various neural language processing (NLP) techniques, including string-based metrics (N-gram, Levenshtein, etc.), syntactic operations (lemmatization, stop word removal, etc.), and semantic analysis (synonyms, antonyms, etc.) are applied since the semantically related entity types are most likely lexically related \cite{cheatham2013string}. For instance, edit-distance \cite{ristad1998learning} is applied to calculate the string similarity between two entity types \cite{li2004lom,resnik1995using} for entity type alignments. The main idea of such aligning methods is relatively straightforward, where correlated entity type pairs are identified if the string similarity result is higher than a specific threshold \cite{li2008rimom}. However, the limitation of the string-based aligning methods is that they cannot solve the word sense disambiguation problem. 

To solve this problem, Gracia et al. \cite{gracia2007solving} exploited a synonymy measurement to perform disambiguation for correctly aligning entity types from two schemas. Lin and Sandkuhl \cite{lin2008survey} introduced a synonym-improved similarity measurement, where WordNet \cite{fellbaum1998wordnet} is applied as a dictionary to look for synonyms for disambiguation and achieved better performance on the schema aligning task. To refine health records searching outputs, Olivier et al. \cite{cure2015formal} introduced a aligning method based on formal concept analysis (FCA) which assists users in defining their queries. Meanwhile, a bottom-up schema merging approach was proposed where FCA lattices were used to keep track of the schema hierarchy \cite{stumme2001fca}. Sun et al, \cite{sun2015comparative} reviewed a wide range of string similarity metrics and proposed the entity type alignment method by selecting similarity metrics in different scales. Although string-based methods can lead to effective performance in some cases, selecting the right metric for aligning specific datasets is the most challenging part.

\subsection{Contextual-based Methods}
Researchers also explore the contextual information that is critical in knowledge graphs, for identifying semantically related entity types. Such studies suppose that two entity types have a higher possibility of being aligned if they have the same super-class or sub-class. Giunchiglia et al. \cite{giunchiglia2004s} developed S-match system that can calculate the semantic relation between two entity types by taking the schemas as trees, where five semantic relations are introduced to identify if two entity types are semantically related. LogMap \cite{jimenez2011logmap} is a logic-based schema aligning method with built-in reasoning and diagnosis capabilities, where a lexical matcher is used to provide evidence for reasoning contextually related entity types. The authors structurally indexed the input schemas for presenting the anchor mapping and finally output the confidence value of each of the possible aligning pairs. AML \cite{faria2013agreementmakerlight} introduced a schema aligning system that consists of a string-based matcher and a structure-based matcher, building internal correspondences by exploiting \textit{is-a} and \textit{part-of} relationships. Furthermore, Niepert et al. \cite{niepert2010probabilistic} presented a probabilistic-logical framework to map entity types by Markov logic reasoning.

\subsection{Machine Learning-based Methods}
Machine learning (ML) is a widespread technique in many research areas, and it has also been applied in schema aligning methods. Such methods model the entity type aligning task as a classification task under supervised learning constraints, trying to encode the information like lexical or contextual similarities as features for model training. In this case, classifiers are designed to recognize the correspondences between two entity types by learning from aligned and unaligned entity type pairs. Faria et al. \cite{faria2014agreementmakerlight} introduced an automated system for large-scale schema aligning, where an ML-based component is introduced for aligning entity types and properties between two schemas. Amrouch et al, \cite{amrouch2016decision} developed a decision tree model by exploiting lexical and semantic similarities of the entity type labels to align schemas. By encoding the lexical similarity of the superclass and subclass as structure similarity, Bulygin and Stupnikov \cite{bulygin2019applying} improved the former method and achieved promising results. Bento et al. \cite{rauchbauer2020proceedings} exploit improved convolutional neural networks (CNNs)  for effectively extracting information from features.

Unsupervised learning is more likely applied when dealing with dedicated problems. For solving the large-scale schema aligning, Algergawy et al. \cite{algergawy2011clustering} proposed a clustering-based aligning approach that breaks up the large aligning problem into smaller aligning problems. They also introduced a seeding-based clustering method \cite{algergawy2015seecont} which can decrease the complexity of large-scale schema aligning. Representation learning is commonly applied with ML-based methods for schema aligning. For instance, Nkisi-Orji et al. \cite{nkisi2019ontology} calculated semantic similarities between entity types by applying word embedding; then the similarities were used to train a random forest model as a classifier. DeepAlignment \cite{kolyvakis2018deepalignment} presented an unsupervised aligning by pre-trained word vectors and generating ontological entity type descriptions. By encoding entity types as vectors into latent space, Zhu et al. \cite{zhu2017iterative} introduced a schema alignment method via joint knowledge graph embeddings, which can align semantically related entity types by calculating the distance of two vectors.

\section{Entity Typing}

Entity typing is a key process of relation extraction systems and many NLP applications, which aims to map entities with concrete entity types. According to the different usage and motivation, studies on entity typing focus on three main directions: (1) recognizing the type of \textit{entities from text} \cite{xin2018improving}; (2) recognizing the type of entities from the single knowledge graph for \textit{knowledge graph self-completion} \cite{yaghoobzadeh2017multi}; (3) recognizing the type of entities from different knowledge graphs for \textit{knowledge graph extension} \cite{sleeman2015entity}.
In the field of entity typing for natural language texts, Ling and Weld model entity typing as a classification task, where 112 fine-grained entity types are pre-set as tags to classify the words in natural language texts \cite{ling2012fine}. They also introduce a baseline method that includes a conditional random field (CRF) model for text segmentation and an adapted perceptron algorithm for entity type classification, respectively. Shimaoka et al. \cite{shimaoka2017neural} introduced a model training method by combining learned features with additional hand-crafted features, including syntactic heads and phrases, to improve model performance. Onoe and Durrett propose a generic encoder-decoder framework by exploiting bi-LSTM layers and the attention mechanism for disambiguating between related entities \cite{onoe2020fine}. By using pre-trained language models (PLMs), Ding et al. \cite{ding2022prompt} proposed an effective prompt-learning pipeline for few-shot entity typing.

Existing knowledge graphs often contain entities that lack the association with specific entity types, due to the limited performance of automatic knowledge extraction and reconstruction \cite{zhao2020connecting}. The objective of entity typing within knowledge graphs is to infer the absent entity types for those entities that are categorically part of the knowledge graph. The knowledge graph self-completion task requires that only the knowledge from the reference knowledge graph itself can be utilized. Thus, Zou et al. \cite{zou2022knowledge} proposed a novel entity typing framework consisting of a relational aggregation graph attention encoder and a convolutional neural network (CNN)-based decoder, to encode the relation information between entities. Biswas et al. \cite{biswas2021cat2type} designed a graph embedding-based method, Cat2Type, which exploits the Wikipedia category graph embedding to enhance the learning of missing entity types. Obraczka et al. \cite{obraczka2021embedding} uses knowledge graph embeddings and attribute similarities as inputs for an ML classifier for generic entity resolution with several entity types.

Different from the former two tasks, our study focuses on the field of recognizing the type of entities from other knowledge graphs for extending the reference knowledge graph automatically, which is also named entity type recognition. In this field, some dedicated methods are proposed for specific datasets  \cite{dsouza2021towards, shalaby2016entity}. Rather than using label-based methods, some previous studies consider properties as a possible solution,  \cite{sleeman2015entity,sleeman2013type} propose a solution by modeling entity type recognition as a multi-class classification task. However, a pre-filtering step is needed since only properties shared across all candidates are counted for training and testing, which means there will be a few properties remaining after such filtering and a large amount of critical information will be discarded. Thus, the adaptation of such methods will be limited when applied in practice.  Giunchiglia and Fumagalli \cite{giunchiglia2020entity} propose a set of metrics for selecting the reference knowledge graph to improve the above method, which achieves improved performance with the support of a large number of knowledge graphs. However, there are still limitations since these studies consider all properties with the same weight and neglect to distinguish properties that will contribute differently during entity type recognition.  

\section{Entity Aligning}
The entity alignment task aims to find entities in different knowledge graphs referring to the same real-world identity, which is a process of aligning concrete entities between different knowledge graphs or knowledge bases \cite{sun2018bootstrapping}. Researchers have tried string-based similarities to identify similar entities based on their names, textual descriptions, and attribute values. As an improvement, researchers investigate knowledge graph embedding techniques, which aim to enhance the deep representation of target entities. Hu et al. \cite{hu2019multike} proposed an entity aligning framework that leverages multi-view knowledge graph embeddings. Trisedya et al. \cite{trisedya2019entity} exploited large numbers of attribute triples existing in the knowledge graphs and generated attribute character embeddings for aligning entities. DeepAlignment proposed an unsupervised method that refines pre-trained word vectors aiming at deriving ontological entity descriptions that are dedicated to the entity aligning task \cite{kolyvakis2018deepalignment}. Georgala et al. \cite{ngomo2020applying} proposed a specific process for exploiting semantically linked entities to increase the aligning efficiency. Besides, FCA is also widely applied for entity aligning. For instance, FCA-Map \cite{chen2019identifying} introduced a novel aligning system for large knowledge graph aligning by exploiting FCA lattices to describe how the lexical tokens are shared and extracted across knowledge graphs.

\section{Knowledge Graph Completion}
The knowledge graph completion task refers to the process of extending the target knowledge graph by extracting knowledge from additional resources, and it has been regarded as a promising topic in the research community. In this section, we focus on three types of knowledge graph completion methods, including refinement, self-completion, and extension.

\subsection{Knowledge Graph Refinement}
Knowledge refinement refers to the process of error detection and correction of knowledge graphs. Standard refinement methods include erroneous type assertions, correcting erroneous relations, detecting erroneous literal values and erroneous interlinks. Reasoning-based methods are widely used for error detection and correction. Jens and Lorenz \cite{lehmann2010ore} introduced a system for repairing and enriching knowledge graphs, which allows extending a knowledge graph through semi-automatic supervised learning. After adding axioms, the model can detect errors and repair the knowledge graph automatically. Ebeid et al. \cite{ebeid2021biomedical} demonstrated a reasoning-based method that comprehensively combines the knowledge graph embedding and logic rules that support triplets for eliminating conflicts and noises in biomedical knowledge graphs. As an opposite solution, Dong et al. presented a classification method to distinguish relations that should hold in knowledge graphs from those that should not \cite{dong2014knowledge}. They proposed a supervised deep-learning model for fusing knowledge from diverse resources.

\subsection{Knowledge Graph Self-completion}

Knowledge graph self-completion aims to add missing knowledge by completing the entity/property in the graph without other resources. Existing approaches include graph embedding models, the path ranking algorithm, and rule evaluation models. Socher et al. \cite{socher2013reasoning} introduced an expressive tensor neural network for reasoning over relationships between entities. Jiang et al. \cite{jiang2017attentive} introduced a dedicated recurrent neural network (RNN) architecture that can complete the complex relationship between entities by decomposing the path into a series of relationships. Embedding methods like pairwise-interaction differentiated embeddings model \cite{zhao2015knowledge} predict the possible truth of additional facts to extend the knowledge base by embedding pairwise entity relations into the lower-dimensional vector representation. Zhao et al \cite{zhao2020connecting} introduced a local-global strategy to jointly learn knowledge from existing entity type assertions, aiming at inferring possible missing entity type instances, which achieved promising results on various benchmarks. Biswas et al. \cite{biswas2022entity} present GRAND, a novel approach that leverages several graph walk strategies in graph embedding together with the deep representation of textual entity descriptions to improve the performance of recognizing missing entity types. Such graph embedding-based and deep representation-based methods provide new solutions to KG-related tasks, including the self-completion task. However, such methods require massive data during the training procedure for obtaining powerful deep representations, which limits the application on lots of small-scale KGs. At the same time, these data-driven models also request balanced data categories with enough samples (in the KGs they need entities and entity types), which is another challenge for the training based on KGs. Thus, few-shot learning and deep representation learning methods are also discussed to be applied in knowledge graph self-completion tasks, e.g., Zhang et al. \cite{zhang2020few} introduced a few-shot relation learning model for discovering facts of new relations with few-shot references in KGs.

\begin{table*}[!t]
\centering

\caption{A summary of all related topics and their representative target tasks.}
  \label{comparisonOfSOTA}

\begin{tabular}{c|c|c}

\toprule
\textbf{Topics} & \textbf{Related Concepts} & \textbf{Target Tasks} \\ \midrule

\multirow{2}{*}{Schema Alignment} & \multirow{2}{*}{Property, Entity Type} & \{Property - Property\} Alignment \\ & & \{Entity Type - Entity Type\} Alignment \\ \midrule

Entity Typing       & Entity, Entity Type  & \{Entity - Entity Type\} Alignment  \\ \midrule
Entity Aligning        & Entity  & \{Entity - Entity\} Alignment  \\ \midrule
\multirow{3}{*}{KG Refinement}       & \multirow{3}{*}{Property, Entity} & Entity Type Assertion
  \\ & & Property Correction \\ & & Erroneous Value Detection \\ \midrule
\multirow{2}{*}{KG Self-completion}         & \multirow{2}{*}{Property, Entity} & \{Entity - Entity Type\} Alignment \\ & & Property Completion  \\ \midrule
\multirow{4}{*}{KG Extension}         & \multirow{4}{*}{Property, Entity, Entity Type} & \{Entity Type - Entity Type\} Alignment \\ & & \{Property - Property\} Alignment \\& & \{Entity - Entity Type\} Alignment\\ & & Concept Merging\\ 
\bottomrule

\end{tabular}
\end{table*}

\subsection{Knowledge Graph Extension}
Knowledge graph extension aims to integrate additional knowledge from other knowledge graphs. It is different from the self-completion task which aims to add missing knowledge (concepts and properties) into the reference knowledge graph without exploiting other resources. In the context of the semantic web, most of the current extension cases locate the corresponding entity pairs in knowledge graphs and then directly merge knowledge graphs by taking their union \cite{hogan2021knowledge}. Several rule-based approaches for integrating knowledge graphs have been proposed, which are given in terms of theories of classical first-order logic and rule bases. They either cast rules into classical logic or limit the interaction between rules and schemas \cite{steve1998integrating,fahad2015initial}. As one of the rules-based methods, Bruijn et al. \cite{bruijn2011embedding} presented three embeddings for ordinary and disjunctive nonground logic programs under the stable model semantics to select knowledge graphs for integration. Since the current work on knowledge graph extension does not consider the schema as input, Wiharja et al. \cite{wiharja2018more} improved the correctness of knowledge graph combination based on schema-aware triple classification, which enables sequential combinations of knowledge graph embedding approaches. However, knowledge graph extension by the above-mentioned methods lacks the context information and taxonomy standard of the knowledge graph, which will decrease the quality of the output integrated knowledge graph and lead to ambiguities. Meanwhile, these approaches only work for schemas in specific-domain knowledge graphs and still have limitations when applied in practice. Novel methods for automatic schema integration are needed to extend general-domain knowledge graphs efficiently and accurately.

\chapter{Task Statements and Objectives}
\section{Introduction}
In this chapter, our primary focus is on the rational modeling of the knowledge graph extension task, specifically constructing the target task into a framework comprising several end-to-end sub-tasks. In this process, we introduce key settings utilized in this study and define them using symbolic language to ensure the corresponding concepts are consistently applied in subsequent chapters. We also explain several technical terminologies to clarify their specific meanings as presented in the thesis. Subsequently, we provide an overview of the knowledge graph extension framework, including detailed introductions of the framework's component functionalities and discussions of the challenges encountered during the implementation.

\section{Definition of Terminology}

\subsection{Data Descriptions}
A knowledge graph is conceptualized as a hierarchy of concepts, wherein properties are utilized to characterize these concepts. To demonstrate the subsequent contents of this study, we formulate the following foundational definitions.

\begin{figure}[!t]
    \centering
    \includegraphics[width=0.87\linewidth]{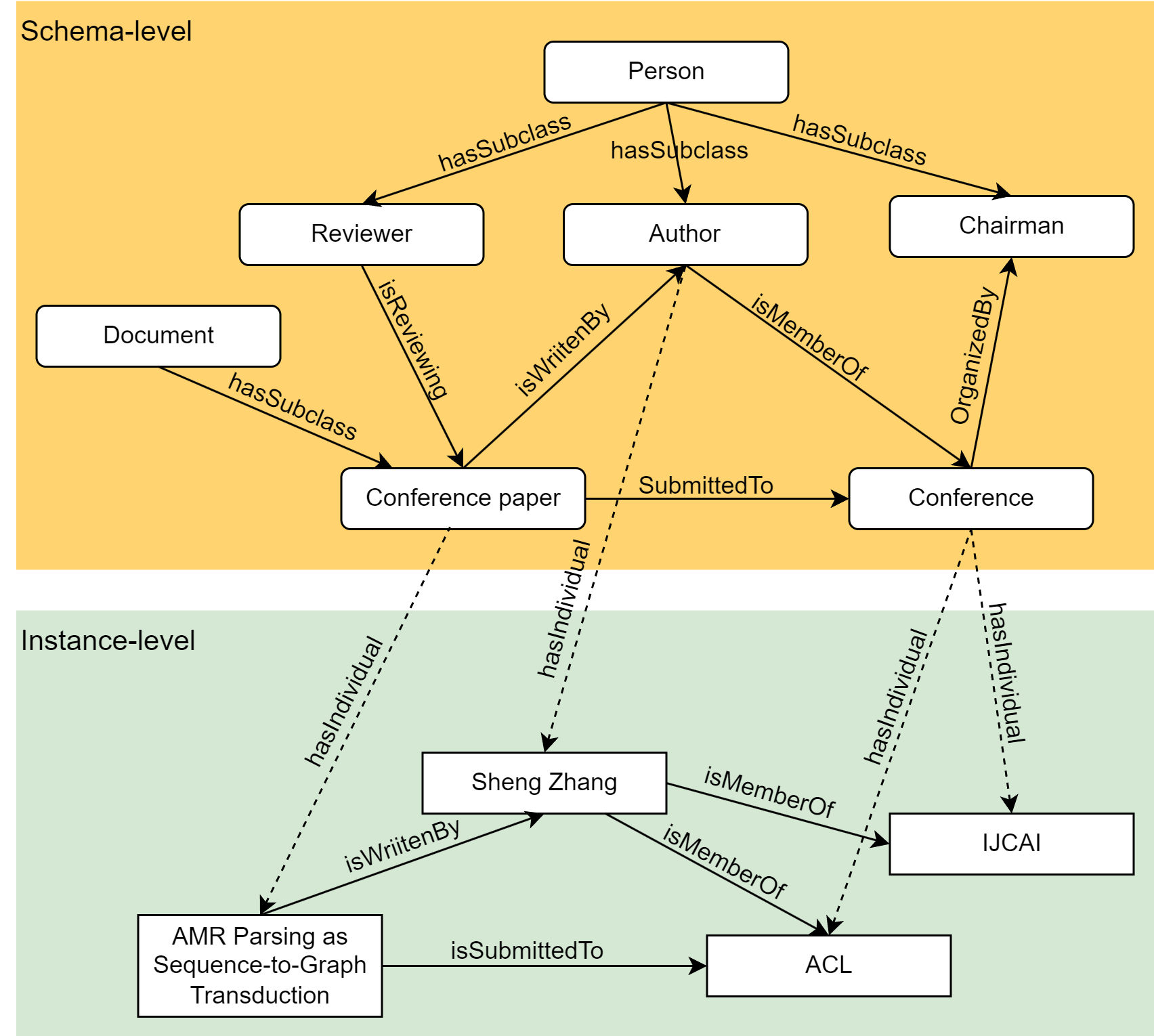}
    \caption{Example of schema-level and instance-level knowledge graphs.}
    \label{schema-instance}
\end{figure}

\noindent\textbf{Schema-level knowledge graph:}\\
A schema constitutes a framework that delineates categories of information and their interrelationships \cite{giunchiglia2017teleologies}. People create a schema as a system to organize extant knowledge in the real world. Within the realm of knowledge representation, a schema-level knowledge graph refers to the schema layer used to organize the entities in the knowledge graph. Structurally, a schema manifests as a graph-like configuration, comprising nodes and edges. Here, a node represents a category of entities, termed an entity type; whereas an edge signifies a potential relationship between two distinct entity types, known as a property. Specifically, we define the schema of a knowledge graph and its intrinsic relations as $KGS = \langle E, P, R \rangle$, where $E = \{E_1, ..., E_n\}$ being the entity types, $P = \{p_1, ..., p_m\}$ being the set of properties, $R = \{\langle E_i, T(E_i) \rangle|E_i \in E \}$ being the set of correspondences between entity types and properties, and function $T(E_i)$ returns properties associated with $E_i$. In Figure \ref{schema-instance}, we take an example from a specific domain knowledge graph \cite{jimenez2019logmap}. The yellow area illustrates a segment of the schema-level knowledge graph, nodes like \textit{Author} and \textit{Chairman} represent entity types, and edges define probable properties between entity types, such as \textit{an author can be a member of a conference}. A schema delineates a high-level framework for entities within the knowledge graph to follow, which offers many pivotal benefits \cite{paulheim2017knowledge}, such as: \textit{i).} human comprehensibility; \textit{ii).} a stable and distinct perspective over a flow of varied and multiple data streams; and \textit{iii).}  a structure resembling a tree or grid, enabling the location of each piece of information by sequentially addressing a specific set of questions. These benefits facilitate the depiction of high-efficiency solutions to extensive categorization challenges, specifically those about storage and retrieval.

\noindent\textbf{Instance-level knowledge graph:}\\
The instance-level knowledge graph can likewise be depicted as a graph comprising nodes and edges. In such a graph, an entity represents a specific instance of an entity type, and a relation signifies a specific property between two entities. Specifically, as for the entities $I$, we define $I = \{I_1, ..., I_l\}$ being the set of entities in the knowledge graph $KG$, where each entity $I_i$ can be identified by a specific entity type $E$. $t(I_i)$ refers to the set of associations between entities and properties. We consider that the property $p_i$ is used to describe an entity type $E_i$ or an entity $I_i$ when the property belongs to set $T(E_i)$ or $t(I_i)$, respectively. The green area in Figure \ref{schema-instance} exemplifies a portion of the instance-level knowledge graph. All nodes within this green area are entities, and the edges denote specific properties between pairs of entities. For example, the entity \textit{Sheng Zhang}, in the capacity of an author, is depicted as a member of the \textit{ACL} conference. Entities are crucial for representing real-world objects and establishing complex relationships, serving as fundamental components in the construction and conveyance of information within knowledge graphs.

\subsection{Task Statements}
In this section, we introduce several critical tasks that are integral to the implementation of automatic knowledge graph extension.

\begin{figure}[!t]
    \centering
    \includegraphics[width=0.95\linewidth]{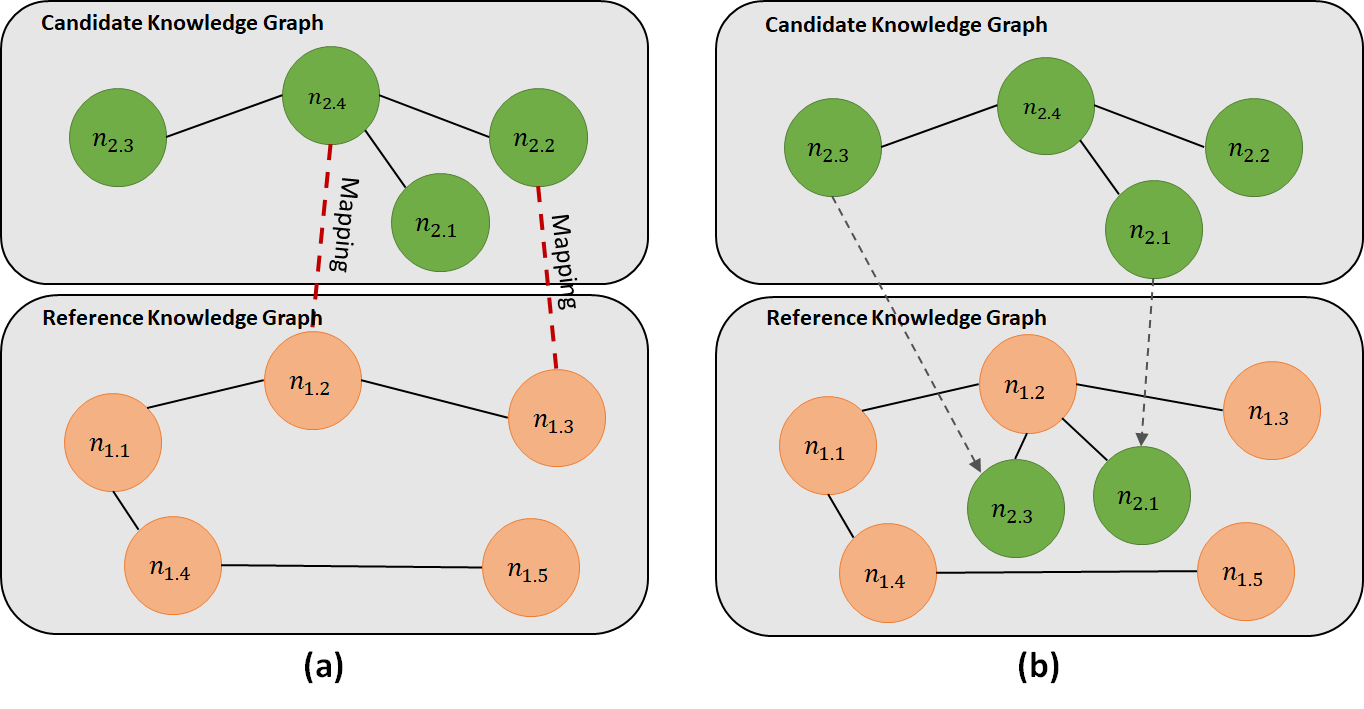}
    \caption{Demonstrations of the critical tasks involved in the extension framework.}
    \label{extension demo}
\end{figure}

\noindent\textbf{Entity type recognition:}\\
Referring to the conceptualization articulated in the pioneering study \cite{giunchiglia2020entity} by Giunchiglia and Fumagalli, we define the entity type recognition task. This task is primarily introduced to deal with the heterogeneity present in knowledge graphs. It is geared towards recognizing the types of concepts by leveraging the information encapsulated within the knowledge graph. Entity type recognition is generally modeled as a aligning task, with a focus on identifying correspondences across concepts within heterogeneous knowledge graphs. For instance, as depicted in Figure \ref{extension demo}(a), the node $n_{1.2}$ and $n_{1.3}$ are aligning with the node $n_{2.4}$ and $n_{2.2}$, respectively. Regarding different aligning objects, our study predominantly concentrates on (1) alignments between entity types, and (2) alignments between entities and entity types. Consequently, we have defined two tasks:

\begin{itemize}
    \item \textbf{Schema-level entity type recognition} corresponds to the definition of the schema-level knowledge graph, with its objective being the alignment between entity types that represent the same meaning across different schemas, e.g., the alignment between entity types \textit{Person} and \textit{Human being}. 

    \item \textbf{Instance-level entity type recognition} concentrates on entities from the instance-level knowledge graph, aiming at aligning the entities with their corresponding entity types, e.g., the alignment between the entity \textit{Usain Bolt} and entity type \textit{Athlete}. 
\end{itemize}
    
\noindent\textbf{Knowledge graph extension:}\\
Knowledge graph extension aims to extract and integrate knowledge from heterogeneous data resources into a knowledge graph. In our study, the heterogeneous data sources refer to a set of candidate knowledge graphs $KG_{cand}$, while we define the reference knowledge graph $KG_{ref}$ as the target being extended. Figure \ref{extension demo}(b) illustrates an extension example, where the node $n_{2.3}$ and $n_{2.1}$ are integrated into the $KG_{ref}$, being a subclass of the node $n_{1.2}$. Note that during the extension process, we focus on both the schema-level and instance-level knowledge graphs, where the reference knowledge graph schema is extended by candidate entity types $C_{cand}$, and the candidate entities $I_{cand}$ are integrated into the reference entity types $C_{ref}$.

\section{Knowledge Graph Extension Framework}
We model the knowledge graph extension task into a framework that encompasses multiple sub-tasks, as depicted in Figure \ref{Project framework}, where different sub-tasks are distinguished by colors. In response to automatic extension, we detail the objectives of each sub-task as follows:

\begin{figure}[!t]
    \centering
    \includegraphics[width=1\linewidth]{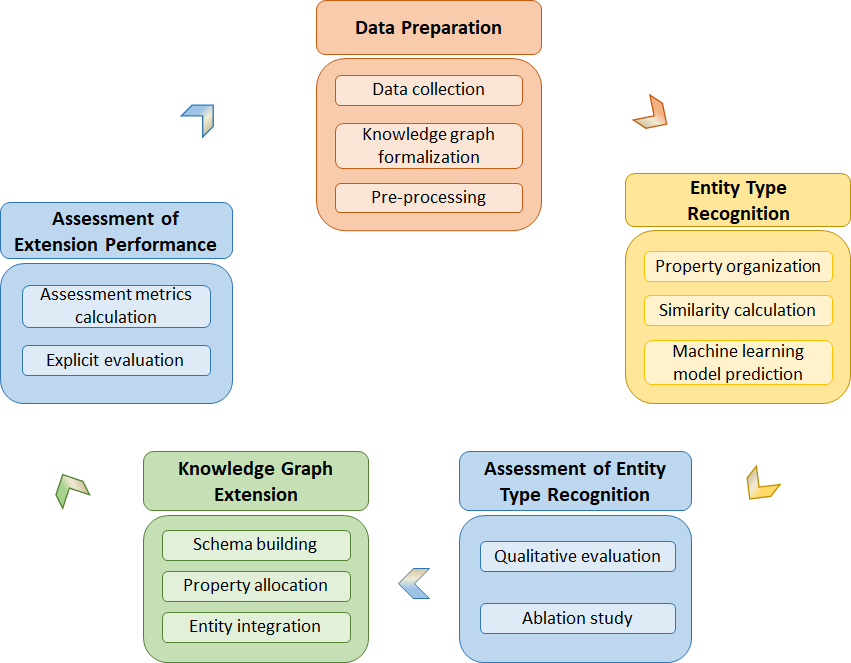}
    \caption{The proposed knowledge graph extension framework.}
    \label{Project framework}
\end{figure}

\begin{itemize}
    \item \textbf{Data preparation:} The data preparation stage marks the commencement of the entire framework, aiming to select reference and candidate knowledge graphs from the collected resources. Subsequently, the chosen knowledge graphs will be applied with pre-processing methods like parsing and formalization to support subsequent stages.

    \item \textbf{Entity type recognition:} Entity type recognition is responsible for accurately aligning concepts within the knowledge graph at both schema and instance levels. This process involves filtering and aligning properties, similarity metrics calculation, and selecting appropriate machine learning algorithms. The output of this stage including $<E_i, E_j>$ and $<E_i, I_k>$ alignments from schema-level and instance-level recognition respectively, where $E_i \in KG_{ref}$ and  $E_j, I_k \in KG_{cand}$.

    \item \textbf{Knowledge graph extension:} The knowledge graph extension stage employs a customized extension algorithm to integrate the requisite knowledge into the reference knowledge graph. By utilizing the results of schema-level entity type recognition, we aim to extend the reference schema by organizing entity types and properties $E_j, P_m \in KG_{cand}$. Subsequently, we will use the extension algorithm to insert entities $I_k \in KG_{cand}$ into the extended reference schema.

    \item \textbf{Performance assessment:} We assess the knowledge graph extension and its intermediary processes utilizing both implicit and explicit approaches. As depicted in Figure \ref{Project framework}, the blue blocks refer to two distinct assessment processes. Initially, we validate the results of the entity type recognition, quantifying the precision of the alignments using a range of metrics. Following this, we assess the quality of the extended knowledge graph, conducting this evaluation using the metrics we designed specifically.
\end{itemize}
Utilizing this framework enables us to achieve automatic knowledge graph extension, encompassing stages from data collection to the assessment of the knowledge graph. Additionally, the designed framework allows for the processing and integration of multiple candidate knowledge graphs through the implementation of multiple iterations. In the subsequent chapters, we will provide a detailed exposition for each stage in the demonstrated framework, encompassing both the implementation of its methods and the scientific problems addressed.



\section{Summary}
In this chapter, we introduce a framework designed for the automatic knowledge graph extension, serving as a guide for the planning and execution of subsequent studies. Initially, we present the data used for the extension task and outline several definitions of the corresponding tasks. Subsequently, we demonstrate the overall framework and architectural details of the designed framework.

\chapter{Similarity Measurements}

\section{Introduction}
In this chapter, we introduce the proposed property-based similarity measurements for measuring the relevance between knowledge graph concepts, which aims to support the implementation of two kinds of entity type recognition. The work present in this section is based on the publication \cite{giunchiglia2021property, shi2023recognizing}.

Entity type recognition constitutes a fundamental task in the process of knowledge graph extension, as entity types are instrumental in defining the schema for entities aggregated within a knowledge graph. When the entity types of candidate entities align with those of the reference knowledge graph, it facilitates an automatic integration of these entities into the graph. Consequently, there exists an imperative to evaluate the relevance between reference entity types and candidate entity types/entities. Such evaluation is crucial to achieving the objective of large-scale, automated entity type recognition, ensuring that the extension of the knowledge graph maintains structural and semantic integrity.

Existing relevance evaluation methods mainly exploit natural language processing (NLP) based \cite{cheatham2013string,sun2015comparative} and structure-based \cite{jimenez2011logmap,faria2013agreementmakerlight} techniques. Both techniques enforce entity type label alignment as a prerequisite. NLP-based methods utilize diverse lexical-based similarity metrics and synonym analysis to align entity type labels, which raises limitations when applied in practice. Labels may suggest a wrong entity type \cite{sleeman2015entity}, where the same concept can be labeled differently by knowledge graphs. For instance, an eagle can be labeled as \textit{Bird} in a general-purpose knowledge graph and \textit{Eagle} in a domain-specific knowledge graph. In turn, the same label may present different concepts in heterogeneous knowledge graphs, which will also lead to wrong recognition results. 

Structure-based methods model the knowledge graph hierarchy as an additional input, where the structure of a hierarchy is used to drive the label alignment, e.g., it is suggested to perform label alignment to sub-classes or super-classes. However, these methods may also mislead the conclusions as properties assigned to an entity type in the hierarchy are cumulative and depend only on nodes in the path from the root and, therefore, do not depend on the order by which they are assigned \cite{giunchiglia2020entity}. In addition, the difference in taxonomy between knowledge graphs will increase the impact of such mistakes, e.g., the super-class of entity type \textit{Eagle} can be \textit{Animal} in one knowledge graph and \textit{Bird} in another knowledge graph. On the other hand, the alignments between entity types and entities are also critical since the reference knowledge graph can also be extended by candidate entities that are not included in aligned entity types. However, it is not easy to identify such alignments due to the lexical labels are not applicable to alignment entities and entity types. For instance, entity \textit{apple} can be a company, a fruit, or the name of a pet, but there is no lexical similarity between entity \textit{apple} and entity types \textit{company}, \textit{fruit} and \textit{pet}. Thus, current NLP-based and structure-based methods poorly perform the entity type recognition task either.

As a solution to the above problems, in this chapter, we present three \textit{property-based} metrics to measure the entity type relevance for supporting better recognition performance. 
The similarity metrics characterize the role that properties have in the definition of given entity types from different aspects. They capture the main idea that the number of aligned properties affects the contextual similarity between entity types and entities. The proposed similarity metrics will contribute to the following machine learning-based entity type recognition algorithms from a property-aware perspective. 

In particular, we conduct a comprehensive statistical assessment of the distribution of entity types and properties across various knowledge graphs to demonstrate the intuition of measuring entity type similarities by properties. Our analysis encompasses knowledge graphs from diverse domains and scales. We introduce the \textit{knowledge lotus diagram} as a visual representation tool to visibly illustrate these statistics. Meanwhile, we introduce a formalization strategy for knowledge graphs, where we organize a knowledge graph and its inner alignments between nodes (including entity types and entities) and properties based on the use of formal concept analysis (FCA) lattices \cite{ganter2012formal}. Then, we introduce the detailed processing to calculate the proposed property-based similarities and the experimental analysis which validates the effectiveness of the proposed similarity measurements.

\section{Intuition and Analysis}
Generally, knowledge graphs can be modeled as knowledge contexts, where each context encapsulates a unique perspective on real-world concepts. The researcher noted a dual aspect within these knowledge contexts: the homogeneity in depicting identical concepts and the heterogeneity in delineating distinct concepts \cite{giunchiglia2020entity}. Suppose we aim to investigate the homogeneity between any two similar concepts, e.g., given any two entity types, their homogeneity relates:

\begin{itemize}
\item to how many properties they do share within one or more (or all) contexts;
\item to how many contexts they do share, associated with one or more (or all) of their properties.
\end{itemize}

\begin{table*}[!t]
\centering
\caption{Shared properties of entity type \textit{person} across different knowledge graphs.}
  \label{PropertyExample}
\resizebox{1\columnwidth}{!}{

\begin{tabular}{@{}lcccl@{}}
\toprule
\textbf{Contexts}      & & \textbf{Tot.} &  & \textbf{Shared Properties} \\  \midrule
OpenCyc \& DBpedia     & & 39  & & \textit{birth, education, title, activity, ethnicity, employer, status...} \\
OpenCyc \& Schema.org  & & 21  & & \textit{contact, suffix, tax, job, children, works, worth, gender, net...} \\
DBpedia \& FreeBase    & & 33  & & \textit{title, number, related, birth, parent, work, name...} \\ 
DBpedia \& Schema.org  & & 22  & & \textit{death, sibling, point, member, nationality, award, parents...} \\ 
\bottomrule                    
\end{tabular}
}
\end{table*}

Here we present several examples to show the homogeneity of entity types.  In Table \ref{PropertyExample}, we demonstrate shared properties of the entity type \textit{Person} across different knowledge contexts, e.g., \textit{birth} and \textit{education} are applied in both OpenCyc \cite{paulheim2017knowledge} and DBpedia \cite{auer2007dbpedia}. All these examples present features of properties, namely, the homogeneity for describing the same or similar concepts. Such homogeneity can also be found in Table \ref{ContextExample}, where the identical entity types are distributed across different contexts to some extent, e.g., entity type \textit{Event} and \textit{Place} are encoded by four well-known general-purpose knowledge graphs, OpenCyc, DBpedia, SUMO\cite{niles2003mapping} and Schema.org \cite{guha2016schema}, and over 30 entity types are shared between OpenCyc and DBpedia.

Meanwhile, we also find the heterogeneity we defined for distinguishing different concepts across contexts in Table \ref{PropertyExample}. For instance, a \textit{Person} can be distinguished from a \textit{Place} by the property \textit{birth}, which is a crucial step to identify the entity types. As a result, the intuition of utilizing properties for entity type recognition comes from the property being one of the most basic and critical elements for implicitly defining entity types \cite{giunchiglia2021property}. Within the realm of artificial intelligence, the purpose of categorization is typically realized through the employment of well-structured and effective information constructs, known as schemas, where prominent examples include \textit{knowledge graph schema layers} \cite{qiao2016knowledge}. For each schema, entity types play the role of categorization, and properties aim to draw sharp lines so that each entity in the domain falls determinedly either in or out of each entity type \cite{giunchiglia2019knowledge}. Meanwhile, we have the following observations:

\begin{itemize}
\item In a specific knowledge graph, a set of properties describes each entity type, whereas most of the properties are distinguishable according to the belonging entity types, and a small number of properties are shared across different entity types;
\item Same or similar properties are shared across different knowledge graphs for describing the same concepts.
\end{itemize}

\begin{table*}[!t]
\centering
\caption{Shared entity types across different knowledge graphs}
  \label{ContextExample}
\resizebox{1\columnwidth}{!}{

\begin{tabular}{@{}lcccl@{}}
\toprule
\textbf{Contexts}      & & \textbf{Tot.} &  & \textbf{Shared Entity Types} \\  \midrule
OpenCyc \& DBpedia \& SUMO \& Schema.org     & & 4  & & \textit{Person, Organization, Event, Place} \\
OpenCyc \& Schema.org \& DBpedia  & & 7  & & \textit{SportsTeam, SportsEvent, Action, ...} \\
OpenCyc \& Schema.org \& SUMO  & & 8  & & \textit{City, PoliticalParty, Language, Animal, ...} \\ 
OpenCyc \& DBpedia  & & 31  & & \textit{Hospital, SportsLeague, BodyOfWater, Sport, ...} \\ 
OpenCyc \& Schema.org  & & 16  & & \textit{Offer, EducationalOrganization, Message, Role, ...} \\ 
OpenCyc \& SUMO  & & 27  & & \textit{UnitOfMeasure, CreditAccount, ComputerNetwork, ...} \\ 
DBpedia \& SUMO  & & 12  & & \textit{Region, Ship, MilitaryUnit, Agent, ...}\\
\bottomrule                    
\end{tabular}
}
\end{table*}

\begin{figure*}[!t]
	\centering
	\includegraphics[width=1\linewidth]{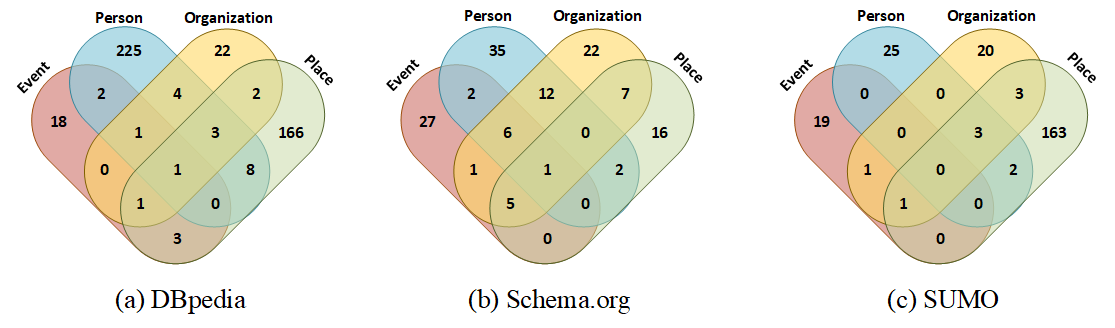}
	\caption{The shareability of properties that occur within each knowledge graph. 
	\label{Venn-etype}}
\end{figure*}

\begin{figure*}[!t]
	\centering
	\includegraphics[width=1\linewidth]{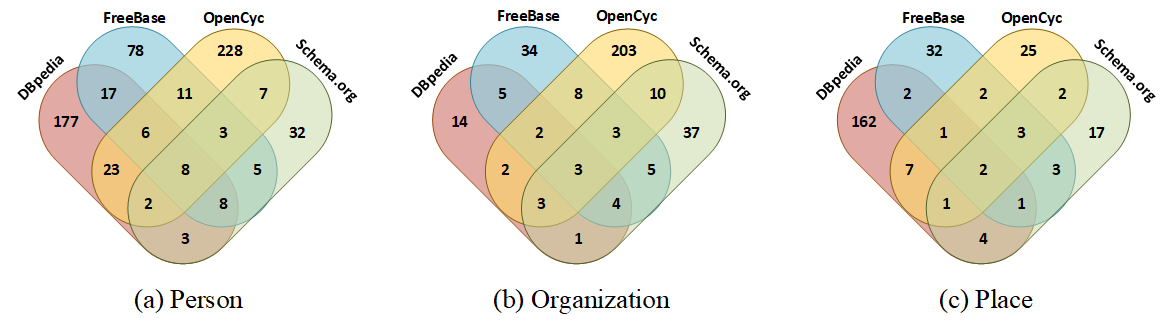}
	\caption{The shareability of properties that occur across different knowledge graphs. 
	\label{Venn-property}}
\end{figure*}

To visually present these observations, we introduce a special type of \textit{Venn graphs}, namely \textit{knowledge lotuses}, to prove our observations by representing the shareability of properties that occur within and across knowledge graphs. Knowledge lotuses provide a synthetic view of how different knowledge graphs or entity types overlap in properties \cite{giunchiglia2020entity}. We present several examples in Figure \ref{Venn-etype} and Figure \ref{Venn-property}, where we assume that we have several contexts built from (parts of) the representative knowledge graphs, i.e., OpenCyc, DBpedia, Schema.org, SUMO, and FreeBase \cite{bast2014easy}. Each value in a knowledge lotus refers to the number of shared properties.

Figure \ref{Venn-etype} demonstrates the intersection of properties amongst entity types within each knowledge graph. For example, within DBpedia, the entity types \textit{Person}, \textit{Organization}, and \textit{Place} share four properties, whereas the entity type \textit{Event} possesses 18 unique properties, not common with other entity types. The insights derived from the knowledge lotuses reveal that within individual knowledge graphs, various entity types exhibit a minimal overlap in their associated properties, suggesting that entity types can be distinctly identified based on their property profiles. Concurrently, Figure \ref{Venn-property} highlights the commonality of properties in knowledge graphs when depicting identical entity types. For instance, OpenCyc and DBpedia share 39 properties for the entity type \textit{Person}, while OpenCyc and Schema.org share 19 properties for \textit{Organization}. This observation underscores that identical or similar properties are recurrent across different knowledge graphs for representing similar concepts. Hence, it is imperative to leverage and quantify properties for the accurate recognition of entity types.

\section{FCA-based Formalization}

The main motivation for applying FCA-based formalization methods is that entity types are the basic elements that populate knowledge contexts. Here, the word “populate” is used on purpose, meaning that entity types have for contexts the same role that entities have for entity types. In the same way as the schema of a single entity type collects entities at the instance level, the schema of a single knowledge context collects entity types at the schema level; and, in both cases, properties are what allow to discriminate among elements, i.e., entity types or entities. Based on the above intuitions, we aim to model the extent and the intent of the knowledge context. In other words, a knowledge context is a set, actually a lattice of concepts, which refers to the FCA lattice we introduced in this section.

To utilize the property information, we formalize the relation between properties and entity types/entities as \textit{associations}. Thus, two cases are considered. At the schema level, the knowledge graph schema will be flattened into a set of triples, where each triple encodes information about \textit{entity type-property} associations, e.g., triple “organization-domain-LocatedIn” encodes the “organization-LocatedIn” association. Instance-level cases generally define triples as “entity-property-entity", where two associations are encoded. For instance, triple “Eiffel Tower-Located In-Paris” encodes “Eiffel Tower-Located In" and “Paris-Located In". To generate an FCA-based formalization, we have the following settings:  

\begin{itemize}
\item Compared to general FCA lattices, we consider both entity types $E$ and entities $I$, where we associate an entity with its properties $t(I_i)$, and also an entity type with its properties $T(E_i)$;
\item We introduce the notion of \textit{undefined} to describe an additional relation between entity types and properties.
\end{itemize}

As an example, we generate an FCA lattice whose information is extracted from DBpedia \cite{auer2007dbpedia}, as shown in Figure \ref{ETG_example}. We adopt the following conventions. The value box with a cross represents the fact the property is associated with the entity type, e.g., \textit{citizenship} is associated with \textit{Person}. The value box with a circle means the property is unassociated with the entity type, e.g., \textit{date} as for \textit{Person}. The value “U” (for undefined) represents the fact that the property is unassociated with the entity type but associated with one of its sub-classes. The intuition is that the property might or might not be used to describe the current entity type, e.g., \textit{academy award} is used to describe \textit{Artist} and it might be used to describe \textit{Person}. Thus, undefined means the target entity type can be described by the specific property but not specified in the current schema. Similar to entity types, we can also find formalized entities and their properties. Note that these entities are selected from the knowledge graph with a hierarchical schema, thus, they can inherit the \textit{unassociated properties} from their entity types, e.g., as an \textit{Athlete}, \textit{Usain Bolt} does not have property \textit{duration}.

\begin{figure*}[!t]
	\centering
	\includegraphics[width=1\linewidth]{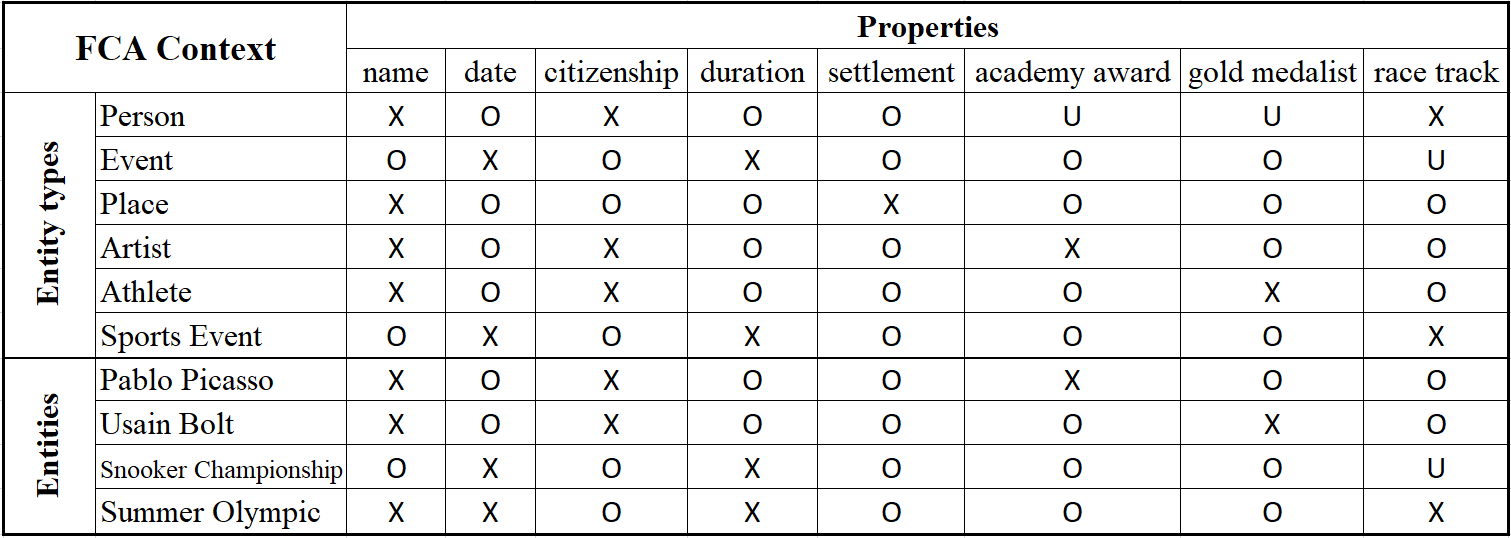}
	\caption{An example of formalizing knowledge graph into FCA contexts 
	\label{FCA_example2}}
\end{figure*}

We encode the above-mentioned three correlations as the parameter $w_E(p)$. Considering the correlation of “associated with” is positive for a property-based description, the correlation of “unassociated with” is negative and the correlation of “undefined” is neutral, we define the parameter as:
\begin{equation}
w_E(p) = \left\{
\begin{tabular}{ll}
1,  & if $p \in prop(E)$\\
0,  & if $p \notin prop(E) \land p \in prop(E.subclass)$ \\
-1, & if $p \notin prop(E) \land p \notin prop(E.subclass)$ 
\end{tabular}\right.
\label{equ:1}
\end{equation}
where we suppose $E$ is a general concept that unifies the definitions of entity (types) in a knowledge graph, $p$ is the target property, $E.subclass$ refers to the sub-classes of the entity type $E$, and $prop(E)$ refers to the properties associated with $E$. Need to notice a special case that \textit{undefined properties} also exists where a specific entity misses the inherited property, such as \textit{race track} is used to describe \textit{Sports events} but missed for its entity \textit{Snooker Championship}, which will also make $w_E(p)=0$. Thus, the crosses, “U”s and circles in Figure \ref{FCA_example2} are set to 1, 0 and -1, respectively. Please note that the proposed FCA-based formalization will be a pre-processing step for calculating the property-based similarities.

\section{The Property-based Similarity Metrics}
\begin{figure}[!t]
	\centering
	\includegraphics[width=0.7\linewidth]{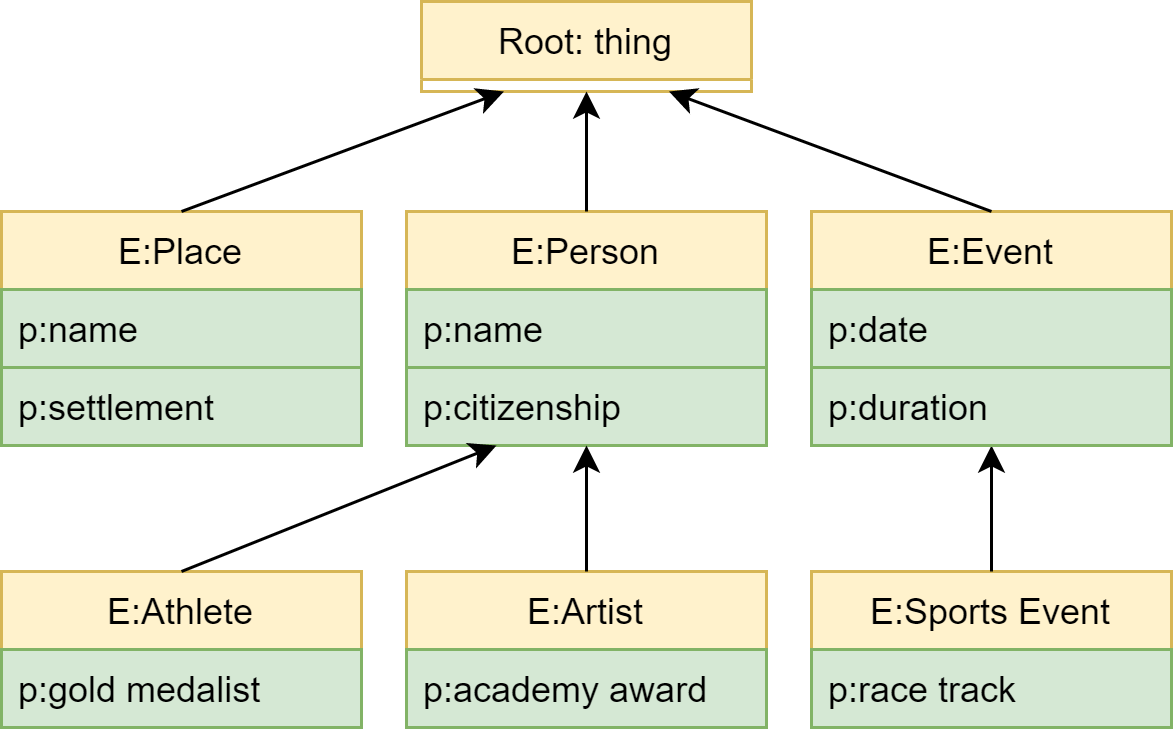}
	\caption{An example of the hierarchical schema in knowledge graph. 
	\label{ETG_example}}
\end{figure}

A foundational prerequisite of our research is the discrimination of entity types and entities via properties that are essential for delineating knowledge graph concepts \cite{giunchiglia2020entity}. The relevance can be observed between the reference and candidate entity types/entities when there is an overlap in their properties. Intrinsically, it is these properties that define an entity type, a definition that transcends the specific labels and hierarchical positioning \cite{ganter2012formal}. This perspective allows for the conceptualization of entity types within hierarchical structures, wherein subordinate entity types inherit properties from their superordinate counterparts, and the entities classified under a given type also populate all superior entity types. Consequently, this framework posits knowledge graph alignment as a task of mapping these inheritance hierarchies. Figure \ref{ETG_example} illustrates such a hierarchy, serving as a continual example throughout this chapter.

In practice, prominent knowledge graphs like DBpedia \cite{auer2007dbpedia} and OpenCyc \cite{farber2015comparative} are characterized by a profusion of properties, a consequence of the inherent objective of knowledge graph extension to broaden the spectrum of properties. Hence, quantifying the overlap in properties between entity types becomes vital, considering the rarity of complete property overlap in real-world knowledge graphs \cite{fumagalli2021ranking}. Concurrently, we engage the underlying principles of the “get-specific” heuristic, as formalized in \cite{giunchiglia2007formalizing}, to differentiate the significance of various overlapping properties. The central insight here is that properties with varying specificity levels hold disparate importance in entity type identification. Specifically, a more specific property yields greater informational content, thereby facilitating a more precise concept definition.

Accordingly, this section commences with a review of existing methodologies for measuring entity type similarity, approached through a series of comparative analyses. Subsequently, we introduce three distinct concepts: \textit{horizontal specificity}, \textit{vertical specificity}, and \textit{informational specificity}, along with their respective similarity metrics, to evaluate the extent of property overlap.

\subsection{General Similarity Measurements}

Currently, there are numerous methodologies for measuring the relevance of entity types, with the majority focusing on quantifying the semantic similarity between two entity types. This is predominantly achieved through widely utilized string-based metrics, semantic-based metrics, and approaches grounded in knowledge contexts \cite{nezhadi2011ontology}. In this section, we introduce several representative similarity measurements within each approach, serving as a baseline for measuring similarity and also for supporting the entity type recognition task.

\subsubsection{Ngram Similarity}
The introduced Ngram similarity is a string-based method used to quantify the degree of resemblance between two text sequences based on their Ngram representations. The Ngram representation \cite{kondrak2005n} is a pivotal tool in computational linguistics and natural language processing, operates on the principle of dividing text into chunks, or $n-grams$, where $n$ represents the number of contiguous characters. Essentially, Ngram similarity involves breaking down each text sequence into sets of Ngrams and then comparing these sets to determine how similar the sequences are. One common approach is to use the Dice coefficient with bigrams (DICE)\footnote{Dice coefficient with bigrams (DICE) is a particularly popular word similarity measure.}, which is defined as the size of the intersection divided by the size of the union of the sets of Ngrams. Given the two texts A and B, the N-gram similarity can be mathematically represented by: 

\begin{equation}
Sim_{ngram}(A, B)= 2 * \frac{|Ngrams(A) \cap Ngrams(B)|}{|Ngrams(A)| + |Ngrams(B)|}
\label{F(Simngram)}
\end{equation}
where $Ngrams(A)$ is a multi-set of letter n-grams in text A. Note that the range of Ngram similarity is from 0 to 1.

\subsubsection{Longest Common Sub-sequence}

The Longest Common Sub-sequence (LCS) similarity \cite{euzenat2007ontology} is a string-based measure used to assess the similarity between two texts by identifying the longest sub-sequence present in both texts. A sub-sequence is defined as a sequence that appears in the same relative order, but not necessarily contiguously. Given two texts, A and B, the LCS similarity is often calculated as the length of the longest common sub-sequence relative to the lengths of the input sequences, as: 

\begin{equation}
Sim_{LCS}(A, B)= 2 * \frac{|LCS(A,B)|}{Length(A) + Length(B)}
\label{F(SimLCS)}
\end{equation}
where the function $LCS(A, B)$ refers to the length of the longest common sub-sequence between A and B, functions $Length(A)$ and $Length(B)$ represent the length of texts A and B, respectively. This formula normalizes the length of the LCS by the total length of the two texts, giving a value between 0 and 1.

\subsubsection{Levenshtein Distance}
Levenshtein distance, also known as edit distance \cite{yujian2007normalized}, is a string-based measurement of the difference between two texts. It is defined as the minimum number of single-character edits (insertions, deletions, or substitutions) required to change one word into the other. When using the Levenshtein distance as a basis for calculating the similarity between two texts, the formula typically involves normalizing the distance by the length of the longer string, thereby converting the distance into a similarity measure that ranges from 0 (completely dissimilar) to 1 (identical). The formula is as follows:

\begin{equation}
Sim_{LD}(A, B)= 1 - \frac{Levenshtein Distance(A,B)}{max(|Length(A)|,|Length(B)|)}
\label{F(SimLD)}
\end{equation}
where the function $LevenshteinDistance(A, B)$ refers to the minimum number of single-character edits required to transform text A to B, and the function $max(\cdot)$ returns the length of the longer text.

\subsubsection{WordNet Similarity}
WordNet similarity refers to a group of methods used to compute the semantic similarity between words or concepts based on the WordNet database \cite{richardson1994using}. WordNet organizes English words into sets of synonyms called synsets and records various semantic relations between these synsets, including is-a relationships, subset relationships, and part-of relationships. Different WordNet similarity measures exploit these relationships to assess the degree of semantic similarity or relatedness between words. Here we introduce a measurement namely Wu-Palmer Similarity \cite{palmer1994verb} based on information content. The similarity exploits the least common subsumer, which is the most specific ancestor node common to both synsets in the WordNet hierarchy. Thus, given two texts A and B, we have:

\begin{equation}
Sim_{WordNet}(A, B)= 2 * \frac{LC_{subsumer}(A,B)}{Depth(A) + Depth(B)}
\label{F(SimWordNet)}
\end{equation}
where the function $LC_{subsumer}(A, B)$ is the depth of the lowest common ancestor of A and B, and function $Depth(\cdot)$ refers to the depths of synsets in the taxonomy. Note that the texts A and B need to be found as synsets in WordNet.

\subsubsection{Word2Vec Similarity}
Word2Vec similarity is a concept derived from the Word2Vec model that produces word embeddings \cite{church2017word2vec}. Word2Vec model is a shallow, two-layer neural network trained to reconstruct linguistic contexts of words. The similarity between two words in the Word2Vec model is often calculated using the cosine similarity between their vector representations. Cosine similarity measures the cosine of the angle between two non-zero vectors of an inner product space, which in this case are the Word2Vec vectors of the two words. Given the texts A and B, we have:

\begin{equation}
Sim_{Word2Vec}(A, B)=  \frac{Vec(A) \cdot Vec(B) }{||Vec(A)|| * ||Vec(B)||}
\label{F(SimWord2Vec)}
\end{equation}
where $Vec(\cdot)$ represents the vector of the text produced by the Word2Vec model, $Vec(A) \cdot Vec(B)$  is the dot product of two vectors, and $||Vec(A)|| * ||Vec(B)||$ are the Euclidean norms of vectors. The Word2Vec similarity ranges from -1 meaning exactly opposite, to 1 meaning the same, with 0 typically indicating independence.

\subsection{Horizontal Similarity}
For measuring the specificity of a property, a possible idea is to horizontally compare the number of entity types that are described by a specific property, namely the shareability of the property \cite{giunchiglia2020entity}. If a property is used to describe diverse entity types, it means that the property is not highly characterizing its associated entity types. Thus, for instance, in figure \ref{FCA_example2}, the property \textit{name} is used to describe \textit{Person}, \textit{Place}, \textit{Athlete} and \textit{Artist}, where \textit{name} is a common property that appears in different contexts. Dually, \textit{settlement} is a horizontally highly specific property since it is associated only with the entity type \textit{Place}. Based on this intuition, we consider the specificity of a property as related to its shareability. Therefore, we propose $HS$ (\textit{Horizontal Specificity}) for measuring property specificity. More precisely, $HS$  aims to measure the number of entity types that are associated with the target property in a specific knowledge graph. We model $HS$ as: 

\begin{equation}
HS_{KG}(E,p)= w_E(p) * {e^{\lambda(1-|K_v|)}} \in [-1, 1]
\label{F(HS)}
\end{equation}
where $p$ is the input property; $K_v$ is the set of entity types described by the input property in a specific knowledge graph and $|K_v|$ is the number of entity types in $K_v$, thus $|K_v| \ge 1$; $e$ denotes the natural mathematical constant \cite{finch2003mathematical}; $\lambda$ represents a constraint factor. The reason for using an exponential function to model the $HS$ is that we aim to normalize the horizontal specificity. The motivation is that different properties may have a larger difference on $|K_v|$ in large knowledge graphs.

We have modeled the horizontal specificity of properties. Then, we define horizontal similarity metrics based on the corresponding specificity to measure the property overlapping between two concepts. Given two knowledge graphs, the reference knowledge graph $A$ and candidate knowledge graph $B$, we model the horizontal similarity $Sim_H$ between entity types/entities from $A$ and $B$: 

\begin{equation}
Sim_H(E_a,E_b) =\frac{1}{2}\sum_{i=1}^{k} \left ( \frac{HS_{A}(E_a, p_i)}{|prop(E_a)|} + \frac{HS_{B}(E_b, p_i)}{|prop(E_b)|} \right) \in [0, 1]
\label{F(SimH)}
\end{equation}

\noindent where we take $E_a$, $E_b$ as the input entity types from $A$ and $B$ respectively, thus $E_a \in A$ and $E_b \in B$; $prop(E)$ refers to the properties associated with the specific entity type and $|prop(E)|$ is the number of properties in $prop(E)$; $HS_{A}(E_a,p_i)$ and $HS_{B}(E_b,p_i)$ refer to the horizontal specificity of the aligned property $p_i$ in $A$ and $B$ respectively; $k$ is the number of aligned properties which are associated with both entity type $E_a$ and $E_b$. As a result, we obtain the horizontal similarity metric $Sim_H$. Note that each similarity metric is symmetric, more specifically, $Sim_H(E_a,E_b) = Sim_H(E_b,E_a)$. Note also that we apply z-score normalization \cite{ali2014data} to similarity metrics at the end of calculations, and the range of $Sim_H$ is between $0$ to $1$. 

\subsection{Vertical Similarity }
Entity types are organized into classification hierarchies such that the upper-layer entity types represent more abstract or more general concepts, whereas the lower-layer entity types represent more concrete or more specific concepts \cite{giunchiglia2007formalizing,rios2013learning}. Correspondingly, properties of upper-layer entity types are more general since they are used to describe general concepts, vice versa, properties of lower-layer entity types are more specific since they are used to describe specific concepts. We assume that lower-layer properties will contribute more to the identification of an entity type since they are more specific. For instance, in Figure \ref{FCA_example2}, as a lower-layer entity type, \textit{Artist} can be identified by the property \textit{academy award} but not by the property \textit{name}. Based on this intuition, we propose $VS$ for capturing the vertical specificity, as follows:

\begin{equation}
VS_{KG}(E,p)= w_E(p) * \theta * \min_{E \in K_v} layer(E) \in [-1, 1]
\label{F(VS)}
\end{equation}
where $\theta$ is a constraint factor which normalized the range of the function; where $layer(E)$ refers to the layer of the inheritance hierarchy where an entity type $E$ is defined. Note that all $E$ in set $K_v$ are described by $p$. Then, we define the vertical entity type similarity metric $Sim_V$ as from below:

\begin{equation}
Sim_V(E_a,E_b) =\frac{1}{2}\sum_{i=1}^{k} \left ( \frac{VS_{A}(E_a,p_i)}{|prop(E_a)|} + \frac{VS_{B}(E_b,p_i)}{|prop(E_b)|} \right) \in [0, 1]
\label{F(SimV)}
\end{equation}

\noindent Similar to the definition of $Sim_H$, we have candidate entity types  $E_a \in A$ and $E_b \in B$; and the properties $prop(E)$ associated with the entity type $E$. The key difference is that $Sim_V$ exploits the property specificity based on the layer information $VS_{KG}$. $VS_{A}(E_a, p_i)$ and $VS_{B}(E_b, p_i)$ refer to the highest layer of the aligned property $p_i$ in $A$ and $B$, respectively. Notice that $Sim_V$ is symmetric as well. $Sim_V$ is also normalized by z-score normalization, in the same way as  $Sim_H$. Finally, the range of $Sim_V$ is between $0$ to $1$.

\subsection{Informational Similarity}
Horizontal specificity allows measuring the shareability of properties. Notice that $HS$ is independent and does not change (increase or decrease) with the number of entities populating it.  We take into account this fact by introducing the notion of informational specificity $IS$. The intuition is that $IS$ will decrease when the entity counting increases. Thus, for instance, the $IS$ of \textit{gold medalist} decreases when there are increasing entities of \textit{athletes}, as from the schema in Figure \ref{FCA_example2}. Clearly, $IS$, differently from $HS$, can be used in the presence of entities.

The definition of informational specificity is inspired by Kullback–Leibler divergence theory \cite{van2014renyi}, which is introduced to measure the difference between two sample distributions $Y$ and $\hat{Y}$. More specifically, given a known sample distribution $Y$, assume that a new coming attribute $x$ changes $Y$ to $\hat{Y}$. Then Kullback–Leibler divergence theory says that the importance of $x$ for defining $Y$ is positively related to the difference between $Y$ and $\hat{Y}$. In the definition of informational specificity, we need to exploit some notions from information theory. We define the informational entropy of the knowledge graph: 
\begin{equation}
H(K)= \frac{- \sum_{i=1}^{|K|} F(E_i)\log\frac{F(E_i)}{|K|}}{|K|}
\label{F(entropy)}
\end{equation}
where $K$ refers to the set of all entity types in an knowledge graph and $|K|$ is the cardinality of $K$; $H(\cdot)$ represents the informational entropy of an entity type set; $E_i$ is a specific entity type in set $K$, thus $|K|\ge i \ge 1$; $F(E_i)$ refers to the number of samples of entity type $E_i$.  Need to notice when we calculate the informational entropy $H(K)$ for knowledge graphs without entities, $F(E_i) = 1$ since each knowledge graph includes one entity type instance. For a knowledge graph with entities, $F(E_i)$ depends on the number of instances of the given entity type. After obtaining informational entropy, the informational specificity $IS$  of property for describing a set of entity types $K$ is defined as:
\begin{equation}
IS_{KG}(E,p)= w_E(p) * (H(K) - \sum \frac{|K_v|}{|K|} H(K_v)) \in [-1, 1]
\label{F(IS)}
\end{equation}
where $K$ is the set of all entity types in $KG$; $K_v$ is the subset of $K$ described by the input property $p$. We weight each informational entropy $H(K_v)$ by the proportion of $|K_v|$ to $|K|$. Being subtracted by the overall informational entropy $H(K)$, $IS$ presents the importance of the property $p$ for describing the given entity type set $K$. Then, the informational similarity $Sim_I$ can be calculated by $IS$, as:

\begin{equation}
Sim_I(E_a,E_b) =\frac{1}{2}\sum_{i=1}^{k} \left ( \frac{IS_{A}(E_a,p_i)}{|prop(E_a)|} + \frac{IS_{B}(E_b,p_i)}{|prop(E_b)|} \right) \in [0, 1]
\label{F(SimI)}
\end{equation}
Analogous to the former similarities, our approach involves candidate entity types $E_a$ and $E_b$, along with the properties $prop(E)$ associated with the entity type $E$. The primary distinction lies in that $Sim_I$ leverages the informational entropy to distinguish the weights of different aligned properties. It is noteworthy that $Sim_I$ exhibits symmetry. Additionally, $Sim_I$ also undergoes normalization through a z-score process. Consequently, the value range of $Sim_I$ spans from $0$ to $1$.

\section{Qualitative Evaluation}
We conduct experiments on several real-world datasets used for entity type recognition. Our approach focuses on knowledge graphs that contain entity types associated with a fair number of properties. For the evaluation of \textit{entity type-entity type} alignments $<E_i, E_j>$ and \textit{entity type-entity} alignments $<E_i, I_k>$, we involve ConfTrack $ra1$\footnote{https://owl.vse.cz/ontofarm/} version \cite{zamazal2017ten} 
from Ontology Alignment Evaluation Initiative\footnote{http://oaei.ontologymatching.org/2021/} (OAEI) tracks which are the main references for most of the schema aligning methods, where dataset CONF contains 21 annotated reference graph pairs.

\subsubsection{Schema-level Evaluation}
Table \ref{Schema-level Explicit Evaluation} provides representative examples to show the entity type similarity metrics between candidate entity type $E_{cand}$ with reference entity type $E_{ref}$ pairs in schema-level knowledge graphs. Note that the former four results are from \textit{cmt-confof} and the latter are from \textit{cmt-conference} in ConfTrack. Value box “M" demonstrates if two entity types refer to the same concept, where $\times$ refers to a positive answer. We also introduce two representative similarity metrics for string-based and semantic-based measurements as the baseline, i.e., Levenshtein Distance $Sim_{LD}$ and Word2Vec Similarity $Sim_{Word2Vec}$.

\begin{table}[!t]
\centering
\caption{Representative samples of property-based similarity $Sim_V$, $Sim_H$ and $Sim_I$ on entity type-entity type pairs.}
  \label{Schema-level Explicit Evaluation}

\begin{tabular}{|c|c|c|c|c|c|}
\hline

$E_{ref}$   & $E_{cand}$   & $Sim_V$ & $Sim_H$ & $Sim_I$ & M\\ \hline
Paper       & Contribution       & 1      & 0.853  & 0.730 & $\times$ \\ \hline
SubjectArea & Topic             & 0.756  & 0.740  & 0.857 & $\times$ \\ \hline
Author & Topic         & 0.198  & 0.353  & 0.018 & \\ \hline
Meta-Review & Poster          & 0      & 0.312  & 0.262 & \\ \hline
Chairman    & Chair           & 1      & 0.559  & 0.554 & $\times$ \\ \hline
Person      & Person         & 1      & 0.970  & 0.678 & $\times$ \\ \hline
Person      & Document       & 0.02   & 0.06   & 0 & \\ \hline
Chairman    & Publisher      & 0      & 0.07   & 0.195 & \\ \hline
\end{tabular}

\end{table}





We find that our property-based similarity metrics indeed capture the contextual similarity between relevant entity types, where aligned entity types output higher values (e.g., \textit{Paper-Contribution}), in turn, non-aligned entity types return lower values (e.g., \textit{Person-Document}). Meanwhile, the introduced metrics $Sim_{LD}$ and $Sim_{Word2Vec}$ demonstrate the same trend in the general examples, e.g., \textit{SubjectArea-Topic}, and \textit{Chairman-Chair}. However, the string-based Levenshtein Distance metric does not work well for measuring entity type pairs that have very different text descriptions, e.g., \textit{Paper-Contribution}. Our proposed similarities can distinguish such entity type pairs since the similarity values are calculated based on the properties. With an explicit observation of the metric values, we consider the property-based similarity metrics $Sim_H$, $Sim_V$, and $Sim_I$ valid for \textit{entity type-entity type} pairs.

\subsubsection{Instance-level Evaluation}

Table \ref{Instance-level Explicit Evaluation} delineates a selection of similarity metric examples, drawn to compare reference entity types with candidate entities from the dataset. Echoing the settings in Table \ref{Schema-level Explicit Evaluation}, the ``M" notation within a value box, marked with a $\times$, denotes the candidate entity belongs to a specific reference entity type. Notably, in this experimental setup, we have excluded both string-based and semantic-based similarity measures. This decision stems from the observation that textual descriptions of entities and their corresponding entity types often do not exhibit direct string or semantic alignments\footnote{This is also one of the reasons that current NLP-based and structure-based measurements poorly perform the instance-level entity type recognition task.}. 

\begin{table}[!t]
\centering
\caption{Representative samples of $Sim_V$, $Sim_H$ and $Sim_I$ on entity type-entity pairs.}
  \label{Instance-level Explicit Evaluation}

\begin{tabular}{|c|c|c|c|c|c|}
\hline
$E_{ref}$   & $I_{cand}$        & $Sim_V$ & $Sim_H$ & $Sim_I$ & M \\ \hline
Person       & MiltHinton       & 1      & 0.787  & 0.873 & $\times$ \\ \hline
Person      & Jadakiss          & 1      & 0.645  & 0.305 & $\times$ \\ \hline
Person      & Boston            & 0.264  & 0.173  & 0.041 & \\ \hline
Place       & Boston            & 0.720  & 1      & 0.433 & $\times$ \\ \hline
Place    & Jadakiss             & 0.128  & 0.093  & 0.148 & \\ \hline

Organization & MiltHinton       & 0.070  & 0.022  & 0.092 & \\ \hline

\end{tabular}
\end{table}

Similar to what happened in schema-level entity type pairs, we find the similarity metrics of aligned instance-level pairs return much higher values than non-aligned pairs, which proves the effectiveness of the proposed property-based similarities for demonstrating contextual similarities between instance-level pairs. Thus, we conclude the proposed similarity metrics are valid for measuring both schema-level and instance-level cases.

\section{Summary}

In the current chapter, a suite of innovative entity type similarity metrics has been introduced, tailored for tasks involving entity type recognition. Initially, we dive into the notion that properties implicitly delineate entity types, offering a fresh perspective for their identification. This concept is further demonstrated through the introduction of the `knowledge lotus', a visual tool designed to graphically represent these statistics. Additionally, we outline a modeling strategy for knowledge graphs, organizing the arrangement of a knowledge graph and its intrinsic associations between entity types/entities and properties, utilizing the framework of FCA lattices.

Subsequently, we propose three novel metrics to quantify the contextual similarity between reference entity types and their candidate counterparts: horizontal similarity (\(Sim_H\)), vertical similarity (\(Sim_V\)), and informational similarity (\(Sim_I\)). Within the same discourse, we undertake a comparative analysis of various existing entity type similarity metrics, primarily those employing string-based and semantic-based methodologies, establishing them as baselines for comparing with our proposed metrics.

The validation of our metrics encompasses both the schema and instance levels, with empirical results underscoring the efficacy and reliability of our similarity metrics from a qualitative standpoint.

\chapter{Entity Type Recognition}

\section{Introduction}
In this chapter, we propose a machine learning-based framework to achieve automatic entity type recognition, aiming to conduct knowledge extraction on both schema-level and entity-level knowledge graphs. This study is based on the work presented in \cite{shi2023recognizing, shi2022simple, wu2023knowlab, li2023fcc}.

The challenge of semantic heterogeneity emerges prominently in the context of extending knowledge graphs by varied sources \cite{lonsdale2010reusing}. A well-known strategy to address this issue involves conducting \textit{entity type recognition} by leveraging the knowledge encoded within one or several reference knowledge graphs. The literature delineates two primary methodologies in this regard. The first approach, schema-level entity type recognition, is oriented toward aligning a collection of candidate entity types with a corresponding set of reference entity types, utilizing schema-level data. This approach is typically seen in the realm of schema alignment, as exemplified by studies like \cite{euzenat2007ontology,giunchiglia2004s,algergawy2018results}. The second approach, instance-level entity type recognition, focuses on the information intrinsic to individual entities, as illustrated in research such as \cite{portisch2021background,shalaby2016entity}. Both approaches play a pivotal role in data-intensive applications, notably in fields like data integration and coreference resolution \cite{lenzerini2002data,lee2017end}.

The entity type recognition task is primarily concerned with the accurate identification and categorization of entities within a given text or dataset into predefined types. It is fundamental in structuring unstructured data by assigning relevant types to entities, thus facilitating enhanced data retrieval, organization, and understanding. Specifically, we focus on the case of knowledge graph extension, which requires knowledge extraction by recognizing the type of entities or entity types from additional candidate knowledge graphs following the entity types given by the reference knowledge graph. Thus, this task necessitates the analysis of contextual and semantic attributes of entities, leveraging methodologies and insights from NLP and ontology. Considering the different data levels of knowledge graphs, in the chapter, we aim to conduct:
\begin{itemize}
\item Schema-level entity type recognition: Align the given candidate entity types with the pre-set reference entity types, e.g., \textit{Chairman-Chair};
\item Instance-level entity type recognition: Align the candidate entities with the pre-set reference entity types, e.g., \textit{Usain Bolt-Athlete}.
\end{itemize}

In the field of entity type recognition, existing methods employ various metrics as primary features to assess possible alignments. Rule-based methods were introduced in the past decade, exploiting string-based similarity metrics as the core measurement for aligning lexical-related entity type pairs \cite{cheatham2013string}. Then, researchers considered the contextual relevance and encoded knowledge graph structures into the similarity metrics, e.g., s-match \cite{giunchiglia2012s}. To facilitate a more comprehensive assessment of the entity type relevance, an increasingly diverse range of similarity metrics was employed. The main intuition of these studies is to predict two entity types align each other when the value of the similarity metrics is higher than a pre-set threshold. However, rule-based methods perform limitedly since most of them are dedicated methods for using one or a few metrics with artificial pre-set thresholds. Meanwhile, such methods grounded in predefined heuristics and explicit rules, exhibit a rigidity that limits their ability to generalize beyond the pre-set scenarios.

To solve the above-mentioned problems, researchers attempt to exploit learnable methods, i.e., machine learning-based methods, which uncover subtle patterns and relationships within the data during entity type recognition. This learning process enables machine learning models to adapt to new trends and variances in the data, leading to more robust and accurate predictions, e.g., work presented in \cite{bulygin2019applying}. Furthermore, machine learning models can handle high-dimensional data and nonlinear relationships more effectively than rule-based systems. Thus, in contexts requiring adaptability and handling of complex data structures, e.g., instance-level entity type recognition, machine learning-based methods inherently outperform rule-based counterparts, demonstrating superior predictive performance and versatility.

As a result, we introduce a novel entity type recognition framework that exploits machine learning algorithms, aiming to improve the model predictions and adaptability. Meanwhile, the proposed framework deploys property-based similarity metrics, together with various string-based and structure-based similarity metrics, to comprehensively assess candidate entity types for better recognition performance. Thus, we demonstrate the whole recognition pipeline and details of each framework component. We also have a comprehensive review of machine learning models that are expected to be applied in the framework, where we have an experimental comparison to determine the model selection. Lastly, we conduct several experiments to demonstrate the effectiveness of the proposed entity type recognition framework, the selection of the machine learning model, and the proposed property-based similarity metrics.

\section{Entity Type Recognition Framework}
\begin{figure*}[!t]
	\centering
	\includegraphics[width=0.62\linewidth]{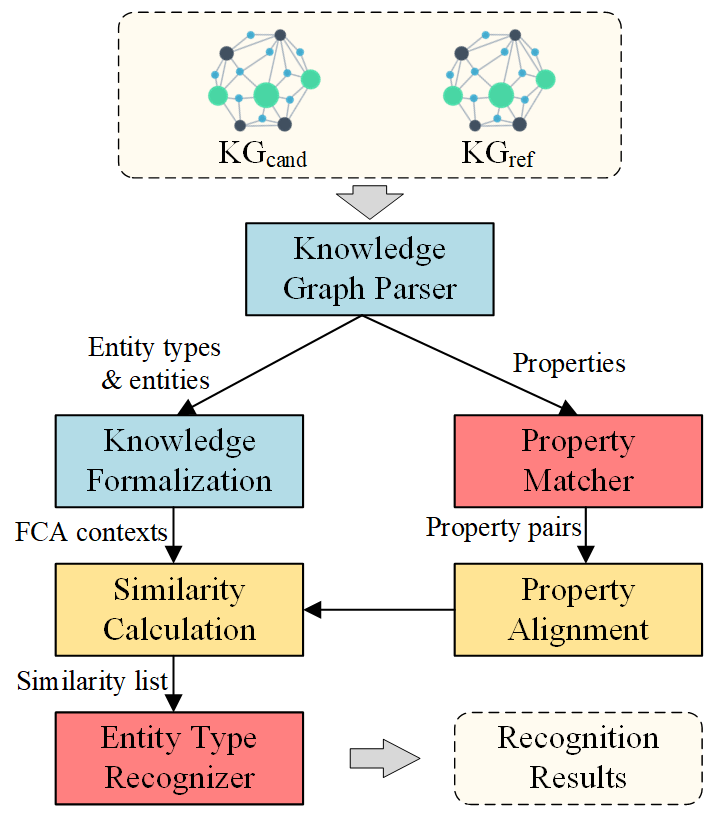}
	\caption{The proposed framework for entity type recognition. 
	\label{framework}}
\end{figure*}

In order to predict the entity types of unknown concepts from candidate knowledge graphs, we propose a machine learning-based framework that exploits the similarity metrics defined above, as shown in Figure \ref{framework}. The framework mainly consists of five modules, namely knowledge graph parser, knowledge formalization module, property matcher, similarity calculation module, and entity type recognizer. Notice that modules are marked in different colors to distinguish their usage. 

Different from knowledge graph completion tasks \cite{zhao2020connecting,yaghoobzadeh2017multi} which manipulate entities in a single knowledge graph, there will be two (or more) knowledge graphs involved in the entity type recognition task, where we have $KG_{ref}$ and $KG_{cand}$ as inputs, as a step during knowledge graph extension. The knowledge graph parser aims to parse the input knowledge graph as a structured set of entity types, entities, and properties. The knowledge graph will be flattened into a set of triples, where associations will be extracted from triples. Then the triples and associations will be used to generate an FCA for each input knowledge graph by the knowledge formalization module. Note that the knowledge graph parser and the formalization module are marked in blue since they are data pre-processing modules in the framework. 

After all properties from the input knowledge graphs have been collected, they will be sent to the NLP-based property matcher. Different labels of properties may express similar meanings since many of them are variations of the same label. Thus, an NLP pipeline is designed to normalize all input properties, where phrase segmentation, lemmatization, and stop-word removal are introduced for better normalization performance. String-based and semantic-based similarity metrics are exploited for aligning the properties by normalized labels \cite{bella2017language,bella2016domain}. Then, we align properties from input knowledge graphs following the results of the property matcher.

In the next phase, we generate the proposed property-based similarity metrics $Sim_H$,  $Sim_V$, and $Sim_I$ by inputting the FCA contexts and aligned properties. According to the functions presented in the last chapter, three similarity values will be generated for each entity type pair, which will then be passed to the entity type matcher together with general string-based and semantic-based similarity metrics. We will map reference entity types with candidate entity types and entities by the output of the entity type recognizer. Note that two ML-based modules, including the property matcher and entity type recognizer, are both highlighted in red. The final output of the framework is a set of \textit{entity-entity type} and \textit{entity type-entity type} aligning pairs, which represent the successful recognition results.

\subsection{Similarity Calculation Algorithm}

\begin{algorithm}[!t] 
\caption{Calculating horizontal similarity $Sim_H$ between reference and candidate knowledge graphs. $L_H=SimCal_H(f_a,f_b)$} 
\label{Similarity} 
\begin{algorithmic}[1] 
\REQUIRE ~~\\ 
Reference and candidate FCA contexts $f_a$, $f_b$;\\
\ENSURE ~~\\ 
List of all horizontal similarities $L_H$;\\

\STATE 	$PM = (p_a \times p_b) = PropertyMatcher(f_a,f_b)$; \{$PM$ is formed as a set of aligned property pairs, where $p_a \in f_a$, $p_b \in f_b$.\}
\STATE $EM = (E_a \times E_b) = EtypeSelector(f_a,f_b)$; \{entity types $E_a, E_b$ from $f_a, f_b$ are assembled as candidate calculation pairs $EM$.\}
\FOR {all $(E_j,E_k) \in EM$}
\STATE $Sim_H(E_j,E_k) = 0$; \{initialize the value of horizontal similarity $Sim_H(E_j,E_k)$.\}
\FOR {all $(p_n,p_m) \in PM$}
\IF {$p_n \in prop(E_j) \wedge p_m \in prop(E_k)$}
\STATE $Sim_H(E_j,E_k).add(\frac{HS_A(E_j,p_n)}{|prop(E_j)|} {+} \frac{HS_B(E_k,p_m)}{|prop(E_k)|})$; \{calculate the corresponding specificity $HS$ for the similarity $sim_H(E_j,E_k)$.\}

\ENDIF
\ENDFOR
\STATE $L_H.stack(\frac{1}{2} * Sim_H(E_j,E_k))$; 
\{stack the value of horizontal similarity to the list $L_H$.\}

\ENDFOR

\RETURN $L_H$
\end{algorithmic}
\end{algorithm}

The property-based similarity calculation is one of the critical parts of this study. We detail the calculation, as shown in Algorithm \ref{Similarity}. After formalizing reference and candidate knowledge graph $KG_{ref}$ and $KG_{cand}$, we assume that the two FCA contexts $f_a$ and $f_b$ are generated correspondingly. Then we obtain property aligning pairs $PM$ from the property matcher. To calculate the similarity of entity types, we need to generate candidate entity type pairs $EM$ for further processing. For each candidate pair in $EM$, we check their correlated properties and update the specificity values to $Sim_H$,  $Sim_V$, and $Sim_I$ when their properties are aligned. After traversing all the candidate pairs, we obtain a complete entity type similarity list $L$ which will be used for training the ML model and aligning candidate entity types. Notice that we present the algorithm for calculating the horizontal similarity $Sim_H$ in function \ref{F(SimH)}, the metrics vertical similarity $Sim_V$ and informational similarity $Sim_I$ will be calculated by the corresponding functions.

\subsection{Entity Type Recognizer}
The entity type recognizer introduced in Figure \ref{framework} is developed by learnable algorithms, where two critical cases are considered for knowledge graph extension. Here we present more details of training and setting the recognizer. 

\subsubsection{Schema-level Recognition: Aligning Entity Types}
We develop a machine learning-based method that deals with entity type aligning as a binary classification task. The main idea is to predict if two incoming entity types are aligned with each other, output true or false. For applying this method, a list of candidate pairs is generated by pairing entity types from $KG_{cand}$ and $KG_{ref}$. We will record candidate entity type pairs $EM_{ali}$ when the result of classification is true. The proposed property-based similarity metrics $Sim_H$,  $Sim_V$, and $Sim_I$ are introduced to train the machine learning models for aligning entity types. For better performance, we also exploit string-based and semantic-based similarity metrics, along with property-based similarity metrics for training the model for alignment.

\subsubsection{Instance-level Recognition: Aligning Entities with Entity Types}
The strategy for using the recognizer to align entities with entity types is very similar to the above case, where we predict if candidate entities match with target entity types, and output true or false. Thus, we will also generate candidate pairs that consist of entities from $KG_{cand}$ and entity types from $KG_{ref}$. The true candidate pairs will be recorded and the corresponding entity types will be output as the final recognition results. Compared to the entity types alignment cases, the main difference is that the aligning case mainly uses property-based similarity metrics as features for model training since the lexical labels are not applicable to align entities and entity types. Thus, property-based similarity metrics $Sim_H$,  $Sim_V$, and $Sim_I$ are applied.

\subsection{Property Matcher}
The objective of the property matcher is to facilitate the alignment of properties across knowledge graphs, employing both string-based and semantic-based similarity metrics for model training. The aligning strategy is the same as the entity type recognizer, where the candidate property pairs are generated to predict the true or false as the aligning results. The efficacy of a property matcher is crucial, as the outcomes in both scenarios within the entity type recognizer are contingent upon the results yielded by the property matcher. This section demonstrates various solutions we applied to mitigate the impact of misaligned properties, thereby enhancing the accuracy and reliability of the property aligning process.

\begin{itemize}
    \item Use of the \textbf{formalization parameter $w_E(p)$}. As we introduced in Section 4.3, besides “associated” (positive) and “unassociated” (negative) properties, we also defined “undefined” properties (neutral). Since misaligned properties will not be used for similarity calculation, they are treated as “undefined” properties which will not affect the model training and reduce the additional interference. However, additional interference from misaligned properties appears if “unassociated” and “undefined” properties are not distinguished. 
    
    \item Use of \textbf{similarity metrics}. Similar to lexical-based similarity metrics, our property-based similarity metrics also allow us to align entity types by soft matching, even if there are few properties not aligned. This will increase the robustness of our entity type recognition approach. 

    \item Use of \textbf{learnable models}. By learning from the practical data from different resources, machine learning models will propose a learnable strategy rather than a fixed threshold for determining alignments, which will maximize the use of existing aligned properties and minimize the effect of misaligned properties. 

\end{itemize}

\section{Algorithm Selection}
In this section, we endeavor to explore recent developments of machine learning algorithms, to identify reasonable algorithms for applying the entity type recognizer. Our focus is centered predominantly on two types of methods: classic machine learning algorithms and neural networks. Here we investigate several algorithms as follows:

\begin{enumerate}

    \item \textbf{Logistic Regression}: Logistic regression (LR) \cite{ng2001discriminative} is a statistical model that uses a logistic function to model a binary dependent variable. In regression analysis, logistic regression estimates the parameters of a logistic model; it is a form of binomial regression.

    \item \textbf{SGDClassifier}: Stochastic gradient descent (SGD) \cite{kabir2015bangla} classifier is a linear classifier (such as SVM or logistic regression) under the SGD training framework. It is especially useful for large-scale and sparse machine learning problems often encountered in text classification and natural language processing. This classifier iteratively minimizes a given loss function using a stochastic gradient descent method.

    \item \textbf{Decision Tree}: A decision tree (DT) \cite{safavian1991survey} is a non-parametric supervised learning method used for classification and regression. The goal is to create a model that predicts the value of a target variable by learning simple decision rules inferred from the data features. The algorithm is simple to understand and interpret but can be prone to overfitting.

    \item \textbf{Random Forest}: The random forest (RF) \cite{pal2005random} algorithm is an ensemble learning method, primarily used for classification and regression tasks. It operates by constructing a multitude of decision trees during training and outputs the mode of the classes (classification) or mean prediction (regression) of the individual trees. Compared with decision trees, the random forest algorithm is less susceptible to the issue of overfitting.

    \item \textbf{XGBoost}: XGBoost \cite{chen2016xgboost} stands for eXtreme gradient boosting, which is also an ensemble learning method. It is an implementation of gradient-boosted decision trees designed for speed and performance. XGBoost provides a parallel tree boosting that solves many data science problems in a fast and accurate way.

    \item \textbf{ANNClassifier}: An artificial neural network (ANN) classifier \cite{nath2021automated} is a computational model that is distinguished by its adeptness in modeling intricate information. It consists of layers of interconnected nodes (neurons), where each connection represents a weight that is adjusted during the learning process. ANN classifier is versatile and can model complex nonlinear relationships in data, which is capable of handling tasks with high-dimensional data, like image and speech recognition.
\end{enumerate}

Classical machine learning algorithms demonstrate efficacy in tabular data learning, excelling in feature analysis. Algorithms with straightforward structures, such as logistic regression or decision trees, are notable for their interpretability and proficiency in processing datasets characterized by linear boundaries. Nevertheless, these algorithms are inclined towards overfitting and exhibit limitations in managing noisy data. Both SGDClassifier and ensemble-based algorithms are adept at processing non-linear data and larger datasets. Specifically, the SGDClassifier and ANNClassifier are well-suited for high-dimensional data, while XGBoost is distinguished for its high performance and speed. 

In our prior research, we explored various feasible configurations for neural network classifiers, distinguished by their network structure's scale. In our work \cite{wu2023knowlab}, we developed a large language model employing a bidirectional transformer specifically for clinical report extraction tasks. Additionally, in \cite{shi2022simple}, we utilized a pre-trained language model, founded on contrastive learning, to effectively address long text pairing tasks. Meanwhile, we have investigated the impact of data scale and network structure scale on the performance of classifiers \cite{li2023fcc}. Our experiments have consistently indicated that neural network models do not invariably exhibit enhanced performance in classification tasks as the complexity of their network architecture increases. This is particularly evident in classification tasks that utilize tabular data as the training set, where complex neural networks are prone to overfitting, likely due to the encapsulation of manually extracted features within tabular data. 

Therefore, considering our training data already includes features that have been artificially engineered, we have shown a preference for neural networks of moderate size. Pursuing this approach, in our study \cite{shi2023recognizing}, we introduced a shallow artificial neural network — a classification model built through a five-layer fully connected Sigmoid architecture. The rationale behind this selection was driven by the specific characteristics of our training data, which had undergone feature engineering, necessitating the use of a neural network that is optimally scaled to suit our requirements.

For the algorithm selected to apply in entity type recognizer, we plan to conduct evaluations of all the above-mentioned algorithms in schema-level experimental settings, given that the volume of training data for schema-level recognition is generally moderate and not excessively large.  For instance-level recognition, we plan to conduct the algorithms that can deal with large datasets, i.e., SGDClassifier, random forest, XGBoost, and ANNClassifier.

\section{Model Training}

In this section, we introduce the model training strategies and technical details, aiming to achieve advanced entity type recognition performance. 

\subsection{Feature Selection}
Our approach models the recognition task by a generic binary classification\footnote{The objective of binary classification is to map an input feature vector $x \in \mathbb{R}^{n}$, to a discrete label, $y \in \{0,1\}$.} strategy. Distinct from dedicated techniques, this approach maintains independence from the specific selection of machine learning models. Binary classification tasks necessitate annotated datasets with both positive and negative labels to facilitate prediction output. Here, the labeling of data as positive or negative samples corresponds to the alignment status of either entity type-entity type pairs or entity type-entity pairs.

In addition to our property-based similarity metrics, we have elected to incorporate a selection of the most prevalent string-based and semantic-based similarity metrics. This integration is undertaken to enhance the overall performance in entity type recognition, leveraging these additional metrics to provide a more robust and effective classification. We conduct three kinds of data features for model training, including:
\begin{itemize}
    \item String-based similarity metrics: N-gram similarity \cite{kondrak2005n}, longest common sub-sequence similarity \cite{euzenat2007ontology}, Levenshtein distance \cite{yujian2007normalized}, substring similarity \cite{euzenat2007ontology}, and Needleman-Wunsch distance \cite{cohen2003comparison}; 
    \item Semantic-based similarity metrics: Wu and Palmer similarity \cite{palmer1994verb}, Word2Vec \cite{church2017word2vec}, and transformer embedding similarity \cite{laskar2020contextualized};
    \item Property-based similarity metrics: horizontal similarity $Sim_H$, vertical similarity $Sim_V$, and informational similarity $Sim_I$.
\end{itemize}
These similarity metrics aim to measure different aspects of the relevance between the reference entity type and candidates. Since all the above-mentioned similarity metrics are symmetric, the order of entity type/entity in the candidate pair will not affect the final results. Moreover, we apply only property-based similarity metrics ($Sim_H$, $Sim_V$ and $Sim_I$) for instance-level entity type recognition, since the label of entity is commonly not relevant to its entity type.

\subsection{Model Training Strategy}
In our exploration of candidate pairs for binary classification across various entity type recognition tasks, we have observed a practical imbalance between positive and negative samples within these pairs. More specifically, there tends to be a significantly higher production of negative samples compared to positive ones. For example, the alignment of two knowledge graphs, each containing 1,000 entity types, can generate one million candidate pairs, amongst which merely hundreds may be positive samples. This disproportionate ratio often leads to overfitting in machine learning models trained on such unbalanced datasets.

To alleviate this imbalance issue, we introduce a model training strategy that emphasizes the weight of positive samples, aiming to foster a more balanced training dataset and mitigate the risk of overfitting. This is achieved by duplicating a portion of the positive samples, thereby optimizing entity type recognition performance by maintaining a positive-to-negative sample ratio of approximately 1:10. It is important to note that this data augmentation strategy is not extended to the testing set. Instead, the candidate pairs in the testing set are randomly selected to ensure the absence of any bias or interference in the evaluation process.

\subsection{Dealing with Trivial Samples}
In order to decrease the negative samples and avoid generating unnecessary candidate pairs, we prune the trivial samples which are obviously to be negative. Regarding different tasks, we have:
\begin{itemize}
    \item For the schema-level recognition task, we apply string-based measurements to filter the obvious negative samples. Given two entity types $E_a$ and $E_b$ from a candidate pair, we define the pre-selection factor $PS_s$ as: 
\begin{equation}
PS_s = Ngram(E_a,E_b) + Word2Vec(E_a,E_b)
\label{PS_s}
\end{equation}
    where $Ngram(\cdot)$ and $Word2Vec(\cdot)$ are two similarity measurements that are lexical-based and semantic-based, respectively. Thus, we consider $(E_a,E_b)$ to be an obvious negative candidate entity type pair if $PS_s$ is greater than the threshold $th$. Experimentally, we find that $th = 0.3$ will lead to better results.
    
    \item For the instance-level recognition task, the obvious negative candidate pair is identified when two entities $I_a$ and $I_b$ have no shared property. Thus, such candidate pairs will be pruned before inputting to the entity type recognizer.
\end{itemize}
The implementation of a pre-selection process for trivial samples significantly contributes to the reduction of runtime in model training phases. Furthermore, this strategic approach aids in diminishing the propensity for overfitting. It achieves this by eliminating superfluous samples, thereby streamlining the training dataset. This refinement of the training process not only enhances the efficiency of model training but also improves the recognition performance. 

\section{Quantitative Evaluation}

In this section, we aim to evaluate our proposed entity type recognition method for achieving knowledge extraction. First, we introduce the experimental setups, including the datasets we used and the evaluation strategy. Then we present the analysis and quantitative evaluation results. Finally, we conduct the ablation studies to explain the setting of parameters.

\subsection{Experimental Setup}

\subsubsection{Dataset Selection.}
For evaluating the result of entity type recognition, we exploit OAEI as the main reference for the selection of the schema-level recognition cases. As of today, this in fact the major source of schema alignment problems. Our approach focuses on knowledge graphs that contain entity types associated with a fair number of properties. As a result, we have selected the following cases: the bibliographic ontology dataset (BiblioTrack) \cite{euzenat2009results} and conference track (ConfTrack) \cite{zamazal2017ten} ($ra1$\footnote{https://owl.vse.cz/ontofarm/} version). From the bibliographic ontology dataset, we select \#101-103 and series \#301-304, which present real-life ontologies for bibliographic references from the web. We select the alignment between \#101 and \#304 as the training set for training our ML-based entity type matcher, and the rest of the ontology alignments as the testing set. The conference track contains 16 ontologies, dealing with conference organizations, and 21 reference alignments. We set all 21 reference alignments from the conference track as the testing set to validate our entity type matcher. Notice that we select the training and testing set from different datasets since we aim to prove the adaptation of our approach, which also prevents our approach from overfitting. 

For evaluating the performance of instance-level recognition, we build a dataset called EnType, since there is no publicly released dataset for such recognition task between two knowledge graphs. We exploit DBpedia infobox dataset\footnote{http://wikidata.dbpedia.org/services-resources/ontology} as the reference knowledge graph for providing reference entity types. DBpedia is a general-purpose knowledge graph that contains common entity types in the real world, where sufficient properties are applied for describing these entity types. Then we select candidate entities from DBpedia, SUMO\footnote{https://www.ontologyportal.org/}, and several domain-specific datasets.
The entities we selected mainly according to common entity types, more specifically, \textit{Person, Place, Event, Organization} and their sub-classes. Finally, we obtain 20,000 entity type-entity candidate pairs, where 6,000 from DBpedia (EnType$_{Self}$) and 14,000 from the remaining resources (EnType$_{Gen}$). Need to notice, that this dataset will be randomly separated into the training and testing set to implement the ML model.

\subsubsection{Evaluation Metrics.}
In our experiment, we exploit standard evaluation metrics including precision (\textbf{Prec.}), recall (\textbf{Rec.}) and $F_1$-measure (\textbf{$F_{1}$}\textbf{-m.}), and additional $F_{0.5}$-measure (\textbf{$F_{0.5}$}\textbf{-m.}) and $F_2$-measure (\textbf{$F_{2}$}\textbf{-m.}) \cite{pour2020results} to comprehensively validate our method and compare it with state-of-the-art methods. We form the recognition candidates as pairs, where each pair consists of a reference entity type and a candidate entity type/entity. Precision refers to the fraction of correctly identified pairs among all identified pairs. Recall is the fraction of correctly identified pairs among all pairs in the ground-truth alignment.
$F_{\beta}$-measures (\textbf{$F_{\beta}$}\textbf{-m.}) are defined as the harmonic mean of recall and precision:
\begin{equation}
F_{\beta}\textbf{-m.} = (1+\beta^2) * \frac{Prec.* Rec.}{\beta^2*Prec.+Rec.}
\label{Fscore}
\end{equation}
where $\beta$ is the configuration factor that allows for weighting precision or recall more highly if it is more important for the use case. And metrics \textbf{$F_{1}$}\textbf{-m.}, \textbf{$F_{2}$}\textbf{-m.} and \textbf{$F_{0.5}$}\textbf{-m.} are defined when $\beta = \{1, 2, 0.5\}$, respectively. We consider the $F_{\beta}$\textbf{-m.} to be the most relevant metrics for evaluation since it reflects both recall and precision.

\subsection{Schema-level Entity Type Recognition}


\begin{table*}[!t]
\centering
\caption{Quantitative comparisons on schema-level entity type recognition.}
  \label{comparison1}

	\resizebox{1\linewidth}{!}{%

\begin{tabular}{@{}lcccccccccc@{}}
\toprule
\multirow{2}{*}{\textbf{Methods}} & \multicolumn{5}{c}{\textbf{ConfTrack}} & \multicolumn{5}{c}{\textbf{BiblioTrack}}   \\ \cmidrule(l){2-6}  \cmidrule(l){7-11} & \multicolumn{1}{l}{\textbf{ Prec. }} & \multicolumn{1}{l}{\textbf{ Rec. }} & \multicolumn{1}{l}{ \textbf{$F_{0.5}$}\textbf{-m.} } & \multicolumn{1}{l}{ \textbf{$F_1$}\textbf{-m.} } & \multicolumn{1}{l}{ \textbf{$F_2$}\textbf{-m.} } & \multicolumn{1}{l}{\textbf{ Prec. }} & \multicolumn{1}{l}{\textbf{ Rec. }} & \multicolumn{1}{l}{ \textbf{$F_{0.5}$}\textbf{-m.} } & \multicolumn{1}{l}{ \textbf{$F_1$}\textbf{-m.} } & \multicolumn{1}{l}{ \textbf{$F_2$}\textbf{-m.} }\\ \midrule
FCAMap \cite{chen2019identifying}  & 0.680  & 0.625  & 0.668 & 0.651 & 0.635 & 0.820  & 0.783  & 0.812 & 0.801 & 0.790  \\
AML \cite{faria2013agreementmakerlight}    & 0.832  & 0.630 & 0.782 & 0.717 & 0.662      & 0.869  & 0.822 & 0.859 & 0.845 & 0.830  \\
LogMap \cite{jimenez2011logmap}  & 0.798  & 0.592  & 0.746 & 0.680 & 0.624    & 0.832  & 0.694  & 0.800 & 0.757 & 0.718  \\
Alexandre et. al. \cite{bento2020ontology} & 0.795 & 0.638  & 0.758 & 0.708 & 0.664   & 0.827 & 0.786  & 0.818 & 0.806 & 0.794     \\ 
Nkisi-Orji et. al. \cite{nkisi2018ontology}  & \textbf{0.860}  & 0.514  & 0.758 & 0.643 & 0.559   & 0.858  & 0.682  & 0.816 & 0.760 & 0.711  \\
LogMapLt \cite{jimenez2011logmap} & 0.716 & 0.554  & 0.676 & 0.625 & 0.580   & 0.796 & 0.781  & 0.793 & 0.788 & 0.784     \\ \hline

RF$_{ETR}$ &0.677	&\textbf{0.804}	&0.699	&0.735	&\textbf{0.775}
& 0.529 & \textbf{0.884} & 0.575 & 0.662 & 0.779  \\
SGD$_{ETR}$ &0.795	&0.635	&0.757	&0.706	&0.662
& 0.779 & 0.632  & 0.744  & 0.698  & 0.656 \\
DT$_{ETR}$ &0.624	&0.742	&0.644	&0.678	&0.715
& 0.671 & 0.703 & 0.677  & 0.687 & 0.696   \\
LR$_{ETR}$ &0.638	&0.779	&0.662	&0.701	&0.746
& 0.556  & 0.808 & 0.593  & 0.659  & 0.741 \\
XGBoost$_{ETR}$     & 0.827 & 0.676 & \textbf{0.792}  & \textbf{0.744}  & 0.702    &  \textbf{0.870}   & 0.832 & \textbf{0.862}  & \textbf{0.851}  & \textbf{0.840}    \\
ANN$_{ETR}$    & 0.813   & 0.604 & 0.760 & 0.693 & 0.636                     & 0.797 & 0.811 & 0.799 & 0.803  & 0.808    \\
\bottomrule                    
\end{tabular}}
\end{table*}

We evaluate the validity of the proposed recognition framework, where we apply several above-mentioned algorithms in the entity type recognizer, including random forest, stochastic gradient descent classifier, decision tree, logistic regression, XGBoost, and artificial neural network classifier (ANN), namely RF$_{ETR}$, SGD$_{ETR}$, DT$_{ETR}$, LR$_{ETR}$, XGBoost$_{ETR}$, and ANN$_{ETR}$, respectively. We compared our work with state-of-the-art entity type alignment methods\footnote{as most of them came out of previous OAEI evaluation campaigns}, including FCAMap \cite{chen2019identifying}, AML \cite{faria2013agreementmakerlight}, CNN-based schema aligning \cite{bento2020ontology}, word-embedding-based ontology alignment \cite{nkisi2018ontology}, LogMap and LogMapLt \cite{jimenez2011logmap}. We calculate the mentioned evaluation metrics for a comprehensive comparison. Notice that we focus on the result of \textit{entity type-entity type} alignment in this experiment.

Table \ref{comparison1} presents the outcomes of our approach utilizing various algorithms, compared with the results of state-of-the-art methods. Firstly, we can find our entity type recognizer yields different results when applying different machine learning algorithms. Both XGBoost and Random Forest demonstrate superior performance compared to other algorithms, and they also surpass other methods in BiblioTrack and in most scenarios within ConfTrack.  Additionally, ANN also presents a competitive performance for both datasets since neural networks are powerful in encoding complex features. The method by Nkisi-Orji et. al. \cite{nkisi2018ontology} leads the precision on ConfTrack, while it poorly performs on recall for both datasets. AML also shows competitive overall results compared to other state-of-the-art methods. Considering the average results of our approach with different models are performing better or close to the state-of-the-art, we can say that our approach surpasses the state-of-the-art competitors on the entity type alignment task\footnote{All methods do not have significant differences in running times.}. Furthermore, the consistently promising performance across different algorithms signifies the validity and adaptability of our proposed similarity metrics and entity type recognizer.

\subsection{Instance-level Entity Type Recognition}

In evaluating the performance across various instance-level scenarios, we incorporate two subsets, EnType$_{Self}$ and EnType$_{Gen}$, in our experiment. EnType$_{Self}$ comprises candidate entities and reference entity types from the same knowledge graph, signifying \textit{self-recognition}, which is anticipated to be a relatively simpler task. Conversely, \textit{general recognition} pertains to the assessment within EnType$_{Gen}$, where candidate entities are sourced from external resources. Table \ref{comparison2} displays the results of entity type recognition (measured by $F_1$-measure), employing algorithms such as XGBoost$_{ETR}$, SGD$_{ETR}$, RF$_{ETR}$, and ANN$_{ETR}$. Additionally, two state-of-the-art methods \cite{sleeman2015entity, giunchiglia2020entity} are included in this study.

It is observed that our recognizers, utilizing the proposed property-based similarity metrics, significantly outperform the two state-of-the-art methods. In the self-recognition group, XGBoost$_{ETR}$ exhibits superior performance in two types, while ANN$_{ETR}$ also delivers competitive results. In the context of general recognition, the XGBoost-based ETR method maintains stable efficacy, surpassing all comparative methods. The encouraging results in both scenarios attest to the effectiveness of our similarity metrics and recognizer in instance-level entity type recognition. Notably, the precision observed in the general recognition cases is lower compared to self-recognition, which aligns with the relative complexity of these two groups.

\begin{table*}[!t]
\centering

\caption{Quantitative evaluation of instance-level recognition on dataset EnType.}
  \label{comparison2}

\begin{tabular}{@{}lcccccccc@{}}
\toprule
\multirow{2}{*}{\textbf{Methods}} & \multicolumn{4}{c}{\textbf{Self recognition}} & \multicolumn{4}{c}{\textbf{General recognition}}   \\ \cmidrule(l){2-5}  \cmidrule(l){6-9} & \multicolumn{1}{c}{Person} & \multicolumn{1}{c}{Org.} & \multicolumn{1}{c}{Place} & \multicolumn{1}{c}{Event} & \multicolumn{1}{c}{Person} & \multicolumn{1}{c}{Org.} & \multicolumn{1}{c}{Place} & \multicolumn{1}{c}{Event} \\ \midrule

Sleeman et. al. \cite{sleeman2015entity} & 0.633 & 0.509 & 0.618 & 0.712 & 0.365 & 0.274 & 0.420 & 0.323\\
Giunchiglia et. al. \cite{giunchiglia2020entity} & 0.594 & 0.582 & 0.672 & 0.651 & 0.480 & 0.494 & 0.563 & 0.501\\ 
XGBoost$_{ETR}$        & 0.850 & \textbf{0.795}  & 0.820 & \textbf{0.814} & \textbf{0.634} & \textbf{0.509} & \textbf{0.610}  & \textbf{0.514} \\
SGD$_{ETR}$        & 0.698 & 0.703 & 0.764  & 0.732 & 0.418 & 0.462 & 0.535 & 0.472 \\
RF$_{ETR}$        & 0.752 & 0.772 & 0.815  & 0.810 & 0.596 & 0.438 & 0.568 & 0.490 \\
ANN$_{ETR}$        & \textbf{0.863} & 0.760 & \textbf{0.837}  & 0.808 & 0.589 & 0.435 & 0.497 & 0.486 \\
\bottomrule

\end{tabular}
\end{table*}

In acknowledgment of the influence that entity resolutions exert on the performance of entity type recognition, an auxiliary experiment is conducted, focusing on aligning entities with more granular entity types. We have chosen four sub-classes within the entity types \textit{person} and \textit{organization}, along with their respective entities, to form candidate pairs. Given the previously observed promising performance of the XGBoost and ANN algorithms, this experiment continues to compare XGBoost$_{ETR}$ and ANN$_{ETR}$ with the same state-of-the-art methods. Table \ref{comparison3} illustrates the $F_1$-measure outcomes for this recognition exercise. Our methods consistently exhibit superior recognition performance in comparison to other methods across all cases. Both the XGBoost$_{ETR}$ and ANN$_{ETR}$ models demonstrate promising overall effectiveness in recognizing specific entity types. Notably, ANN$_{ETR}$ achieves better performance in categorizing \textit{Comedian} and \textit{Company}, whereas XGBoost$_{ETR}$ excels in the remaining scenarios. The experimental results show that our similarity metrics and approach can also be applied for specific instance-level entity type recognition, which further supports the performance of the knowledge extraction. 

\begin{table*}[!t]
\centering

\caption{Quantitative evaluation of instance-level recognition on different entity resolutions.}
\label{comparison3}
\resizebox{1\columnwidth}{!}{

\begin{tabular}{@{}lcccccccc@{}}
\toprule
\multirow{2}{*}{\textbf{Methods}} & \multicolumn{4}{c}{\textbf{Person}} & \multicolumn{4}{c}{\textbf{Organization}}   \\ \cmidrule(l){2-5}  \cmidrule(l){6-9} & \multicolumn{1}{l}{Athlete} & \multicolumn{1}{l}{MilitaryPerson} & \multicolumn{1}{l}{Artist} & \multicolumn{1}{l}{Comedian} & \multicolumn{1}{l}{MilitaryOrg.} & \multicolumn{1}{l}{Company} & \multicolumn{1}{l}{SportsClub} & \multicolumn{1}{l}{ReligiousOrg.} \\ \midrule
Sleeman et. al. \cite{sleeman2015entity} & 0.639 & 0.472 & 0.690 & 0.597 & 0.479 & 0.463 & 0.637 & 0.510 \\
Giunchiglia et. al. \cite{giunchiglia2020entity} & 0.658  & 0.429 & 0.736 & 0.522 & 0.512 & 0.487 & 0.624 & 0.482 \\  
XGBoost$_{ETR}$     & \textbf{0.756}   & \textbf{0.508}   & \textbf{0.760}  & 0.581   & \textbf{0.656}   & 0.487          & \textbf{0.719}  & \textbf{0.550}\\ 
ANN$_{ETR}$    & 0.712   & 0.507          & 0.732  & \textbf{0.659}  & 0.613   & \textbf{0.528}          & 0.702  & 0.491  \\
\bottomrule
\end{tabular}
}
\end{table*}

\subsection{Ablation Study}
In this section, we conduct ablation studies to substantiate the efficacy of specific components incorporated within our entity type recognition framework.

\subsubsection{Effect of Similarity Metrics}
The first ablation study is to evaluate if each of the proposed property-based similarity metrics is effective. In this experiment, we test the backbones\footnote{trained by all three metrics ($Sim_V, Sim_H, Sim_I$)} (B) which were used in the schema-level and instance-level recognition tasks, respectively. Based on the backbones, we also design a controlled group that includes models trained without one of the property-based similarity metrics (i.e. B-Sim$_V$, B-Sim$_H$ and B-Sim$_I$) and models trained without all metrics (i.e. B-L). If the backbones perform better than the corresponding models in the controlled group, we can quantitatively conclude that each of the property-based similarity metrics ($Sim_V, Sim_H, Sim_I$) contributes to the entity type alignment and recognition tasks. Table \ref{comparison4} demonstrates the $F_1$-measure of each group. We apply ConfTrack and EnType$_{Gen}$ for the recognition cases. Note that we select two models for each case as Table \ref{comparison4} shows. We find that backbones perform better than models in the controlled group, especially for models trained without all metrics. Thus, we consider all property-based similarity metrics to contribute to better recognition performance. Particularly, Sim$_V$ and Sim$_H$ significantly affect the performance of schema alignment cases, and Sim$_I$ affects instance-level cases more.

\begin{table}[!t]
\centering

\caption{Ablation study on property-based similarity metrics.}
\label{comparison4}
\begin{tabular}{@{}llccccc@{}}
\toprule
\multicolumn{1}{c}{Dataset} & \multicolumn{1}{c}{Model}  & Backbone & B-Sim$_V$ &B-Sim$_H$ & B-Sim$_I$   & B-L\\ \midrule
\multirow{2}{*}{ConfTrack} & ANN$_{ETR}$  & \textbf{0.713}    & 0.635     & 0.639    & 0.660  & 0.618     \\
\addlinespace & XGBoost$_{ETR}$                & \textbf{0.740}    & 0.648     & 0.655    & 0.694  & 0.632     \\ \hline
\multirow{2}{*}{EnType$_{Gen}$} & ANN$_{ETR}$ & \textbf{0.537}  & 0.327  & 0.391  & 0.309  & -    \\
\addlinespace & XGBoost$_{ETR}$                    & \textbf{0.559}  & 0.413  & 0.402  & 0.385  & -     \\\bottomrule
\end{tabular}
\end{table}

\subsubsection{Effect of Constraint Factors}

In section 4.4.2, we introduced a constraint factor, denoted as $\lambda$, for the computation of the metric $Sim_H$. The objective of this study is to statistically ascertain the optimal value for $\lambda$. Utilizing the ConfTrack dataset and its two most effective models, we set the value of $\lambda$ at regular intervals ranging from 0.1 to 1, represented by discrete points. Our evaluation focuses on determining whether this pre-defined factor influences the ultimate performance in entity type recognition and deducing the most suitable value of $\lambda$ for generic entity type recognition tasks. Table \ref{comparison5} exhibits these results, with a special emphasis on both the best and second-best outcomes. It is observed that varying values of $\lambda$ indeed impact the final performance in entity type recognition. Both models display a similar trend, indicating that the optimal value of $\lambda$ approximates 0.5. Consequently, we set $\lambda = 0.5$ for the calculation of the metric $Sim_H$ in our experimental framework.

\begin{table}[!t]
\centering

\caption{Ablation study on the constraint factor $\lambda$. The best and second-best results are highlighted in {\color[HTML]{FF0000} red} and {\color[HTML]{0070C0} blue}, respectively.}
\label{comparison5}
\begin{tabular}{@{}clccccccccc@{}}
\toprule
\multicolumn{1}{c}{Factor} & \multicolumn{1}{c}{Model} & 0.1 & 0.2 & 0.3 & 0.4 & 0.5 & 0.6 & 0.7 & 0.8 & 0.9 \\ \midrule
\multirow{2}{*}{$\lambda$} & ANN$_{ETR}$ 
& 0.613 & 0.638     & 0.670   & 0.678  & {\color[HTML]{FF0000} 0.712} & {\color[HTML]{0070C0} 0.685}  & 0.679 & 0.653  & 0.630 \\
\addlinespace & XGBoost$_{ETR}$   
& 0.662 & 0.677 & 0.714   & {\color[HTML]{0070C0} 0.727} &  {\color[HTML]{FF0000} 0.729} & 0.716 & 0.654 & 0.711 &  0.705  \\ \bottomrule
\end{tabular}
\end{table}

\section{Summary}
In this chapter, we introduce a machine learning-based framework for achieving entity type recognition. Initially, we outline the comprehensive structure of this framework, detailing its constituent components: the knowledge graph parser, the formalization module, the similarity calculation module, the property matcher, and the entity type recognizer. Subsequently, we explore a variety of machine learning algorithms aimed at enhancing knowledge extraction performance, encompassing both several classical algorithms and neural networks. Following this, we discuss the training strategy for the critical entity type recognizer, as well as the data processing pipeline. Finally, we assess the effectiveness of our framework across multiple scenarios using various datasets. When compared to state-of-the-art methods, the experimental outcomes affirm the validity of our similarity metrics and the superiority of our proposed recognition framework, both quantitatively and qualitatively. Additionally, ablation studies are conducted to further validate the efficacy of the proposed similarity metrics and to ascertain the optimal parameter settings for the recognition framework.

\chapter{Knowledge Graph Assessment}
\section{Introduction}
In this chapter, we introduce the knowledge graph assessment methods, contributing to the evaluation step of our proposed knowledge graph extension framework. This study is based on our work presented in \cite{fumagalli2021ranking}. 

Knowledge graph assessment is a pivotal procedure in semantic computing and data science, focusing on evaluating the accuracy, completeness, and reliability of knowledge graphs. Such assessments are fundamental for applications including search engines, recommendation systems, and natural language processing tasks. The assessment methods are also applied to reduce the effort in constructing knowledge graphs and to promote their reuse \cite{mcdaniel2019evaluating}.

Significant efforts have been made in the functional, or purpose-driven, evaluation of knowledge graphs, which especially pertains to their purpose-based ranking \cite{butt2016dwrank}. The primary challenge here is to quantify the purpose-driven features of a knowledge graph, which are often heavily reliant on context and application \cite{gangemi2005theoretical}. For example, consider the critical significance of the term ``core” in describing entity types and their varied interpretations \cite{falbo2016ontology}. With the growing abundance of knowledge graphs, the necessity to define methodologies for their reuse intensifies \cite{degbelo2017snapshot}, especially in emerging application domains, such as relational learning \cite{nickel2015review} and transfer learning \cite{fumagalli2020ontology}. This increasing proliferation of knowledge graphs calls for refined approaches to ensure their efficient and effective reuse in diverse contexts.

Following contemporary psychology, the purpose of categorization can be reduced to ``\textit{...a means of simplifying the environment, of reducing the load on memory, and of helping us to store and retrieve information efficiently}'' \cite{rosch1999principles, harnad2017cognize}. This process involves grouping entities based on shared features or similarities, where categorization typically employs well-defined and effective constructs such as knowledge graphs. Accordingly, we introduce a quantifiable and deterministic way to assess knowledge graphs according to their \textit{categorization purpose}. We measure the categorization purpose as a novel perspective of knowledge graph assessment, via metrics that we ground on the notion of categorization defined in cognitive psychology \cite{millikan2017beyond}.

Thus, we propose a set of metrics called \textit{focus} to assess knowledge graphs from their categorization purpose. We will introduce the details of how focus can be used to rank: \textit{i).} the concepts inside a knowledge graph which are more/less informative; \textit{ii).} the concepts across multiple knowledge graphs which are more/less informative; and \textit{iii).} the knowledge graphs which are more/less informative. We also discuss existing assessment metrics that are used to evaluate knowledge graphs from dimensions including completeness, consistency, and data quality. Lastly, to test the utility of the focus measures, we show how focus metrics can be used to support engineers in measuring the relevance of knowledge graphs. That is, \textit{a).} we verify how the knowledge graphs ranking provided by the focus metrics reflects the ranking of the knowledge graphs provided by a group of knowledge engineers, according to guidelines inspired by a well-known experiment in cognitive psychology; \textit{b).} we verify how focus can help scientists in selecting better knowledge graphs for supporting the training process in machine learning tasks.

\section{Existing Assessment Metrics}

Effective assessment of knowledge graphs is crucial for ensuring that they serve their intended purpose accurately and efficiently in various application scenarios. Assessment methodologies often employ a mix of automated tools and manual inspection, especially when comparing the graph's data against a gold standard or benchmark dataset. We introduce several existing ranking metrics as baseline measurements, namely \textit{class match measure (CMM)} \cite{alani2006metrics}, \textit{density measure (DEM)} \cite{alani2006metrics}, \textit{Okapi best matching measure (BM25)} \cite{butt2014ontology} and \textit{term frequency-inverse document frequency measure (TF-IDF)} \cite{salton1988term}.

\begin{enumerate}
    \item \textbf{CMM}: The class match measure is designed to assess the extent to which a knowledge graph encompasses given search terms by identifying entity types with labels that either exactly match (where the label is identical to the search term) or partially match (where the label contains the search term) the search term. This measure posits that a knowledge graph encompassing all the specified search terms will attain a higher score compared to others, and prioritizes exact matches over partial ones. For example, when searching for \textit{student} a knowledge graph with an entity type labeled precisely as \textit{Student} will be rated higher in this measure compared to other knowledge graphs that include partial matches, such as \textit{PhDstudent}. The CMM measure can be mathematically represented by:
\begin{equation}
CMM(K, Q)= \alpha score_{EMM}(K, Q) + \beta score_{PMM}(K, Q)
\label{F(CMM measure)}
\end{equation}
where $score_{EMM}(K, Q)$ and $score_{PMM}(K, Q)$ represent the scores attributed to exact alignments and partial alignments within the knowledge graph $K$ concerning the query $Q$, variables $\alpha$ and $\beta$ serve as weight factors for these two distinct aligning states. Given that exact aligning is preferred over partial aligning, the typical configuration sets $\alpha$ at 0.6 and $\beta$ at 0.4.

    \item \textbf{DEM}: In the context of evaluating an effective representation of a particular concept, people would expect to find a certain degree of detail in the representation of the knowledge pertaining to that concept. The density measure is conceived to estimate the representational density or the information content inherent in classes, thereby reflecting the depth of knowledge details. Currently, the calculations for density are confined to aspects such as the number of subclasses, the number of properties associated with the concept, the count of sibling concepts, and similar elements. The DEM measure can be represented as:
\begin{equation}
DEM(K)=  \frac{1}{n}\sum_{i=1}^{n}\sum_{j=1}^{p}w*M(S_j(c_i),q)
\label{F(DEM measure)}
\end{equation}
where $M(S_j(c_i),q)$ represents the aligning state between the query $q$ and the aspect $S_j(c)$ of the concept $c$, notice that $S_j$ is selected from $S= \{ relations(c), superclasses(c), subclasses(c), siblings(c) \} $, and the factor $p$ is the number of items in set $S$, $n$ refers to the number of alignments, and $w$ is the weight factor set to a default value of 1. 

    \item \textbf{TF-IDF}: The term frequency inverse document frequency measure is an information retrieval statistic that reflects the importance of a word to a document in a collection or corpus. It is also widely used for knowledge graph ranking by calculating relevance between potential knowledge graphs with an automatically gathered concept that describes the domain of interest. TF-IDF can be modeled as:
\begin{equation}
TF-IDF(K, Q)= TF(K, Q) * IDF(K, Q)
\label{F(TF-IDF measure)}
\end{equation}
where $TF(K, Q)$ denotes the term frequency for query $Q$ within the knowledge graph $K$, and $IDF(K, Q)$ pertains to the prevalence of a resource throughout the resources, which is derived by dividing the aggregate count of knowledge graphs by the number of resources that include the query $Q$, followed by calculating the logarithm of the resulting quotient.

    \item \textbf{BM25}: The Okapi BM25 algorithm is a widely used information retrieval technique. Exploiting the probabilistic information retrieval framework, BM25 modifies and extends the TF-IDF measure. BM25 can effectively differentiate between the occurrence of a term in a document relative to its frequency in the corpus, while also preventing overemphasis on longer documents. It excels in scenarios where relevant documents vary in length, as it can adeptly balance the frequency of terms with the length of the document. BM25 can be presented as:
\begin{equation}
BM25(K, Q)= \sum_{i=1}^{n}IDF(K, q_i) \frac{TF(K, q_i)*m +1}{TF(K, q_i)+ m* (1-b+b*\frac{|d|}{avg})}
\label{F(BM25 measure)}
\end{equation}
where $|d|$ is the length of the resources $d$ in words, and $avg$ is the average document length in the text collection from which the resources are drawn, two key parameters $m$ and $b$ control the saturation of term frequency and the length normalization, respectively. 

\end{enumerate}

The above-mentioned assessment metrics are widely applied to measure the knowledge graphs from dimensions including completeness, consistency, and data quality. However, these metrics lack the consideration of the categorization, which is one of the main purposes of the construction of knowledge graphs. Thus, we are attempting to improve the current knowledge graph assessment by proposing a novel set of assessment metrics from the categorization perspective.

\section{Measuring Categorization Purpose by Focus}

\subsection{Intuition}

Envision a scenario where an individual specifies ``the book on my desk" to have another person retrieve that specific book based on her description. Such object descriptions are intended to facilitate the retrieval of the designated objects. The essence of this process is to precisely describe the cluster of objects, which underpins the categorization purpose of an object description. These object descriptions, also termed types or categories, form the foundation of the organization of our mental life \cite{millikan2017beyond}.

In line with contemporary psychological insights, particularly the seminal contributions of Eleanor Rosch \cite{rosch1999principles}, the purpose of categorization can be understood through two principal dimensions: \textit{i).} maximizing the number of features that characterize the members of a specific category, and \textit{ii).} minimizing the number of features shared with other categories. To assess these dimensions, Eleanor Rosch introduced the concept of \textit{cue validity} \cite{rosch1975family}, defined as ``the conditional probability  $p(c_{j} | f_{j})$ that an object falls in a category $c_{j}$ given a feature $f_{j}$". Cue validity was instrumental in identifying basic-level categories, which are those that maximize intra-category feature commonality while minimizing inter-category feature overlap.

Rosch's formulations were initially applied in experiments where individuals were asked to classify objects into specified categories. We adapt Rosch's original approach to the realm of knowledge graph engineering. In our context, each available knowledge graph, such as \textit{schema.org} or \textit{DBpedia}, is conceptualized as a categorization, modeled as a collection of entity types associated with properties. The primary role of these categorizations is to ``draw sharp lines around the types of entities they contain, ensuring each member definitively falls within or outside each entity type" \cite{2019jowo, giunchiglia2020entity}. Knowledge engineers assume a role similar to the participants in Rosch's study. Every knowledge graph schema offers a rich set of categorization instances. Each entity type functions as a category and all entity type properties act as features. The categorization purpose of a knowledge graph, which we term \textit{focus}, is modeled as ``the state or quality of being relevant in storing and retrieving information". We quantify the degree of this pertinence by adapting \textit{cue validity} in the following manner:
\begin{itemize}
\item We take each property to have the same cue validity;
\item For each knowledge graph we equally divide the property cue validity across the entity types the properties are associated with;
\item We quantify the relevance of the knowledge graph and entity types in storing and retrieving information by checking the spreading of cue validity.   
\end{itemize}

Thus, the focus can be then calculated based on the above assumptions and, in turn, it can be functionally articulated in:
\begin{itemize}
\item \textit{Entity Types Focus}: This aspect pertains to the capability to identify those entity types that constitute the most informative categories. These are the categories with heightened relevance for categorization, or more specifically, those that maximize the number of unique properties and minimize the overlap of properties with other categories. To a certain extent, these entity types correspond to what expert users deem as ``core entity types" within a specific domain.

\item \textit{Knowledge Graphs Focus}: This refers to the ability to pinpoint knowledge graphs that encapsulate the greatest number of highly informative (focused) entity types. Such knowledge graphs are characterized, to a certain degree, as ``clean" or ``not-noisy" and are associated with what expert users regard as well-structured knowledge graphs \cite{paulheim2017knowledge}.
\end{itemize}

\subsection{Metric Focus of Entity Types}
According to the notion of entity type focus we introduced above, we model the metric focus of entity type $Focus_{e}$ as:

\begin{equation}
Focus_e(e) =  Cue_{e}(e) + \eta Cue_{er}(e) = Cue_{e}(e)(1 + \frac{\eta}{|prop(e)|})
\end{equation}
where $e$ represents an entity type, the $Focus_{e}$ results from the summation of $Cue_{e}$  and $Cue_{er}$. $Cue_{e}$ represents the \textit{cue validity} of the entity type. $Cue_{er}$ represents a normalization of $Cue_{e}$. $\eta$ represents a constraint factor to be applied over $Cue_{er}$. The constraint factor $\eta$ is used to manipulate the weight of the $Cue_{er}$, where $\eta > 0$, thereby affecting the value of the metric $Focus_{e}$. Specifically, two parts of the function can also be combined, in which $|prop(e)|$ is the number of properties associated with the specific entity type $e$. 

To model $Cue_{e}$ and $Cue_{er}$ we mainly adapted the work presented in \cite{giunchiglia2019knowledge}. In order to calculate $Cue_{e}$, firstly, we define the \textit{cue validity} of a property \textit{p} associated with an entity type $e$, also called $Cue_{p}$, as: 

\begin{equation}
Cue_{p}(p, e) = \frac{PoE(p,e)}{|dom(p)|} \in [0, 1]
\end{equation}
where $|dom(p)|$ presents the cardinality of entity types that are the domain of the specific property $p$. $PoE(p, e)$ is defined as:

\begin{equation}
PoE(p,e) = \left\{\begin{matrix}1, if e \in dom(p)
\\ 
0, if e \notin dom(p)
\end{matrix}\right.
\end{equation}
$Cue_{p}(p, e)$ returns 0 if $p$ is not associated with $e$. Otherwise returns $1/n$, where $n$ is the number of entity types in the domain of $p$. In particular, $Cue_{p}$ takes the maximum value 1 if $p$ is associated with only one entity type.

Given the notion of $Cue_{p}$ we provide the notion of \textit{cue validity} of an entity type. $Cue_{e}$ is related to the sum of the \textit{cue validity} of the properties associated with the specific entity type $e$ and is modeled as follows:

\begin{equation}
Cue_{e}(e) = \sum_{i=1}^{|prop(e)|} Cue_{p}(p_{i}, e) \in [0, |prop(e)|]
\end{equation}
where $Cue_{e}$ provides the \textit{centrality} of an entity type in a given knowledge graph, by summing all its properties $Cue_{p}$. $Cue_{e}$ refers to the maximization of the properties associated with entity type $e$ with the members of its categories.

Given the notion of $Cue_{e}$, we capture the minimization level of the number of properties shared with other entity types, inside a knowledge graph with the notion of $Cue_{er}$, which we define as:

\begin{equation}
Cue_{er}(e) = \frac{Cue_{e}(e)}{|prop(e)|} \in [0, 1]
\end{equation}

After deriving $Cue_{e}$ and $Cue_{er}$ it is possible to calculate $Focus_{e}$. Notice that, to normalize the range of the metrics, we applied log normalization \cite{bornemann1981log} on $Cue_{e}$ since $|prop(e)|$ can be significantly unbalanced between entity types and min-max normalization \cite{jain2011min} on $Cue_{er}$.

\subsection{Metric Focus of Knowledge Graphs}
Following the knowledge graphs focus notion we introduced, we model the metric focus $Focus_{k}$ as follows:

\begin{equation}
Focus_{k}(K) =  Cue_{k}(K) + \mu Cue_{kr}(K) = Cue_{k}(K)(1 + \frac{\mu}{|prop(K)|})
\end{equation}
where we take $K$ as an input knowledge graph and we take $Focus_{k}$ as a summation of $Cue_{k}$ and $Cue_{kr}$. $Cue_{k}$ represents the \textit{cue validity} of the knowledge graph. $Cue_{kr}$ represents a normalization of $Cue_{k}$. $\mu$ represents a constraint factor for $Cue_{kr}$, which we assume $\mu>0$, as for $Cue_{er}$ above. $|prop(K)|$ refers to the number of the properties in $K$. 

The notions and terminology used for entity types, i.e., the notions of $Cue_{e}$ and $Cue_{er}$, can be straightforwardly generalized to knowledge graphs, generating the following metrics:  
\begin{equation}
Cue_{k}(K)= \sum_{i=1}^{|E_{K}|} Cue_{e}(e_{i})  \in [0, |prop(K)|]
\end{equation}
The $Cue_{k}(K)$ is calculated as a summation of the \textit{cue validity} of all the entity types in a given knowledge graph, which in the function is represented by $E_{K}$. $|prop(K)|$ refers to the number of the properties in the knowledge graph, as the maximization of $Cue_{k}$.  

Following the formalization of $Cue_{er}$ we capture the minimization level of the number of properties shared across the entity types inside the graph with the notion of $Cue_{kr}$, which we define as:
\begin{equation}
Cue_{kr}(K)= Cue_{k}(K)/\sum_{i=1}^{|E_{K}|}prop(e_{i}) \in [0, 1]
\end{equation}
where $Cue_{k}$ and $Cue_{kr}$ can be used to assess the focus of a whole knowledge graph. Notice that to normalize the metric $Focus_{k}$, we applied \textit{log normalization} on $Cue_{k}$, since $|prop(K)|$ may be significantly higher in some knowledge graphs than others knowledge graphs and \textit{min-max normalization} on $Cue_{kr}$.

\section{Metrics Validation}

\subsection{Analysis of Knowledge Graph Rankings}
In this section, we discuss the validity of the proposed focus metrics via a set of analyses that assess the collected knowledge graphs. We collected a data set of 700 knowledge graphs, expressed in the \textit{Terse RDF Triple Language (Turtle)}\footnote{ \href{https://www.w3.org/TR/turtle/}{\texttt{https://www.w3.org/TR/turtle/}}} format. Most of these resources have been taken from the LOV catalog\footnote{\href{https://lov.linkeddata.es/dataset/lov}{\texttt{https://lov.linkeddata.es/dataset/lov}}}. The remaining ones, see for instance \textit{freebase}\footnote{\href{https://developers.google.com/freebase}{\texttt{https://developers.google.com/freebase}}} and \textit{SUMO}\footnote{\href{http://www.adampease.org/OP/}{\texttt{http://www.adampease.org/OP/}}} have been added to extend the validation dataset.

\begin{figure}[!t]
  \centering
  \includegraphics[width=1\linewidth]{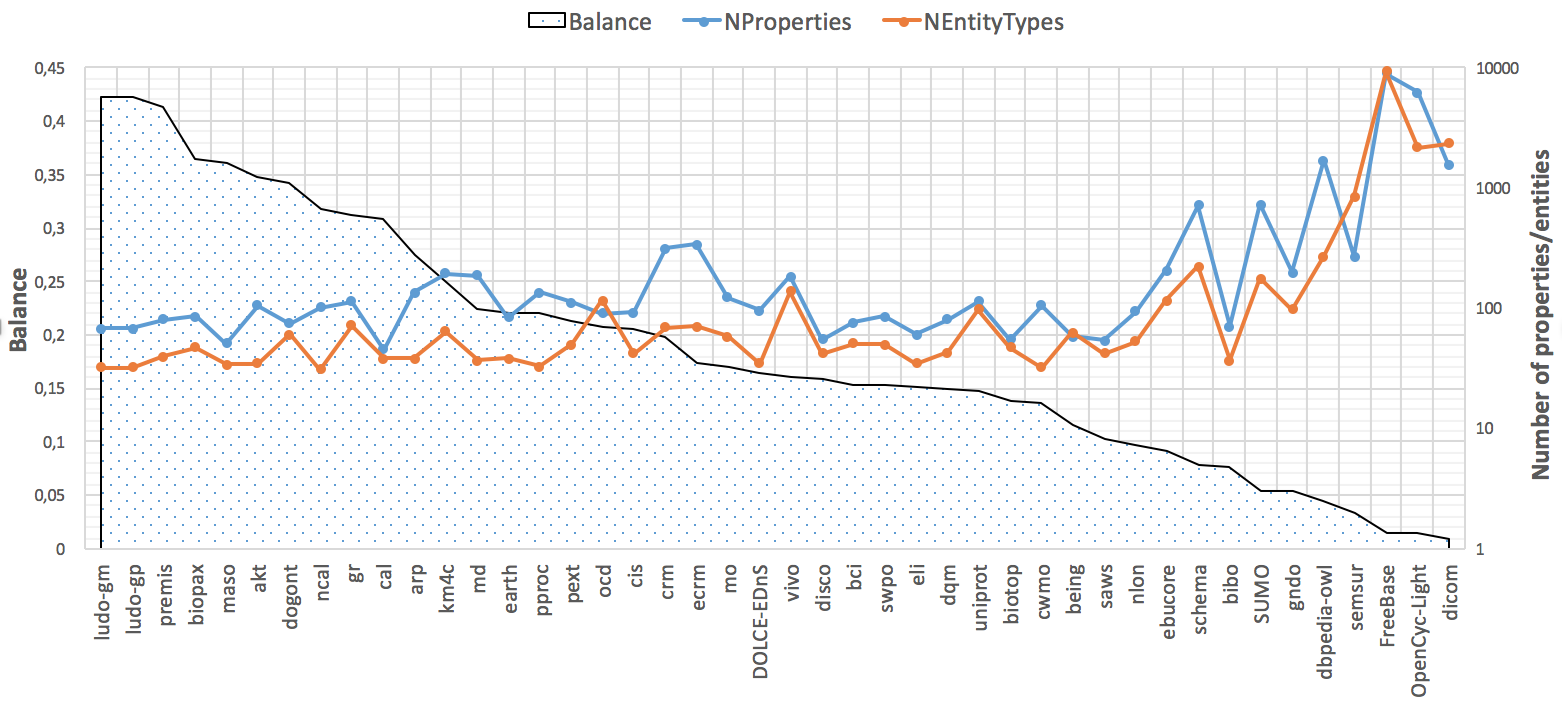}
  \caption{Knowledge graphs selected for the analysis.}
  \label{balanceMetric}
\end{figure}

For analytical purposes, we adopt the framework of FCA delineated in section 4.3, wherein all datasets are transformed into a collection of sets of triples (one set for each entity type), with each triple encoding information about ``entity type-property" associations. Furthermore, to generate the final output datasets, property labels were processed through an NLP pipeline which undertook various steps, including: \textit{i).} splitting a string at the occurrence of every capital letter; \textit{ii).} converting all characters to lower case; \textit{iii).} removing stop-words. This processing enabled us to conduct a more precise analysis.

From the initially processed dataset, we selected a subset by excluding all knowledge graphs containing fewer than 30 entity types. Fig.\ref{balanceMetric} offers a comprehensive overview of the final output dataset, illustrating for each of the remaining 44 knowledge graphs the number of properties, the number of entity types, and their balance. The balance metric reflects the distribution of a knowledge graph's properties across its entity types, which is calculated as follows:
\begin{equation}
Balance(K)=\frac{|prop(K)|}{|prop(e)|_{max} \ast |etype(K)|}
\end{equation}
where $|prop(K)|$ presents the cardinality of the set of properties in the knowledge graph, $|prop(e)|_{max}$ being the cardinality of the maximum set of properties associated with the entity type $e \in K$, and $etype(K)$ refers to the cardinality of the set of entity types in the knowledge graph.

\begin{figure}[!t]
  \centering
  \includegraphics[width=0.65\linewidth]{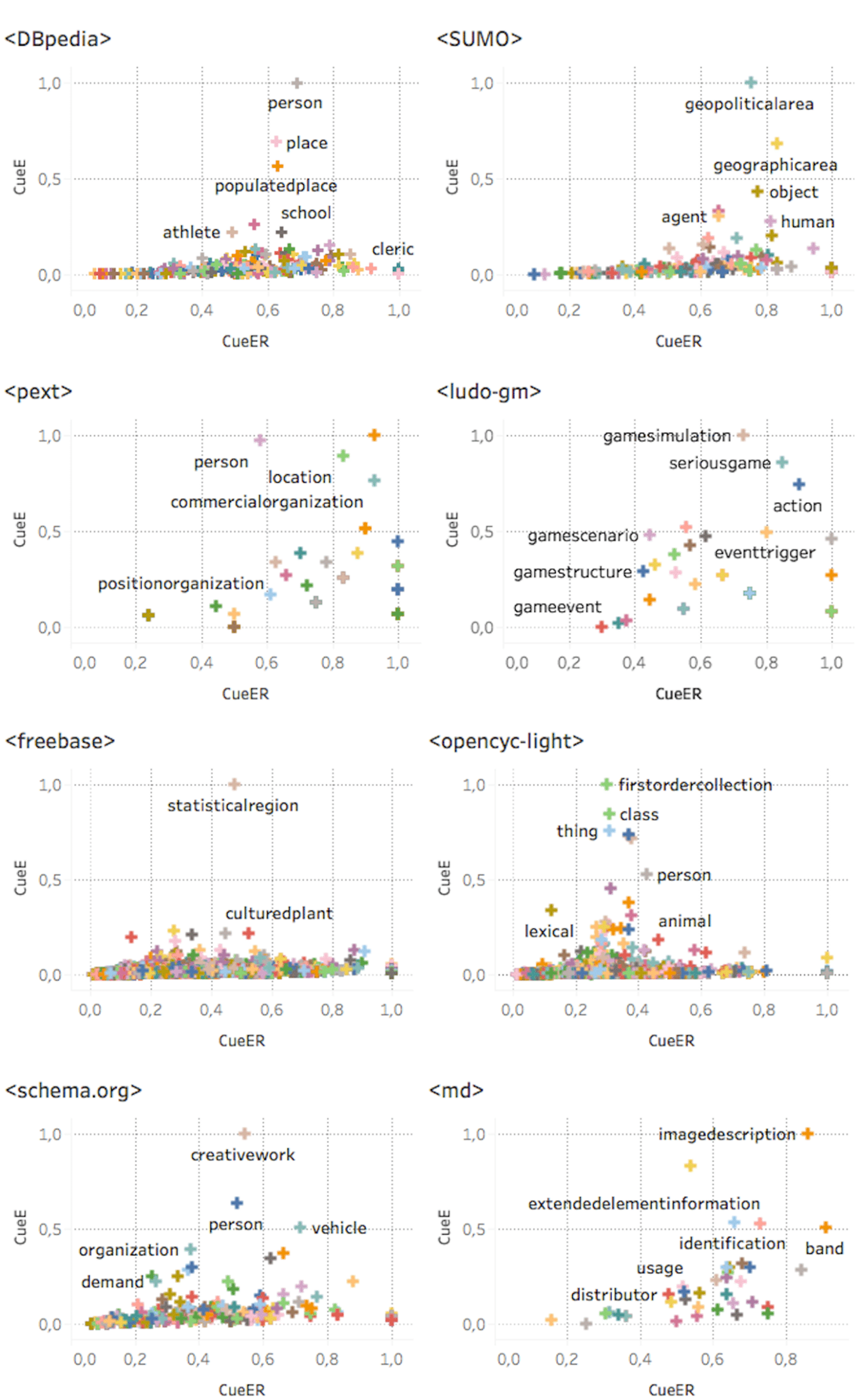}
  \caption{Entity types categorization relevance for example knowledge graphs.}
  \label{CueCategories}
\end{figure}

By applying the Cue metric $Cue_{e}(e)$ and $Cue_{er}(e)$ to the knowledge graphs of the resulting list, we obtained the scores to evaluate the categorization relevance of the entity types for each knowledge graph. Details are provided in Figure \ref{CueCategories}, where we randomly selected eight knowledge graphs and listed them according to the number of entity types. The selected knowledge graphs are: \textit{Freebase}, \textit{OpenCyc}\footnote{ \href{https://pythonhosted.org/ordf/ordf_vocab_opencyc.html}{\texttt{https://pythonhosted.org/ordf/ordf\_vocab\_opencyc.html}}}, \textit{DBpedia}, \textit{SUMO}, \textit{schema.org}, \textit{md}\footnote{\href{http://def.seegrid.csiro.au/isotc211/iso19115/2003/metadata}{\texttt{http://def.seegrid.csiro.au/isotc211/iso19115/2003/metadata}}}, \textit{pext}\footnote{\href{http://www.ontotext.com/proton/protonext.html}{\texttt{http://www.ontotext.com/proton/protonext.html}}} and \textit{ludo-gm}\footnote{\href{http://ns.inria.fr/ludo/v1/docs/gamemodel.html}{\texttt{http://ns.inria.fr/ludo/v1/docs/gamemodel.html}}}. The accompanying scatter plots display the correlations between the min-max normalized $Cue_{e}(e)$ and $Cue_{er}(e)$ for each entity type. Entity types positioned in the top-right quadrant are those with higher categorization relevance. For instance, in the \textit{SUMO} knowledge graph, prominent entity types include \textit{GeopoliticalArea} and \textit{GeographicalArea}, whereas in \textit{DBpedia}, notable examples are \textit{Person} and \textit{Place}. Thus, these observations validate that the metrics $Cue_{e}(e)$ and $Cue_{er}(e)$ indeed capture the informative categories from different knowledge graphs. 

\begin{table} [!t]
\caption{Ranking knowledge graphs by the metric $Focus_k(K)$.}
  \label{ranking1}
  \centering
\begin{tabular}{cccc}
\hline
\textbf{Knowledge Graphs}        & \textbf{$Cue_{k}(K)$} & \textbf{$Cue_{kr}(K)$}   & \textbf{$Focus_k(K)$} \\ \hline
\textbf{Freebase}      & 8981 & 0,21 & 1,15  \\
\textbf{cal}           & 46  & 0,98 & 0,92   \\
\textbf{bibo}          & 71   & 0,97 & 0,92   \\
\textbf{OpenCyc-l}       & 6266 & 0,26 & 0,90  \\
\textbf{swpo}          & 87  & 0,88 & 0,83   \\
\textbf{cwmo}          & 107  & 0,85 & 0,80   \\
\textbf{eli}           & 62   & 0,84 & 0,78   \\
\textbf{ncal}          & 103  & 0,80 & 0,75   \\
\textbf{mo}            & 124  & 0,79 & 0,74   \\
\textbf{akt}           & 106  & 0,79 & 0,74 \\ \hline
\end{tabular}
\end{table}

\begin{table} [!t]
\caption{Ranking entity types by the metric $Focus_e(e)$.}
  \label{ranking2}
\centering
\begin{tabular}{ccccc}
\hline
\textbf{Knowledge graph}    & \textbf{Entity type}   & \textbf{$Cue_{e}(e)$}   & \textbf{$Cue_{er}(e)$} & \textbf{$Focus_e(e)$} \\ \hline
\textbf{DBpedia}  & \textit{person}  & 169,02 & 0,69  & 1,42  \\
\textbf{OpenCyc-l}  & \textit{firstordercoll} & 230,59 & 0,30 & 1,30  \\
\textbf{Freebase} & \textit{statisticalreg}    & 161,53 & 0,48  & 1,17 \\
\textbf{Opencyc-l}  & \textit{class}   & 194,95 & 0,31  & 1,15 \\
\textbf{dicom}    & \textit{ieimage}  & 158,90 & 0,44  & 1,13  \\
\textbf{DBpedia}  & \textit{place}  & 116,97 & 0,63  & 1,13  \\
\hline
\end{tabular}
\end{table}

Meanwhile, we conducted several statistics to observe the performance of focus metrics. Firstly, we generated a ranking of these graphs by implementing the metric $Focus_k(K)$ across the 44 chosen knowledge graphs. The top 11 knowledge graphs in this ranking are presented in Table \ref{ranking1}. Additionally, by applying $Focus_e(e)$ to these knowledge graphs, we derived a ranking of entity types, with the top 6 entity types, in terms of categorization relevance, displayed in Table \ref{ranking2}. Furthermore, by selecting a specific entity type and applying $Focus_e(e)$, it is feasible to identify the most suitable knowledge graph for that particular entity type. Table \ref{ranking3} offers an illustration for the entity type \textit{Person}. The statistics demonstrate the feasibility of ranking knowledge graphs and entity types by focus metrics. Compared to the ranking results by cue metrics, focus metrics consider balanced features to capture categorization relevance.

\begin{table} [!t]
\caption{Ranking the entity type \textit{person} from different knowledge graphs}
  \label{ranking3}
\centering
\begin{tabular}{ccccc}
\hline
\textbf{Knowledge graph}    & \textbf{Entity type}  & \textbf{$Cue_{e}(e)$}   & \textbf{$Cue_{er}(e)$} & \textbf{$Focus_e(e)$} \\ \hline
\textbf{DBpedia}  & \textit{person} & 169,02 & 0,69 & 1,42  \\
\textbf{akt}      & \textit{person} & 8,00   & 1,00 & 1,03  \\
\textbf{OpenCyc-l}  & \textit{person} & 122,14 & 0,43  & 0,95 \\
\textbf{vivo}     & \textit{person} & 10,60 & 0,88 & 0,92  \\
\textbf{swpo}     & \textit{person} & 3,50  & 0,88 & 0,88  \\
\textbf{cwmo}     & \textit{person} & 5,83  & 0,83 & 0,85  \\
 \hline
\end{tabular}
\end{table}

\subsection{Validation of $Focus_e$}
In this section, we validate the efficacy of the $Focus_e(e)$ metric in measuring the categorization relevance of entity types, specifically their ability of information maximization. This analysis involves applying our metrics and corresponding state-of-the-art ranking algorithms to a set of knowledge graphs. We compare the results with a reference dataset formulated by 5 knowledge engineers\footnote{All knowledge engineers are certified Ph.D. candidates and joined the experiments voluntarily.}. These engineers were given a set of instructions for ranking the entity types, drawing inspiration from Rosch's experiment \cite{rosch1999principles}.

\begin{figure}[!t]
  \centering
  \includegraphics[width=1\linewidth]{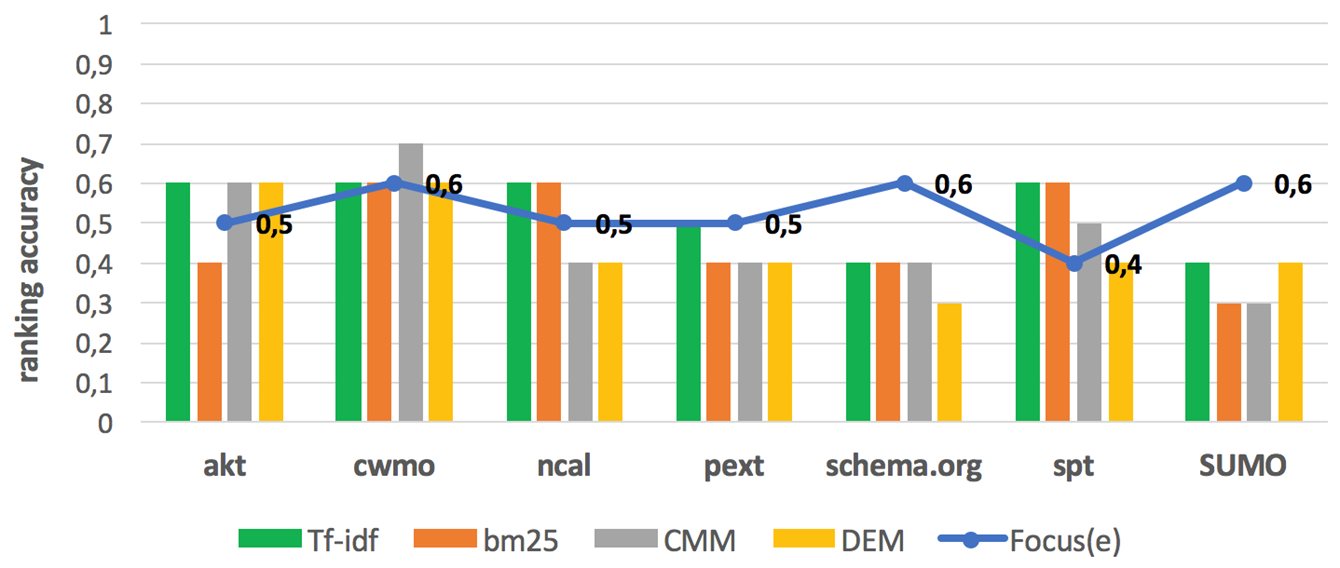}
  \caption{The experimental results on evaluating $Focus_e(e)$.}
  \label{focusE-chart}
\end{figure}

To evaluate our metric, we first select a subset of the knowledge graphs discussed in the previous section, including \textit{akt}\footnote{ \href{https://lov.linkeddata.es/dataset/lov/vocabs/akt}{\texttt{https://lov.linkeddata.es/dataset/lov/vocabs/akt}}}, \textit{cwmo}\footnote{\href{https://gabriel-alex.github.io/cwmo/}{\texttt{https://gabriel-alex.github.io/cwmo/}}}, \textit{ncal}, \textit{pext}, \textit{schema.org}, \textit{spt}\footnote{\href{https://github.com/dbpedia/ontology-tracker/tree/master/ontologies/spitfire-project.eu}{\texttt{https://github.com/dbpedia/\-ontology-tracker\-/tree/master\-/ontologies/spitfire-project.eu}}} and \textit{SUMO}. These particular knowledge graphs were chosen due to their diverse representation in terms of the number of properties and entity types, with almost all their entity type labels being comprehensible to humans\footnote{Many knowledge graphs encode their entity type labels with an ID.}. As a second step, we apply four introduced ranking metrics: \textit{TF-IDF} \cite{salton1988term}, \textit{BM25} \cite{robertson1995okapi}, \textit{Class Match Measure (CMM)} and \textit{Density Measure (DEM)} \cite{alani2006metrics}. The performance of these algorithms served as a baseline, selecting the top 10 entity types for each given knowledge graph, and then compared with the rankings generated by $Focus_e(e)$. The validity of our approach was quantified in terms of accuracy (ranging from 0 to 1), determined by how many entity types in the metrics ranking results aligned those in the ranking lists provided by the knowledge engineers. The outcomes of this experiment are depicted in Figure \ref{focusE-chart}, the blue line represents the accuracy of the ranking trend provided by $Focus_e(e)$. Each bar represents the accuracy of the ranking for the corresponding metric. 

The primary observation is that all referenced metrics exhibit a remarkably similar trend, demonstrating higher accuracy for knowledge graphs such as \textit{akt}, \textit{cwmo}, \textit{ncal}, and \textit{spt} while showing lower accuracy for \textit{schema.org} and \textit{SUMO}. However, this pattern is not mirrored by $Focus_e(e)$. Our metric, while not uniformly superior across all knowledge graphs, is notably effective with large and very noisy knowledge graphs (characterized by lower entity types $Cue_{er}$), as exemplified by \textit{schema.org} and \textit{SUMO}. This outcome aligns with our expectations and can be attributed to the significant emphasis we placed on minimizing the number of overlapping properties. The $Cue_{er}$ for each entity type indeed provides crucial insights into categorization relevance that may not be adequately captured if undue importance is given to the sheer number of an entity type’s properties. Consequently, for smaller and less noisy (or “clean” in terms of overlapping properties) knowledge graphs, other approaches that focus heavily on the quantity of entity type properties perform admirably (as evidenced by the strong performance of the \textit{TF-IDF} algorithm). In contrast, knowledge graphs with a vast array of entity types and low $Focus_e(e)$ more effectively discern categorization relevance. 

The second key observation is that \textit{TF-IDF} and $Focus_e(e)$ emerge as the top metrics in terms of average performance, with both achieving a mean accuracy of 0.52, compared to 0.47 for \textit{BM25} and \textit{CMM}, and 0.44 for \textit{DEM}. This result is underpinned by the fact that \textit{TF-IDF} consistently ranks as the best metric for small and not-noisy knowledge graphs, while $Focus_e(e)$ balances standard performance in small and clean knowledge graphs with high performance in large and noisy knowledge graphs.

\subsection{Validation of $Focus_k$ }
In this section, we analyze how focus can assist scientists in repurposing knowledge graphs for new application domains, such as statistical relational learning. Considering the absence of reference gold standards for assessing a knowledge graph's overall score, the primary objective of this assessment is to demonstrate the ability of the metric $Focus_k(K)$ on the predictive performance in a relational classification task. In this experiment, we utilized the same knowledge graphs selected in the previous experiment for $Focus_e(e)$, addressing relational classification, where the objective is to predict the labels of entity types. Note that we introduce a specific type of relational classification, as defined in \cite{giunchiglia2020entity}. The experimental setup was as follows: \textit{i).} training machine learning models by knowledge graphs selected based on their $Focus_k(K)$ values, where we choose the decision tree \cite{kaminski2018framework} and K-NN \cite{dasarathy1991nearest} algorithms; \textit{ii).} reporting the relative performance of the models in terms of accuracy differences and comparing these performances with the $Focus_k(K)$ for each knowledge graph.

 \begin{figure}[!t]
  \centering
  \includegraphics[width=1\linewidth]{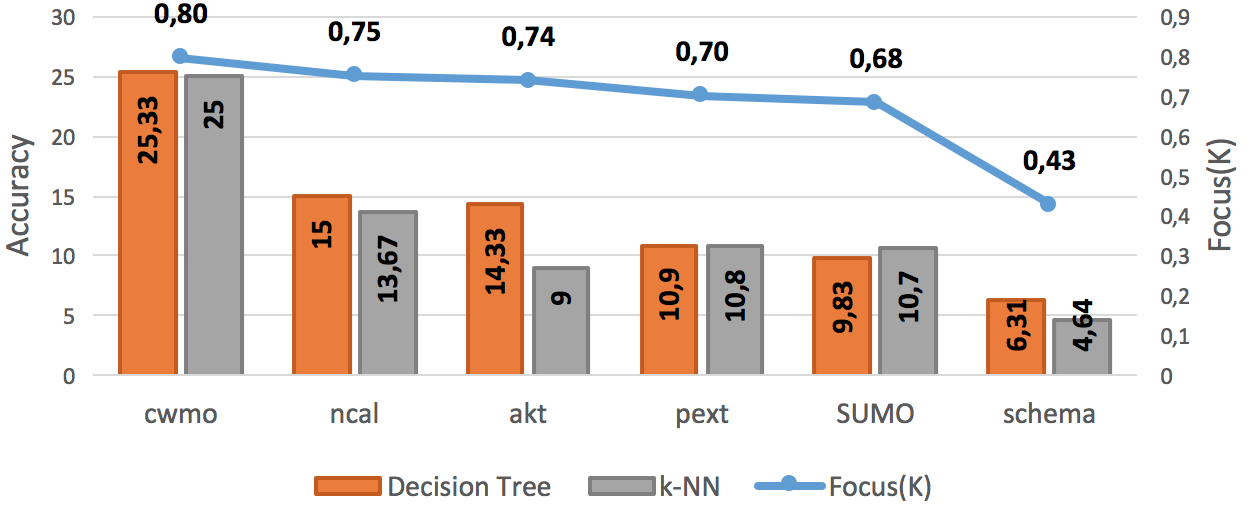}
  \caption{The experimental results on evaluating $Focus_k(K)$.}
  \label{focusC-experiment}
\end{figure}
 
As depicted in Figure \ref{focusC-experiment}, the accuracy is represented as the proportion of correct predictions, within the range of [0\%, 100\%]. The $Focus_k(K)$ is indicated by the values of the line. The knowledge graph \textit{cwmo} scores highest in terms of accuracy (for both trained algorithms) and $Focus_k(K)$, while \textit{schema.org} ranks as the lowest. The key observation is that, as anticipated, the accuracy trend for both models generally aligns with the $Focus_k(K)$ ranking for most knowledge graphs. However, an exception is noted with K-NN for the knowledge graph \textit{pext}, performing better than \textit{akt} despite a lower $Focus_k(K)$. Further analysis reveals that this phenomenon may be explained by the interplay between the number of properties and entity types, particularly the balance of the knowledge graph, which can influence the model's predictive performance. The more balanced the knowledge graph, the higher the likelihood of having entity types with low focus. This effect is particularly noticeable when comparing two knowledge graphs with very similar $Focus_k(K)$ but different balances.

This experiment, while illustrating $Focus_k(K)$ as a measurement of the categorization relevance of a knowledge graph, suggests its practical application in assessing the potential performance of a knowledge graph, or a set of knowledge graphs, in a relational prediction task. These findings could be utilized, for example, to fine-tune knowledge graphs in an open-world data extension scenario.

\section{Summary}
In this chapter, we focus on discussing various assessment methods for the knowledge graphs. Firstly, we introduce several state-of-the-art metrics for evaluating knowledge graphs, considering various dimensions. Then, we propose a formal method to evaluate knowledge graphs according to their focus, namely, what cognitive psychologists call categorization purpose. This in turn has allowed us to describe how this assessment plays an important role in supporting an accurate level of knowledge graphs understanding and reuse. In this regard, as preliminary validation of the proposed metrics we are showing: \textit{a).} how focus knowledge graphs ranking reflects the ranking of the knowledge graphs provided by a group of knowledge engineers, following the guidelines inspired by a well-known experiment in cognitive psychology; \textit{b).} how focus can help to select better knowledge graphs to support down-stream tasks.

\chapter{Knowledge Graph Extension: The Platform}

\section{Introduction}
In this chapter, we introduce an online system designed to collect, process, and analyze existing knowledge graphs, aiming to achieve the goal of knowledge graph extension. This study is based on our publications \cite{shi2020learning, chi2022zinet, fumagalli2023liveschema, wang2020mrp2rec}. 

The endeavor of identifying, processing, and transforming resourcing knowledge graphs can be a time-consuming and labor-intensive process. The researchers and knowledge engineers undertake an extensive exploration across available resources and their associated catalogs, often necessitating the examination of various data versions and formats. Furthermore, upon identifying potential candidate resources, it becomes imperative to conduct a thorough analysis of the data to ascertain their reliability in relation to the specific task at hand. This phase not only entails a detailed examination but also pre-supposes that the processing setup is established and executed directly by the researcher. If the necessary data are successfully located, a transformation procedure can be initiated, purposing the relational data to facilitate the evolution of the resourcing knowledge graphs.

The conception of the LiveSchema platform arose from the pivotal challenge of consolidating all aforementioned operations into a singular, comprehensive system, supported by an array of ready-to-use built-in features. Primarily, the system is designed to assist scientists in more efficiently locating the relational data they require. Utilizing updates from some of the most advanced catalogs, LiveSchema aspires to provide an aggregation service that enables the monitoring of the evolving knowledge representation development community from a unified location. Furthermore, by integrating various essential libraries, LiveSchema is positioned to streamline the processes of data preparation and analysis. Generally, these libraries necessitate an ad-hoc setup and may require the integration of multiple components and environments, demanding some extent of coding and development expertise that may not be inherent to all data scientists. Consequently, LiveSchema is envisioned as a platform that integrates data analysis, data processing, and machine learning model deployment. This integration is intended to render these processes more accessible, reusable, and efficient, thereby reducing the time and complexity typically associated with knowledge representation tasks.

As a result, we present the LiveSchema initiative, a pioneering gateway designed primarily to harness the wealth of data encapsulated within knowledge graphs. Presently, LiveSchema integrates $\sim1000$ datasets from four diverse data sources, offering a range of essential facilities. All these facilities are grouped into two main layers: the \textit{data layer} and the \textit{service layer}. Thus, we first introduce the data layer architecture and the process of data collection for the LiveSchema platform. Then, we describe various services developed for utilization, particularly targeting the extension of knowledge graphs. These capabilities enable users to: \textit{i)}  query and process the aggregated knowledge graphs; \textit{ii)}  encode each knowledge graph for analysis and visualization purposes; \textit{iii)}  transform knowledge graphs to aid downstream tasks; \textit{iv)} continuously evolve and expand a catalog of knowledge graphs by aggregating from other source catalogs and repositories. Additionally, we detail the algorithm designed for extending knowledge graphs. Case studies are then presented to prove the efficacy of the LiveSchema platform and the performance of the extensions. LiveSchema is accessible at \url{http://liveschema.eu/} and ready for demonstration.

\section{Data Layer}
The data layer enables a community of users to maintain whole applications and to suggest and upload new resources or edit some already existing resources. It is mainly managed by a group of expert knowledge engineers, software engineers, and data scientists who contribute to the development and evolution of the whole catalog. Moreover, a group of guest users can also be involved in the collaborative development of the storage, by uploading and editing some new datasets and creating new input reference catalogs following well-founded guidelines. In this section, we are introducing how we design the architecture of the data layer, and the details of the data collecting and processing steps.

\subsection{Data Architecture}
The current version of LiveSchema is built upon the CKAN\footnote{https://docs.ckan.org/en/2.9/user-guide.html\#what-is-ckan} open-source data management system, renowned for being one of the most dependable tools for open data management. In defining the data architecture of LiveSchema, we focus on a critical feature made within CKAN, namely the distinction between datasets and resources\footnote{https://docs.ckan.org/en/538-package-install-docs/publishing-datasets.html}. A dataset is conceptualized as a collection of data (for example, the BBC Sport Ontology), which can encompass multiple resources. These resources represent the tangible forms of the dataset in various downloadable formats (such as the BBC Sport Ontology in Turtle or FCA formats). This distinction allows us, as a major advance from mainstream catalogs such as Linked  Open  Vocabularies \cite{LOV2017}, to exploit fine-grained metadata attributes from the Application Profile for European Data Portals (DCAT-AP)\footnote{https://ec.europa.eu/isa2/solutions/dcat-application-profile-data-portals-europe\_en/}, which mirrors a conceptually identical distinction between \textit{dataset} and \textit{distribution}. The additional benefit of employing DCAT-AP lies in that it organizes attributes into categories: mandatory, recommended, and optional. This classification is deemed pivotal in facilitating varying levels of semantic interoperability amongst data catalogs. By adopting this approach, LiveSchema not only aligns with the sophisticated metadata management of DCAT-AP but also enhances the potential for semantic connectivity and data integration across diverse data sources.

Here we demonstrate the metadata specification, i.e. the selected metadata attributes for datasets and distributions considered in the current version:
\begin{itemize}
    \item [i.] \textbf{Dataset:} 
    \begin{itemize}
        \item Mandatory: \textit{description}, \textit{title};
        \item Recommended: \textit{dataset distribution}, \textit{keyword}, \textit{publisher}, \textit{category};
        \item Optional: \textit{other identifier}, \textit{version notes}, \textit{landing page}, \textit{access rights}, \textit{creator}, \textit{has version}, \textit{is version of}, \textit{identifier}, \textit{release date}, \textit{update}, \textit{language}, \textit{provenance}, \textit{documentation}, \textit{was generated by}, \textit{version}.
    \end{itemize}
    
    \item [ii.] \textbf{Distribution:}
    \begin{itemize}
        \item Mandatory: \textit{access url};
        \item Recommended: \textit{description}, \textit{format}, \textit{license};
        \item Optional: \textit{status}, \textit{access service}, \textit{byte size}, \textit{download url}, \textit{release date}, \textit{language}, \textit{update}, \textit{title}, \textit{documentation}.
    \end{itemize}
\end{itemize}
It is imperative to recognize that the metadata distinction between dataset and distribution is non-trivial, particularly considering that metadata attributes such as format, license, byte size, and download URL are pertinent to the distribution, rather than the dataset itself. This demarcation highlights the specific allocation of metadata, ensuring that each attribute is correctly ascribed to its relevant component within the data management structure.

Our initial observation pertains to the two primary benefits that the aforementioned distinction and metadata specification confer upon LiveSchema. First and foremost, the application of metadata facilitates the ``Fairification"\footnote{The FAIR catalog is accessible at \url{https://github.com/OntoUML/ontouml-models}.} \cite{FAIR2016} of the knowledge schemas. This process effectively makes the schemas findable, accessible, interoperable, and reusable, aligning them with the objectives of downstream tasks targeted by LiveSchema. Secondly, the incorporation of enhanced metadata into the knowledge schemas plays a crucial role in initiating, augmenting, and ensuring reproducibility. This observation addresses the potential for future expansion of the metadata specification. A key advantage of incorporating knowledge graph-specific metadata in LiveSchema lies in the ability it grants users to conduct highly customized searches, which enhances the user experience and ensures the retrieval of schemas that are optimally aligned with specific task requirements.

\subsection{Data Collection}
\begin{figure}
  \centering
  \includegraphics[width=1\linewidth]{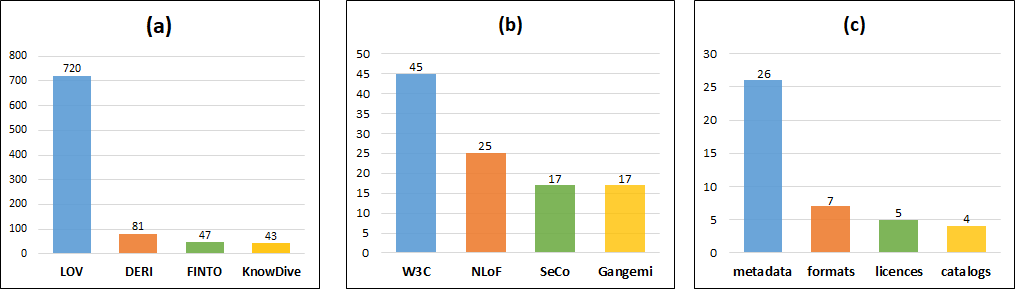}
  \caption{Overall statistics of LiveSchema datasets.}
  \label{LiveSchema_stat}
\end{figure}

At the current state, LiveSchema exploits four principal state-of-the-art catalogs: LOV, DERI,\footnote{\url{http://vocab.deri.ie/}} FINTO\footnote{\url{http://finto.fi/en/}}, and a custom catalog, names KnowDive\footnote{\url{http://liveschema.eu/organization/knowdive}}. The dataset statistics from each of these catalogs are depicted in Figure \ref{LiveSchema_stat}(a)\footnote{Note that the statistics are illustrative since the catalog is currently evolving and may change in the future.}. As it stands, LOV contributes the majority of datasets, establishing itself as one of the most extensively utilized vocabulary catalogs within the semantic web sphere. Each dataset within LiveSchema is linked to a specific provider, which refers to the individual or organization responsible for uploading and maintaining the dataset in its originating catalog. These datasets have been individually scraped and uploaded from each catalog in an automated manner. Figure \ref{LiveSchema_stat}(b) showcases the information regarding these providers, where we find that W3C is the most productive organization in terms of dataset provision.

Figure \ref{LiveSchema_stat}(c) showcases the statistics of several critical pieces of information. From the selected catalogs, 26 types of metadata attributes have been scraped, which include: \textit{i).} catalog attributes, such as \textit{id}, \textit{title} and \textit{description}; \textit{ii).} dataset attributes, like \textit{language}, \textit{publisher}, \textit{license}, \textit{version}, \textit{tags} and \textit{source}; \textit{iii).} provider attributes, such as \textit{id}, \textit{name}, \textit{title} and \textit{uri}. During data scraping, we ensured that no license agreements were breached. Presently, five types of licenses are permitted given their restrictions (all being part of the Creative Commons\footnote{\url{https://creativecommons.org/}} initiative). 

When parsing these datasets in various formats\footnote{\url{https://rdflib.readthedocs.io/en/stable/plugin_parsers.html}}, we serialize them into the most common formats, i.e., \textit{RDF} and \textit{Turtle}. More sophisticated output formats can be generated through the processing operations facilitated by LiveSchema services, such as \textit{CSV} (where all triples and metadata of the input relational data are stored in a datasheet format), \textit{MEX} (providing all the assessment metrics), \textit{FCA} (representing the FCA transformation matrix result), \textit{VIS} (the format enabling visualization services functionalities), and \textit{EMB} (used for generating a statistical model based on a knowledge embedding process).

\subsection{Data Pre-processing}
LiveSchema is managed by a group of knowledge experts who contribute to the development and evolution of the whole system. This group, referred to as \textit{LiveSchema administrators} or simply \textit{admins}, not only addresses maintenance issues but also manages the application of functionalities related to the system's evolution, which are crucial in populating the catalog. Two evolution operations can be executed by the LiveSchema admin. Firstly, LiveSchema encompasses an automated evolution process, consisting of parsing and scraping phases. Admins are provided with a few checkpoints to monitor the outcomes of these automated procedures. Secondly, manual operations such as dataset uploading, reviewing, and managing are also facilitated through LiveSchema services. Some of these datasets are not directly available online and require downloading, unzipping, or editing before being uploaded to GitHub for collection via a URL link. Following data scraping, a secondary key semi-automated parsing process is initiated.

This parsing process, executed iteratively aims to output a set of serialized datasets that are generated by scanning the dataset list and parsing it with the RDFlib Python library\footnote{\url{https://github.com/RDFLib/rdflib}}, a tool used for processing RDF resources. The generated outputs are then used to create more standardized reference formats, currently represented by \textit{RDF} and \textit{Turtle}. The generation of a (\textit{.xlsx} or \textit{.csv}) file encoding all dataset information (e.g., triples) is also supported, thereby facilitating other applications provided by the catalog. In this phase, the admin's key responsibility is to edit the dataset list to exclude unrequired datasets. Then, the dataset will be filtered and transformed into a set of triples as the output of the parsing process.

\section{Service Layer}

\begin{figure}[!t]
   \centering
   \includegraphics[width=0.95\linewidth]{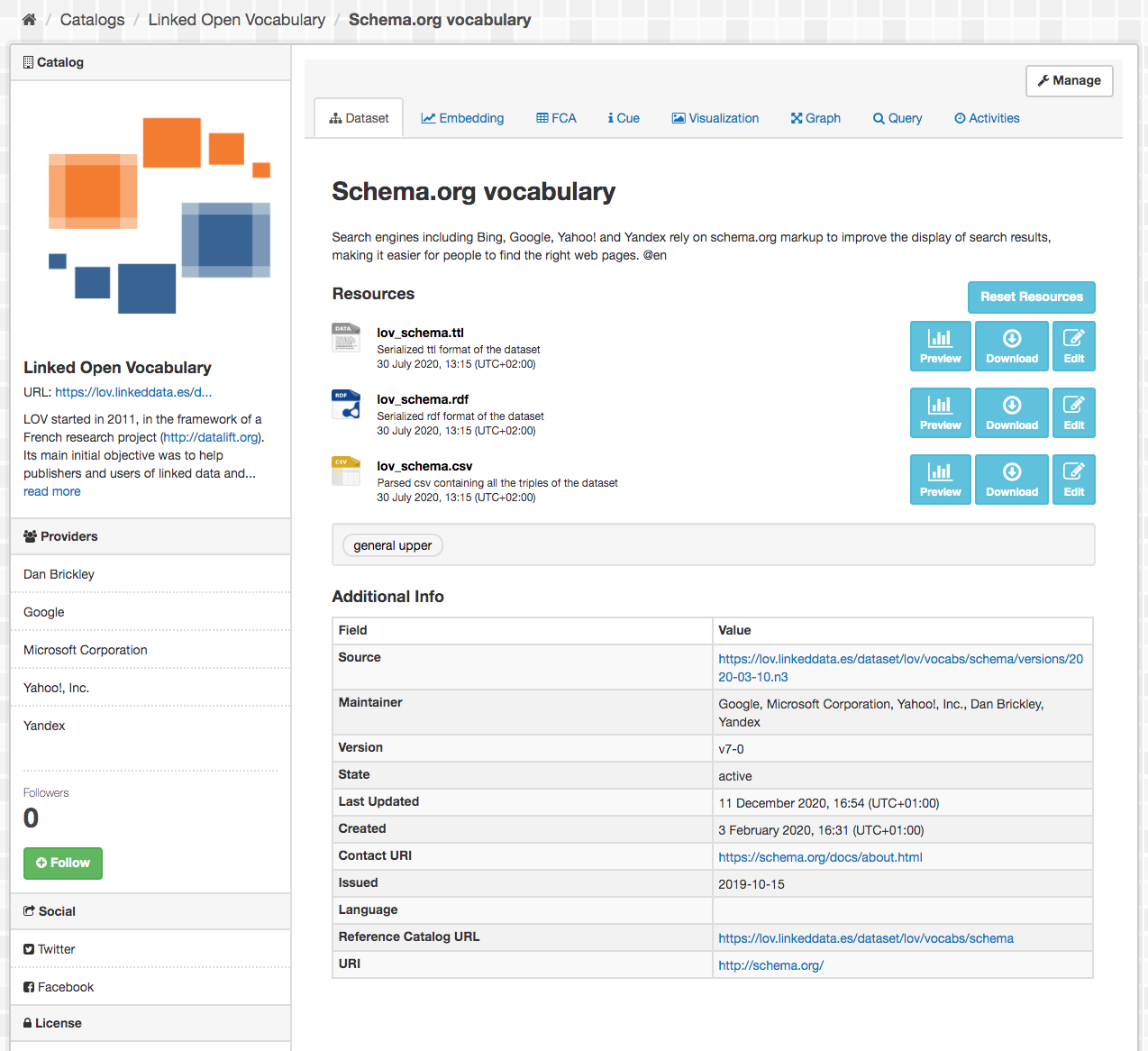}
   \caption{LiveSchema dataset page from an admin perspective.}
   \label{LiveSchema_datapage}
\end{figure}

The LiveSchema service layer offers the connection between users and collected datasets, where all datasets acquired from source catalogs and uploaded to LiveSchema can subsequently be transformed and utilized as input for the platform's available functionalities. The current version of LiveSchema offers several functionalities, including:
\begin{itemize}
    \item Data querying:  the basic function of a knowledge graph repository, enabling the execution of full-text search, fuzzy-matching, and SPARQL queries\footnote{\url{https://rdflib.readthedocs.io/en/stable/intro_to_sparql.html}}, where all datasets and attributes are searchable for providing customized service to the users.

    \item FCA generation: the process of transforming the input knowledge graph format into the FCA format which is designed in section 4.3;

    \item Relational data analyzing: a set of functions that facilitate both visual and parametric analysis of the target knowledge graph, providing knowledge engineers with comprehensive measures for various analysis scenarios.

    \item Metric generation: the process that produces the metrics we introduced in the previous chapters, including similarity measurements for entity type recognition, and Cue validity and Focus metrics for assessing knowledge graphs.

    \item Knowledge graph embedding: the function for encoding the target knowledge graph into high-dimensional deep features using graph neural networks \cite{wang2017knowledge}, aiming to support deep analysis and machine learning pipelines in downstream tasks. 

    \item Knowledge graph visualization: an application for viewing concepts and their associated properties within the target knowledge graphs, via the implementation of the WebVOWL library\footnote{\url{http://vowl.visualdataweb.org/webvowl.html}}.
\end{itemize}
These functionalities are readily accessible and reusable through API services and can be easily extended. Figure \ref{LiveSchema_datapage} presents an example of dataset management, where all functionalities for managing, analyzing, and transforming are displayed\footnote{Current reference resources rdf, ttl, and csv are ready for download.}. Metadata, tags, and information about the reference source catalog are also provided to users on the top left of the link. All the new formats can be accessed on the corresponding functionality page. As some functions, such as FCA generation, have been discussed in earlier chapters, the remainder of this section will focus on detailing the data analysis and knowledge embedding functions, as well as exploring several additional functions anticipated to be available in the forthcoming update.

\subsection{Relational Data Analyzing}
Each LiveSchema dataset is allocated a specific page that presents its information, and from which various processing functionalities can be accessed, as depicted in Figure \ref{LiveSchema_datapage}. A user of LiveSchema has the capability to conduct a straightforward search among the datasets available within the system, subsequently executing analyses to identify the necessary datasets. These analysis functions are accessible via the corresponding tab and facilitate the creation of a corresponding matrix for each knowledge graph. For instance, users have the option to tailor the matrix generation by filtering the target predicates.

In an example scenario, let us consider the task of selecting a reference knowledge graph for recognizing entities of the type `Person' within a collection of tabular data, e.g., originating from an open data repository. This task may entail identifying a knowledge graph that possesses \textit{i).} the target entity type and its corresponding label; \textit{ii).} a substantial number of properties for the target entity type, enhancing the likelihood of aligning properties in the input test data; \textit{iii).} a relatively low count of overlapping properties, aimed at reducing the incidence of false negatives/positives. Suppose the user identifies three potential resources, namely \textit{Schema.org}\footnote{https://schema.org/} (a reference standard for web document indexing), \textit{FOAF}\footnote{http://xmlns.com/foaf/spec/} (a vocabulary commonly used in social network contexts), and the \textit{BBC sport ontology}\footnote{https://www.bbc.co.uk/ontologies/sport} (used by the BBC for modeling sports events and roles). By downloading and comparing all the cue validity and focus values for the entity type \textit{Person} representations across these three sources, as illustrated in Table \ref{Cue values}, a preliminary benchmarking can be conducted. From this analysis, Schema.org emerges as the clear choice, given the high overall evaluation score for the specified entity type.

\begin{table}[!t]
    \centering
    \begin{tabular}{ccccc}
\hline
\textit{Entity type} & \textit{Knowledge graph} & \textit{$Cue_{e}$} & \textit{$Cue_{er}$} & \textit{$Focus_{e}$} \\ \hline
Person & Schema.org & 23 & 0.81 &  0.73  \\
Person & FOAF & 3 & 0.82 & 0.54  \\
Person & BBC & 0.73 & 0.75 & 0.23  \\ \hline
\end{tabular}
    \caption{Analysis of the entity type \textit{person} based on its generated metrics.}
    \label{Cue values}
\end{table}

\begin{figure}[!t]
  \centering
  \includegraphics[width=0.5\linewidth]{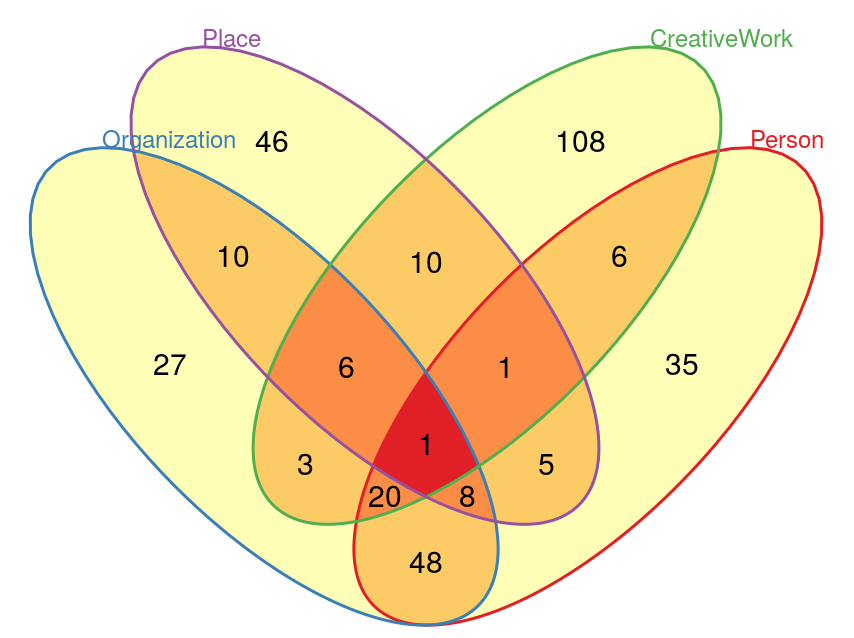}

  \caption{A generated knowledge lotus of the entity type \textit{person}.}
  \label{LiveSchema_lotus}
\end{figure}

In addition to quantifying the generated metrics, further analysis can be conducted by visualizing the intersection of several top entity types from the provided resources. Figure \ref{LiveSchema_lotus} depicts an instance of a knowledge lotus created by LiveSchema, illustrating the diversity of entity types in terms of their shared properties. The yellow petals in the lotus represent the number of properties that are distinctive for the given entity types. In the illustrated example, the entity type `Person' has 35 unique properties, and the red-highlighted petal signifies a property that is shared across all entity types. This kind of visual analysis aids in comprehending the overlap and uniqueness of properties among different entity types within the knowledge graph.

\begin{figure}[!t]
  \centering
  \includegraphics[width=0.85\linewidth]{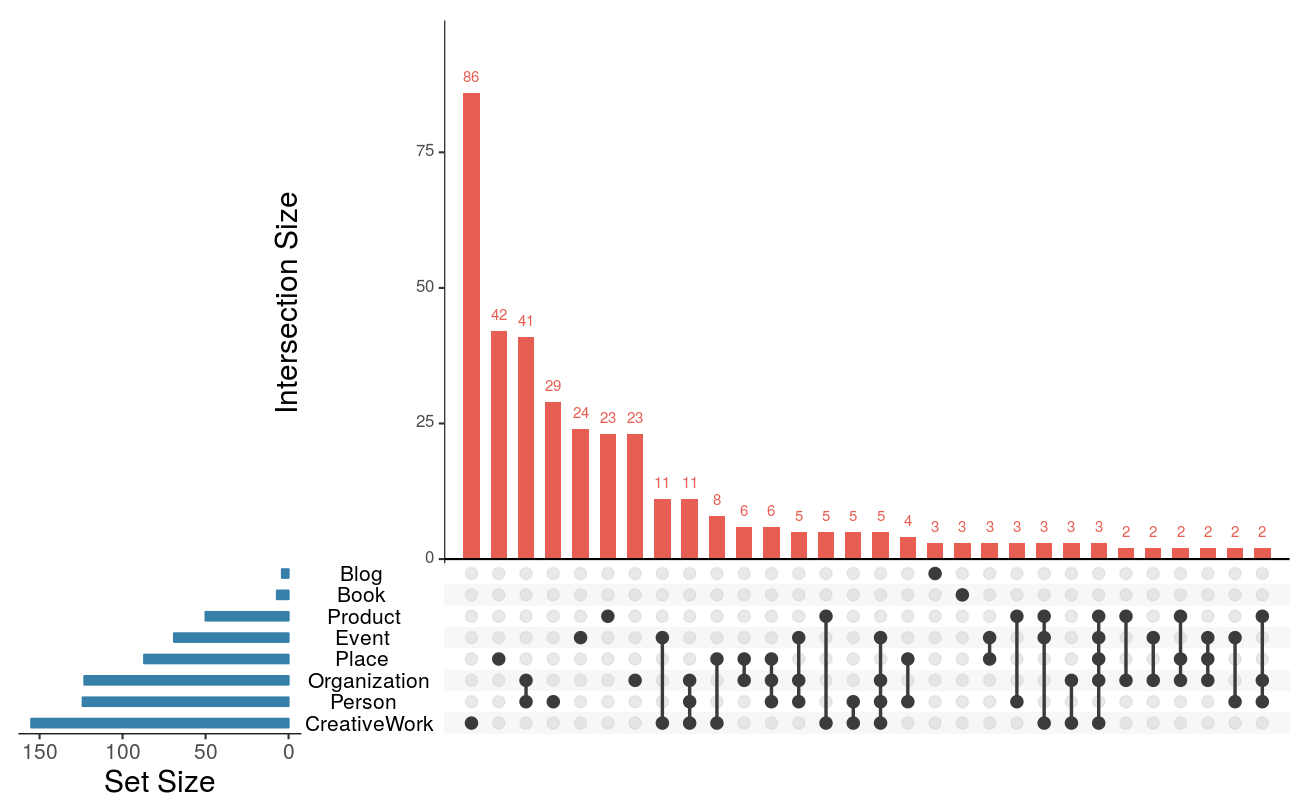}
  \caption{Intersection between entity types and properties by UpSet visualization}
  \label{LiveSchema_upset}
\end{figure}

Further analysis can be conducted through the application of UpSet (multiple set) analysis facilities. LiveSchema facilitates both knowledge lotuses and UpSet visualizations by incorporating the capabilities of the intervention visualization environment\footnote{https://intervene.shinyapps.io/intervene/}. The primary aim of these visualization tools is to simplify the analysis and interpretation of the input resource. An exemplary depiction of a resource using the UpSet module is presented in Figure \ref{LiveSchema_upset}. In this example, eight types are selected. The blue bars on the left side indicate the size of these based on the number of properties. The black dots represent the intersections, while the red bars at the top of the figure denote the size of the intersecting properties set. This approach offers a clear and intuitive understanding of the relational dynamics and commonalities among different resources.

\subsection{Knowledge Graph Embedding}
The knowledge embedding function in LiveSchema encodes the targeted knowledge graph into high-dimensional deep features using graph neural networks. This function is designed to facilitate in-depth analysis, such as graph-based entity similarities, or to support machine learning pipelines in downstream tasks. It is important to note that the current release of LiveSchema incorporates distributional embedding techniques. LiveSchema utilizes the PyKEEN library \cite{ali2021pykeen}, a recent collection encompassing state-of-the-art techniques for graph embedding. PyKEEN is a popular choice for creating custom embedding models. It offers a broad selection of training approaches with various parameters, resulting in a \textit{.pkl} file that can be directly integrated into machine learning pipelines. 

\begin{figure}[!t]
  \centering
  \includegraphics[width=0.9\linewidth]{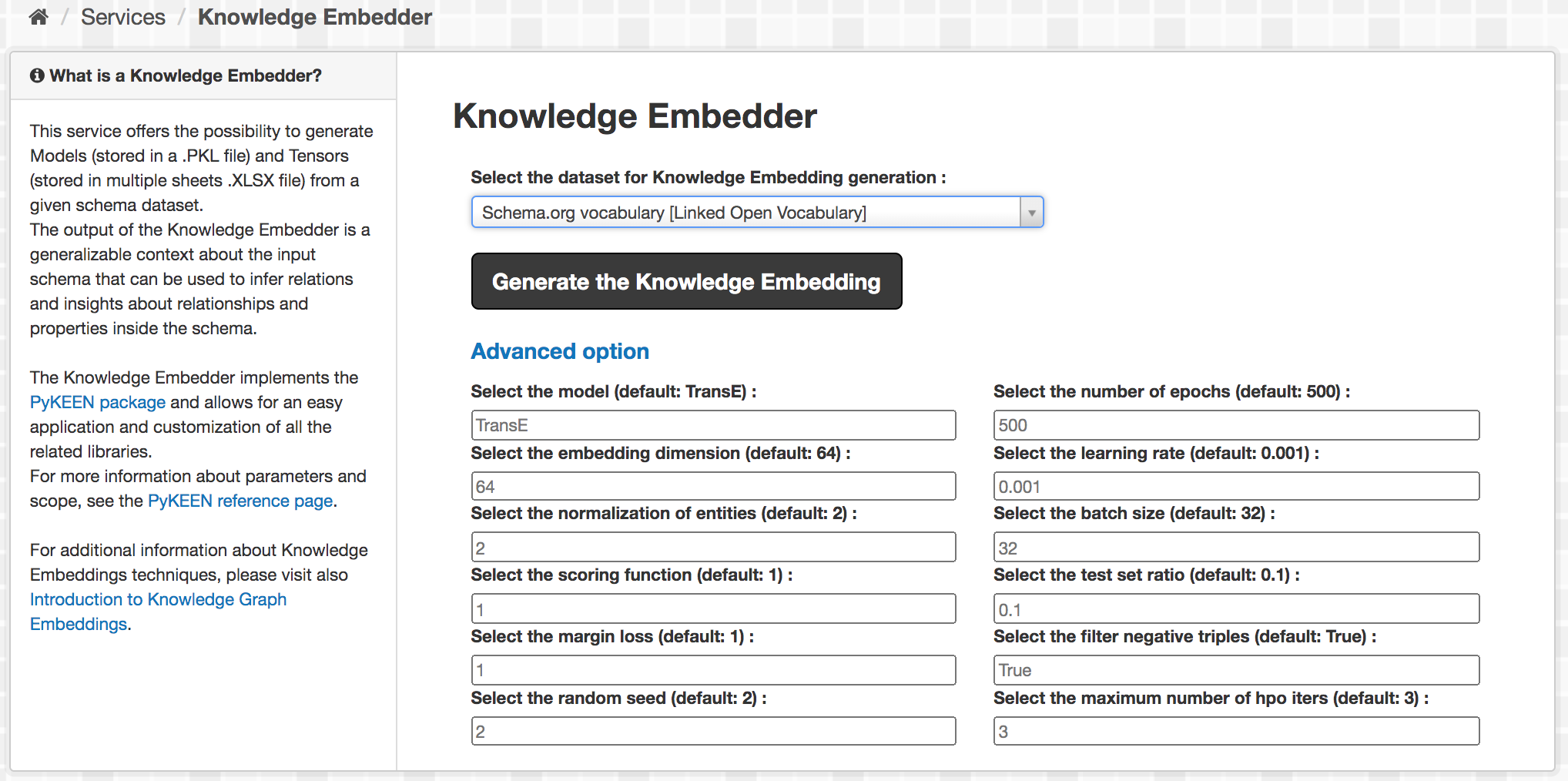}
  \caption{The knowledge embedding interface in LiveSchema platform.}
  \label{LiveSchema_embedding}
\end{figure}

Figure \ref{LiveSchema_embedding} displays a screenshot of the knowledge graph embedding page, where various parameters can be adjusted to achieve specific learning objectives. Users can select the ``embedding model” from a range of state-of-the-art algorithms, such as \textit{TransE}, \textit{RESCAL}, or \textit{DistMult} \cite{wang2017knowledge}. Additionally, settings like the ``loss function” are available. The loss function is typically employed to minimize model error and can condense multiple model features into a single scalar value. This scalar value enables the ranking and comparison of candidate solutions \cite{schmidhuber2015deep}, thereby enhancing the efficacy and applicability of the embedding process.

\subsection{Path Recommendation}
The knowledge graph path recommendation task is an advanced analytical operation in the realm of semantic computing and artificial intelligence. This task involves identifying and predicting optimal paths within a knowledge graph, where a path refers to a sequence of relations connecting a set of entities (types) \cite{shi2016integrating}. The path recommendation task aims to uncover significant paths based on specific criteria or objectives. These objectives could include finding the shortest path, the path toward the most influential concept based on specific properties, or paths that reveal hidden patterns or relationships not immediately apparent. Applications of path recommendation encompass areas such as recommendation systems, semantic search, and data mining. For instance, semantic search can enhance query understanding and result relevance by tracing conceptual links between query terms and information stored in the graph \cite{wang2021learning}. Similarly, recommendation systems can suggest products, services, or content by navigating the relations between user preferences and item attributes. 

Recognizing the significance of the knowledge graph path recommendation task, our research in this domain addresses both methodological and practical aspects. From an application perspective, we have explored the application of path recommendation in online learning environments \cite{saito2020learning}. Effectively sequencing learning objects to create a well-defined learning path can significantly assist e-learners in achieving their learning objectives efficiently and systematically, particularly when dealing with a vast array of fragmented learning content. In response to this need, we developed a knowledge graph-based learning path recommendation system \cite{shi2020learning}. This system utilizes both entity-level and schema-level knowledge to more effectively organize and plan the learning path. Concurrently, we identified six main semantic relations between learning objects within the knowledge graph. Furthermore, we designed a learning path recommendation model that caters to diverse learning requirements and is capable of generating and recommending customized learning paths tailored to the e-learner's specific learning objectives. Our experimental results suggest that the proposed model can successfully generate and recommend personalized learning paths, thereby enhancing the learning experience for e-learners. The implementation of this system significantly improves the depth and accuracy of learning content available to learners, thereby contributing to more efficient and customized educational experiences.

Meanwhile, current knowledge graph path recommendation methods predominantly concentrate on the different selection of deep learning-based knowledge representation models\footnote{Embedding networks like TransE and RESCAL, which encode $<$entity-property-entity$>$ triples into high-dimensional spaces.}. However, these methods generally do not account for the diverse semantic relations between entities as expressed in multi-step relations, which can limit the effectiveness of path recommendation. To address this issue, we introduced a novel framework, namely MRP2Rec \cite{wang2020mrp2rec}, designed to execute path recommendations using latent features derived from both single-step and multiple-step relation paths. More specifically, our approach begins by transforming independent triples into a unified global space to generate multi-step relations. Subsequently, all relations are encoded through a path semantic embedding network, where the encoding facilitates the generation of prediction lists via the inner product in top-K recommendations. Note that the embedding network can be selected from the LiveSchema knowledge embedding functionality. The experimental results demonstrate that the MRP2Rec framework outperforms state-of-the-art methods. This enhanced performance is attributed to the framework's ability to learn more semantic relations from paths involving multiple-step relations. Consequently, our framework can recommend items with implicit features indicative of user collaborative behavior, providing a more nuanced and effective recommendation process.

\section{Knowledge Graph Extension}
In this section, we exhibit the primary application of the proposed platform, specifically focusing on the task of knowledge graph extension, which also constitutes the main objective of this dissertation. Firstly, we will elaborate on the details of the knowledge graph extension algorithm. Subsequently, we will present several case studies to validate the effectiveness of our method.

\subsection{The Algorithm}

In the task of knowledge graph extension, we proceed with the presumption that the input data includes a reference knowledge graph and one or more candidate knowledge graphs. The reference knowledge graph is introduced to selectively integrate content from these candidate knowledge graphs for its expansion. These input knowledge graphs can be identified using the data analysis functionalities available on the LiveSchema platform. Additionally, local knowledge graphs can also be utilized as input sources.

\begin{algorithm}[!t] 
\caption{Extending reference knowledge graph by integrating a candidate knowledge graph. $KG_{ext}=KGExtension(KG_{ref},KG_{cand},EM_{ali})$} 
\label{alg:extension} 
\begin{algorithmic}[1] 
\REQUIRE ~~\\ 
Reference and candidate knowledge graphs $KG_{ref}$, $KG_{cand}$;\\
Aligned entity type pairs $EM_{ali}$; \\
\ENSURE ~~\\ 
The extended reference knowledge graph $KG_{ext}$;\\

\FOR {all $(E_a,E_b) \in EM_{ali}$}
\STATE $E_a.addProperty(E_b.property)$; \{Merging the properties of entity type $E_b$ into $E_a$, where $E_a \in KG_{ref}, E_b \in KG_{cand}$. \}
\STATE $E_a.addInstances(E_b.instances)$; \{Merging the entities of $E_b$ into $E_a$. \}
\FOR {all $E_{sub} \in E_b.subClass$}
\STATE $E_a.addSubClass(E_{sub})$; \{Merging the entity types $E_{sub}$ and their entities into $E_a$.\}
\ENDFOR
\STATE $KG_{ext}.update(E_a)$; 
\ENDFOR

\STATE 	$E_{ref} = listEtypes(KG_{ref})$; 
\STATE 	$I_{cand} = listInstances(KG_{cand})$; 

\FOR {all $I_i \in I_{cand}$}
\IF {$findClass(I_i) \notin EM_{ali}$}
\STATE $E_n = EtypeRecognizer(I_i,E_{ref})$; \{Recognizing the entity type $E_n$ of candidate entities $I_i$, where $E_n, E_{ref} \in KG_{ref}$.\}
\STATE $E_n.addInstances(I_i)$;
\ENDIF

\STATE $KG_{ext}.update(E_n)$
\ENDFOR

\RETURN $KG_{ext}$
\end{algorithmic}
\end{algorithm}
Consequently, let us assume that we are already given the reference and candidate knowledge graphs ($KG_{ref}$ and $KG_{cand}$ respectively), which serve as inputs for the extension algorithm. Let us detail the extension algorithm in Algorithm \ref{alg:extension}. Assume we also have obtained the entity type alignment results by the entity type recognition method presented in Chapter 5. Thus, for each entity type $E_b$ that is aligned with entity type $E_a$ from $KG_{cand}$, we directly add its properties and entities into $E_a$ by $addProperty(\cdot)$ and $addInstances(\cdot)$, respectively. We also consider all subclasses of the aligned entity type $E_b$ since the subclass inherits the properties of the entity type and will bring new entities. If the subclass $E_{sub}$ of $E_b$ is not aligned with any other entity types in $KG_{ref}$, $E_{sub}$ and its associated properties and entities will be merged into $E_a$. Thus, we integrate candidate entities with $KG_{ref}$ when their entity types are able to align with $KG_{ref}$. 

Then, the proposed property-based similarity metrics are applied to align the rest of the candidate entities with entity type $E_{ref}$ from the reference knowledge graph, namely function $Etype Recognizer(\cdot)$, which is also defined in section5.2 as the instance-level entity type recognition. If we successfully align an entity type $E_n$ with the entity $I_i$, $I_i$ will be merged into $E_n$, if not, $I_i$ will be discarded. Regarding the real-world application scenario and topic, the knowledge engineer in our platform will decide if it is needed to integrate the not-aligned entity types with their corresponding entities into $KG_{ref}$. Note that such a strategy for handling the conflict between the $KG_{ref}$ and $KG_{cand}$ is a specific case to the KG extension task since the extension task will consider the usage of the knowledge in the reference KG primarily. However, the conflict of entity types, entities, and properties widely appears between the $KG_{ref}$ and $KG_{cand}$, while the conflict handling strategy can be changed when applying this method to different scenarios and tasks. Finally, all changes will be updated to $KG_{ref}$ and we will obtain an extended knowledge graph $KG_{ext}$ as the final result of our method.  

After we run the extension algorithm, an evaluation process will be conducted to validate the quality of the extended reference knowledge graph, where the metrics introduced in sections 6.2 and 6.3 will be exploited. We compare the metrics on the original reference knowledge graph with the extended version to see if the new-coming concepts from candidate knowledge graphs affect the quality of the reference knowledge graph.

\subsection{Case Study}

We aim to qualitatively assess the extension results by several use cases. In the first case, we demonstrate the process of extending a knowledge graph within a specific domain, focusing on the extension of three datasets that contain Chinese characters from different historical periods, as detailed in the work presented in \cite{chi2022zinet}. Due to the evolutionary nature of characters, there are both semantic continuities and variations among characters from different periods, with some characters maintaining their represented concepts over time while others change or become obsolete. Therefore, it is crucial to properly integrate the consistent characters and to import the new-coming characters as required. The three datasets encompass oracle bone inscriptions, bronze inscriptions, and warring states inscriptions, the statistics of the entity types and entities are illustrated in Table \ref{character statistics}. Each entity type is linked with properties such as sense, glyph, composition structure, deciphering status, and related characters. 

Utilizing the information from these properties, we applied our proposed entity type recognition approach, which facilitated the alignment of 1556 entity types and the identification of the types of 3251 entities. Then, by employing the extension algorithm, we used the warring states inscriptions knowledge graph as the reference to integrate the other candidate knowledge graphs, according to the entity type recognition results. The final extended knowledge graph comprises 6942 entity types and 42151 entities. The correctness of new-coming entities was validated through data verification conducted by three linguists\footnote{The group of linguists consists of PhD candidates and a professor in archeology.}, ultimately confirming the validity of the extended knowledge graph.

\begin{table} [!t]
\caption{The statistics of character knowledge graphs. }
  \label{character statistics}
\centering
\begin{tabular}{ccccc}
\hline
Knowledge graph   & Oracle KG  & Bronze KG   & Warring States KG & Extended KG \\ \hline
Entity Types  & 2543 & 2319 & 5632  & 6942   \\
Entities      & 15175 & 14289  & 28421  & 42151   \\ \hline
\end{tabular}
\end{table}

In the second case study, we endeavor to simulate a real-world situation where individuals extend a general-purpose knowledge graph with domain-specific knowledge graphs to improve usability. We select the widely utilized \textit{schema.org} as the reference knowledge graph. As candidates, we introduce two domain-specific knowledge graphs \textit{Transportation}\footnote{\url{https://carlocorradini.github.io/Trentino_Transportation}} and \textit{Educationtrentino}\footnote{\url{https://alihamzaunitn.github.io/kdi-educationtrentino}} that are specifically designed to support local transportation and education systems, respectively. We chose these knowledge graphs because they offer diverse examples in terms of the number of properties and entity types. Furthermore, the clarity and human-readability of most of their labels facilitate qualitative analysis. We follow the same procedure as in the first case study to recognize the entity types and then proceed to extend the reference knowledge graph. After the extension, the reference knowledge graph encompasses 826 entity types and 1525 properties, reflecting an increase of 23 entity types and 73 properties, respectively. This expansion demonstrates the efficacy of incorporating domain-specific knowledge into a broader knowledge graph, enriching its detail and applicability.

\begin{figure}[!t]
  \centering
  \includegraphics[width=1\linewidth]{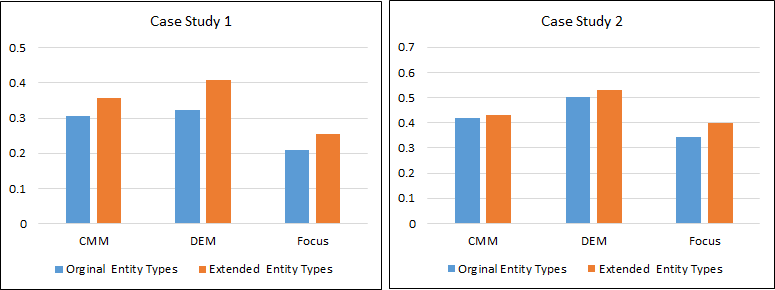}
  \caption{Comparison of the entity types before and after the knowledge graph extension via assessment metrics.}
  \label{extension results}
\end{figure}

To further evaluate the effectiveness of our knowledge graph extension results, we conducted an additional experiment designed to assess the quality of the extended entity types. This experiment is predicated on the idea that our focus is predominantly on the extension of aligned entity types when extracting information from domain-specific resources. For instance, we focus more on aligned entity types such as \textit{University} and \textit{Course} from \textit{Educationtrentino}, \textit{Trip} and \textit{Location} from \textit{Transportation}, or characters that consistently represent the same concepts across different periods. The objective of the experiment is to quantitatively evaluate these aligned entity types using several assessment metrics. Specifically, we compare the metric values of the original and extended entity types to determine whether these entity types have been positively influenced by the extension process. We employ three metrics in this study: \textit{CMM}, \textit{DEM}, and \textit{Focus}, as previously described in section 6.2. These metrics are designed to offer a comprehensive evaluation, encompassing aspects such as coverage, level of knowledge details, and categorization relevance. This experiment encompasses datasets from both case studies, and the results are depicted in Figure \ref{extension results}. Each value represents the average measure of all aligned entity types from the respective case study. It is important to note that the results of all metrics have been normalized to a range of [0,1] to facilitate easy comparison.

The primary outcome of the results indicates that the extension process demonstrates promising performance across all metrics in both cases. In the first case, the objective was to extend a domain-specific knowledge graph, leading to the inclusion of a substantial number of new concepts within the extended entities. The experimental data show significant increases across all three metrics, suggesting considerable enhancement in the informational content. In the second case, a similar increasing trend in metrics was observed, with the extended entity types surpassing their original version in performance. However, the growth across the three metrics was less marked. We attribute this to the fact that the reference schema.org is a general-purpose knowledge graph encompassing a diverse array of concepts, where the limited extension information might not have a significant impact on the metrics.

The second notable observation is that the extended entity types generally demonstrate superior performance across all metrics compared to their original versions. The rise in the \textit{CMM} score indicates that the integration of new concepts has broadened the coverage of entity types. The improvement in the \textit{DEM} score suggests enhanced detail in the entity types of the reference knowledge graph, achieved through the addition of properties and siblings. Moreover, the rising \textit{Focus} score also reveals a trend towards better categorization. Overall, the results from these case studies substantiate the effectiveness of our method in aligning and expanding knowledge graphs.

\section{Summary}
In this chapter, we introduced the LiveSchema platform to collect, process, and analyze existing knowledge graphs, and discussed how to achieve the goal of automatic knowledge graph extension utilizing the services provided by LiveSchema. Initially, we delve into the data layer of the proposed platform, encompassing the unified format and metadata of the collected catalogs, details of data collection, and data pre-processing procedures. We then shifted our focus to the six primary services offered by LiveSchema, detailing the competencies in data analysis, knowledge graph embedding, and path recommendation services. Following this, we elaborate on the knowledge graph extension algorithm, which represents the core purpose of the platform. This includes an explanation of how to leverage the platform's services to facilitate knowledge graph extension, such as similarity calculation, entity type recognition process, and knowledge graph assessment. Finally, we validate the efficacy of the extension through two case studies. These case studies demonstrate the effectiveness of the knowledge graph extension both quantitatively and qualitatively.

\chapter{Conclusion}
\section{Main Conclusion}

The construction of knowledge graphs offers critical support for systems reliant on structured data, particularly for contemporary research in artificial intelligence. Within this context, knowledge graph extension emerges as a pivotal technology, focused on augmenting both the depth and scope of the knowledge graph via the reuse of existing knowledge. Firstly, we discuss and define several primary challenges, i.e., (1) the necessity of a standardized framework to model the task of automatic knowledge graph extension;  (2) the need for precise aligning of knowledge graph concepts to address the description diversity arising from semantic heterogeneity; (3) the requirement of suitable assessment methods to verify the effectiveness of the extended knowledge graph; (4) the demand for a unified platform to customize knowledge graph extension services with consistent processing. In response to these challenges, our main contributions are as follows: (1) we introduce a framework that organizes functionalities into four stages for achieving automatic knowledge graph extension; (2) we propose a method for entity type recognition that employs machine learning models and property-based similarity metrics to improve the accuracy of knowledge extraction; (3) we suggest a set of assessing metrics to validate the quality of the extended knowledge graph; (4) we develop an online platform, namely LiveSchema, which integrates functionalities for the acquisition, management, and extension services of knowledge graphs to benefit knowledge engineers. These principal contributions are then introduced and described in detail in the subsequent chapters.

Therefore, we introduced the modeling of the automatic knowledge graph extension task in Chapter 3. We elaborate on the overall extension framework and its associated functionalities to guide the subsequent contents. At the same time, we also define the utilized data and terminologies in symbolic language. 

In Chapter 4, our focus is primarily on the preparation steps required for entity type recognition. We start by analyzing existing knowledge graphs and discuss the rationale behind using properties to measure similarity between entity types. Additionally, we introduce a method of knowledge formalization based on formal concept analysis lattices. Following this, we present property-based similarity metrics and theoretically contrast them with existing similarity metrics. Finally, we validate the proposed metrics through qualitative experiments.

Chapter 5 presents the machine learning-based entity type recognition method. Initially, we introduce the overall pipeline and its components, which include knowledge pre-processing, the computation of property-based similarity metrics, and machine learning-based recognizers. We distinguish between schema-level and instance-level recognition, as well as the alignment of properties from various sources. A thorough discussion on the selection of learning algorithms is provided, incorporating both classic machine learning models and novel neural networks explored in several studies. Subsequently, we delve into model training, detailing the training procedures and parameter configurations. Lastly, we conduct a series of experiments to evaluate the performance of entity type recognition and the proposed property-based similarity metrics, both quantitatively and qualitatively. 

Chapter 6 is primarily focused on introducing a set of metrics for assessing the quality of extended knowledge graphs. We begin by discussing several existing metrics used in knowledge graph assessment. Following this, we propose our assessing metrics, namely Focus, which are designed to measure the categorization purpose of given entity types based on their associated properties. The experimental results affirm the validity of the Focus metrics.

In Chapter 7, we develop an online knowledge management platform, i.e., Liveschema, with the goal of integrating all the introduced functionalities into a unified system. We detail the data and service layers of the platform, visually depicting Liveschema's development. Importantly, we introduce a knowledge graph extension algorithm that utilizes the functionalities provided by Liveschema. Finally, we present multiple case studies to demonstrate the effectiveness of the extended knowledge graph.

\section{Prospects for Future Work}
In this concluding section of our final chapter, we wish to deliberate on potential avenues for future research. Firstly, there is scope for the entity type recognition method to extend its applicability. Although its current purpose is to identify entity types and entities within candidate knowledge graphs, heterogeneous structured data in real-world research are not confined solely to knowledge graphs but also encompass knowledge bases and tabular data. However, the challenge posed by knowledge conflicts is likely to become more difficult with the increase in data sources. In this paper, we have discussed methods specific to the tasks of entity alignment and knowledge extension in the context of knowledge graph extension, as well as corresponding strategies for conflict resolution. We envisage that the application of the method to a wider range of data types could lead to enhanced depth and accuracy in knowledge extraction. This necessitates further development of the refinement and completion functions of knowledge graphs, which is an insufficient part of our work. Concurrently, we hope for the recognition method to increase processing speeds to facilitate real-time knowledge extraction. Such advancements would significantly boost the effectiveness of the proposed services to knowledge engineers.

With the ongoing progression in the field of artificial intelligence, the demand for structured data represented by knowledge graphs is also increasing. Knowledge graphs frequently serve as a trusted data source for the validation of content generated by artificial intelligence. Consequently, a second potential area for enhancement involves customizing knowledge management services for users based on their unique needs, especially for those requiring more precise domain knowledge. This would entail selectively extracting and integrating target data through a detailed analysis of user requirements. Given the strong foundational information generation capability of large pre-trained language models and their anticipated widespread application scenarios in the future, mining the trustworthiness attributes of knowledge graphs will contribute to addressing the hallucination issues of large language models. The exploration of how to represent the vast amount of domain-specific, trustworthy knowledge within knowledge graphs appropriately and interact with large language models will become an important area of research.

The last direction represents a promising research topic, knowledge graph evolution, namely the continuous enhancement of the reference knowledge graph through the use of streaming data. This procedure can be modeled as a continuous process of knowledge graph extension utilizing multiple data sources. A key challenge is maintaining the integrity and quality of the reference knowledge graph while integrating a multitude of data. One potential approach to address this is by leveraging the concept of continuous learning. In this context, the knowledge graphs are possible to have further refinements after the extension operation, and consequently,  to update machine learning models for the ensuing cycle of knowledge extraction and extension.

\clearemptydoublepage


\thispagestyle{empty}
\makeatletter
\addcontentsline{toc}{chapter}{Bibliography}
\bibliographystyle{plain}
\bibliography{PhD-Thesis}

\clearemptydoublepage









\end{document}